\pdfoutput=1

\documentclass[11pt]{article}
\usepackage[table,xcdraw]{xcolor}
\usepackage[final]{acl}

\usepackage{times}
\usepackage{latexsym}
\usepackage{enumitem}
\usepackage{amssymb}
\usepackage{amsfonts}
\usepackage{amsmath}
\usepackage{bbm}

\usepackage[T1]{fontenc}

\usepackage{booktabs}
\usepackage[utf8]{inputenc}
\usepackage[capitalize]{cleveref}
\usepackage{microtype}
\usepackage{subfigure} 
\usepackage{inconsolata}
\usepackage{graphicx}
\usepackage{stfloats}
\usepackage{adjustbox}
\usepackage[normalem]{ulem}
\useunder{\uline}{\ul}{}

\usepackage{colortbl}
\usepackage{bbm}
\title{Evaluating Mathematical Reasoning of Large Language Models:\\A Focus on Error Identification and Correction}

\author{Xiaoyuan Li\textsuperscript{1},
        Wenjie Wang\textsuperscript{2}\thanks{Corresponding author.},
        Moxin Li\textsuperscript{2},
        Junrong Guo\textsuperscript{1},
        Yang Zhang\textsuperscript{1},
        Fuli Feng\textsuperscript{1}\footnotemark[1] \\
        University of Science and Technology of China\textsuperscript{1} \\
        National University of Singapore\textsuperscript{2} \\
        \texttt{\{xiaoyuanli,godrong,zy2015\}@mail.ustc.edu.cn limoxin@u.nus.edu} \\
        \texttt{\{wenjiewang96,fulifeng93\}@gmail.com}
        }

\begin{document}
\maketitle
\begin{abstract}

The rapid advancement of Large Language Models (LLMs) in the realm of mathematical reasoning necessitates comprehensive evaluations to gauge progress and inspire future directions. 
Existing assessments predominantly focus on problem-solving from the examinee perspective, overlooking a dual perspective of examiner regarding error identification and correction.
From the examiner perspective, we define four evaluation tasks for error identification and correction along with a new dataset with annotated error types and steps. 
We also design diverse prompts to thoroughly evaluate eleven representative LLMs. 
Our principal findings indicate that GPT-4 outperforms all models, while open-source model LLaMA-2-7B demonstrates comparable abilities to closed-source models GPT-3.5 and Gemini Pro.
Notably, calculation error proves the most challenging error type. 
Moreover, prompting LLMs with the error types can improve the average correction accuracy by 47.9\%. These results reveal potential directions for developing the mathematical reasoning abilities of LLMs.
Our code and dataset is available on \href{https://github.com/LittleCirc1e/EIC}{https://github.com/LittleCirc1e/EIC}. 
\end{abstract}

\section{Introduction}

\begin{figure}[htbp]   
\centering
\setlength{\abovecaptionskip}{0.1cm}
\setlength{\belowcaptionskip}{0cm}
\includegraphics[width=0.86\linewidth,scale=1.00]{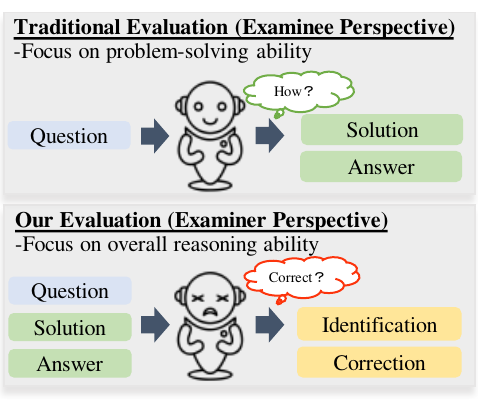}
\caption{Traditional evaluation on problem-solving and our evaluation on error identification and correction.}
\label{overview}
\vspace{-15pt}
\end{figure}

Large Language Models (\citealp{brown2020language}; \citealp{ouyang2022training}; \citealp{anil2023palm}; \citealp{openai2023gpt})  have been successfully applied to mathematical reasoning, particularly in the Math Word Problems (MWP) (\citealp{kushman2014learning}; \citealp{roy2018mapping}). LLMs cultivate a nuanced understanding of number-intensive context and multi-step reasoning. Cutting-edge models such as GPT-4 \citep{openai2023gpt} have demonstrated impressive performance in addressing mathematical problems. For example, it has achieved an accuracy of 97\% on the GSM8K dataset \citep{zhou2023solving}.
With the rapid advancement of LLMs, evaluating their effectiveness and reliability becomes increasingly crucial.

Existing evaluations are mainly from the examinee perspective, which directly assess the problem-solving capability of LLMs regarding the correctness of answers (\citealp{shakarian2023independent}; \citealp{fu2023chain}; \citealp{hong2024stuck}; \citealp{shi2022language}) and the consistency of intermediate reasoning steps (\citealp{wei2022chain}; \citealp{golovneva2022roscoe}; \citealp{zhang2023evaluating}; \citealp{gaur2023reasoning}). However, current research rarely delve into a dual perspective of examiner, \textit{i.e.,} the ability to identify and correct errors (Figure \ref{overview}), which is equally crucial as problem-solving and worthwhile exploring. On one hand, the performance of traditional evaluation tasks is almost approaching saturation, calling an urgent need for new perspectives of evaluation. On the other hand, accurate error recognition and correction can facilitate the development of problem-solving capability.

Aiming to construct fine-grained evaluation on error recognition and correction, we define four distinct tasks. These tasks are as follows: \textit{1)} Error-Presence Identification (\textbf{EP}): Identifying whether any error exists in the entire solution. \textit{2)} Error-Step Identification (\textbf{ES}): Identifying the first wrong step within the solution, which is the root cause of error. \textit{3)} Error-Type Identification (\textbf{ET}):  Identifying the error type present in the first wrong step, such as calculation error. \textit{4)} Error Correction (\textbf{EC}): Rectifying the wrong steps and obtaining the final corrected answer. To our knowledge, we are the first to comprehensively define the four evaluation tasks for error identification and correction regarding mathematical reasoning.


Moving one step further, we consider constructing the evaluation dataset for these four tasks. Given a question, the dataset should include ground-truth answers, solutions with errors, step numbers of wrong steps, and types of errors. To construct this dataset, we need to define the types of error first. By collating examples from existing studies and practical instances, we distill nine common error types. Subsequently, harnessing the exceptional text generation capability of GPT-4 \citep{openai2023gpt}, 
we transform initially correct solutions of GSM8K \citep{cobbe2021training} and MathQA \citep{amini2019mathqa} into solutions featuring single-step and single-type errors. Through this approach, we establish a dataset comprising 1800 instances to evaluate the ability to recognize and rectify errors.

Based on the evaluation dataset, we test closed-source models, including GPT-3.5 \citep{ouyang2022training}, GPT-4 \citep{openai2023gpt}, GLM-4 \citep{du2022glm}, Gemini Pro \citep{team2023gemini}, and their open-source counterparts such as LLaMA-2-7B, LLaMA-2-13B \citep{touvron2023LLaMA}, MetaMath-7B, MetaMath-13B \citep{yu2023metamath},  Mistral-7B \citep{jiang2023mistral}, Llemma-7B \citep{azerbayev2023llemma} and LEMA-7B \citep{an2023learning}. We devise diverse prompts of each task to evaluate the robustness of these LLMs. Through extensive experiments, we derive five key findings:  \textit{1)}  Across all four tasks, GPT-4 exhibits outstanding performance compared to other models with GLM-4 closely following. GPT-3.5, Gemini Pro, and LLaMA-2-7B show varying strengths and weaknesses. \textit{2)}  While GPT-4 and GLM-4 demonstrate overall competence across four tasks, their ability to identify and rectify calculation error lags behind other error types. This suggests a need for further enhancement of the calculation capability of LLMs. \textit{3)}  In the task of \textit{ET}, many error types are easily recognized as the type of calculation error, with the type of missing step proving to be the most challenging to identify. \textit{4)}  In the task of \textit{EC} and \textit{ES}, by providing the error types, the average accuracy can be improved by 47.9\% and 45.9\%, respectively. \textit{5)} Open-source models are highly influenced by prompts, while closed-source models demonstrate a comparatively robust performance.

Our contributions can be summarized as follows: \textit{1)}  We define four tasks for evaluating the mathematical reasoning ability of LLMs regarding error identification and correction. To our knowledge, our work represents the first comprehensive assessment of the fine-grained capability of LLMs in recognizing and rectifying errors. \textit{2)} We define nine common error types and provide a dataset based on these error types. The dataset is intended to facilitate a more nuanced examination of the LLMs' performance in handling different error scenarios. \textit{3)} Through the comprehensive evaluation of four commercial and seven open-source LLMs, we derive key findings that hold insightful implications for the subsequent advancement of LLMs. 

\begin{figure*}[ht]   
\setlength{\abovecaptionskip}{0.1cm}
\setlength{\belowcaptionskip}{0cm}
\centering
\includegraphics[width=0.8\linewidth,scale=1.00]{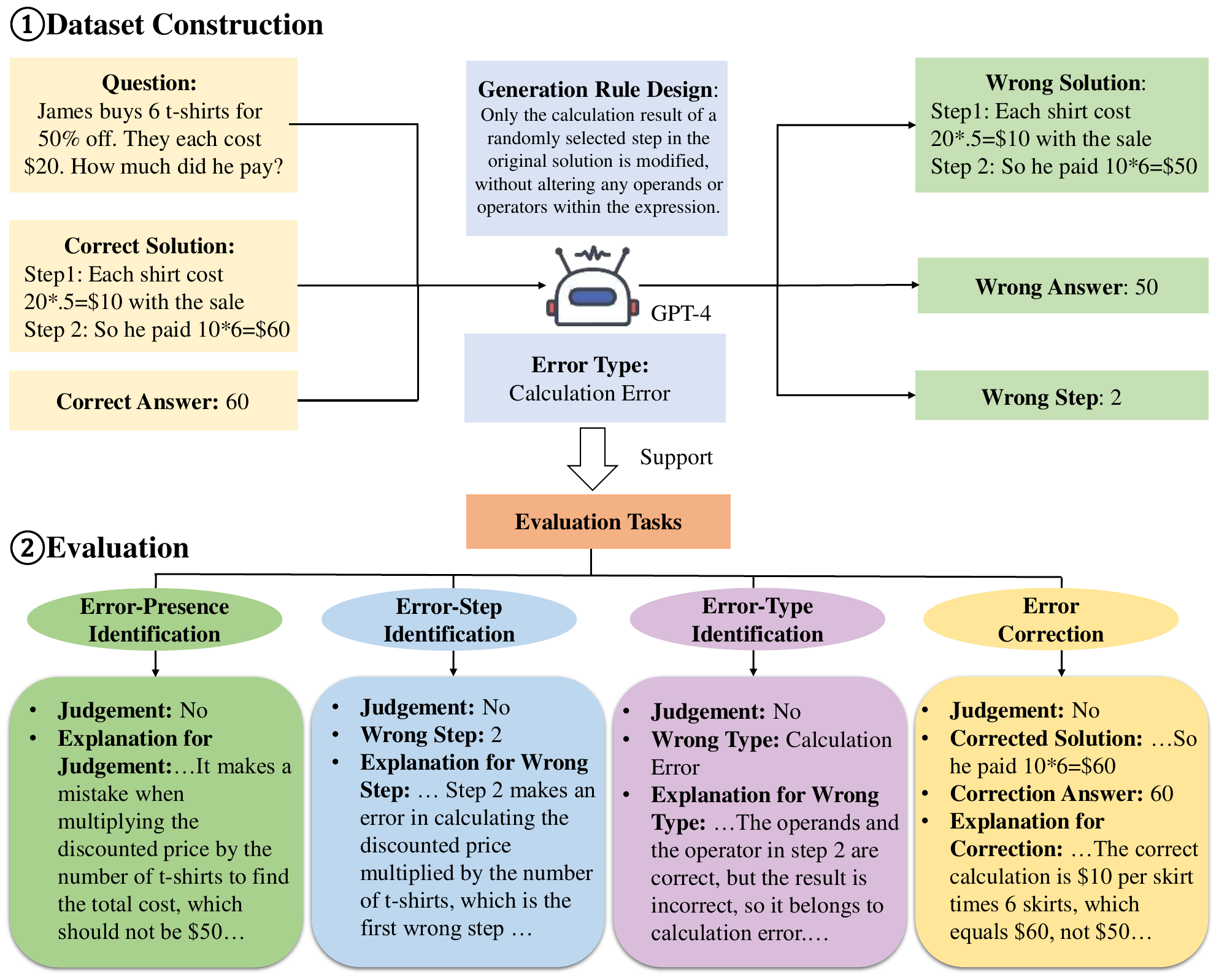}
\caption{Illustration of dataset construction and the four evaluation tasks. For dataset construction, we use GPT-4 to convert ground-truth solutions into wrong solutions containing specific error types. The four evaluation tasks comprehensively access LLMs' error identification and correction abilities from diverse perspectives. 
}
\label{framework_new}
\vspace{-12pt}
\end{figure*}

\section{Task Formulation}
As the proverb goes, \textit{errors are the stepping stones to wisdom}. If a model is adept at mathematical reasoning, it should excel at identifying and correcting errors. Hence, we gauge the mathematical reasoning abilities of LLMs by assessing their proficiency in recognizing and rectifying errors.
To comprehensively accomplish the evaluation, as shown in Figure~\ref{framework_new}, we define four tasks at a fine-grained level of error identification and correction. 
\begin{itemize}[leftmargin=*]
    \item \textbf{Task 1: Error-Presence Identification (EP)} aims to detect whether any error exists in the solution of a mathematical question. Formally, given a mathematical question $q$ with an LLM-generated solution $s$, \textit{EP} estimates the binary label ${y}$ that indicates whether $s$ contains errors. 
    We design three prompts for open-source and closed-source LLMs for \textit{EP}\footnote{Due to certain limitations in the capabilities of open-source models, they may not fully comply with the strict JSON format requirements like closed-source models. Therefore, we design prompts with relaxed formatting tailored for open-source models.}: \textit{Simple} requires LLMs to only output the judgment $\hat{y}$; \textit{Normal} requires to not only output the judgment but also provide an explanation; \textit{Misleading} informs LLMs that there might be errors in the solution and instructs LLMs to generate the judgment with explanation. 
    To save space, we move detailed prompts to Figure \ref{fig: Simple prompt for EP on closed-source models} to \ref{fig: Misleading prompt for EP on open-source models}. 
    For evaluation of \textit{EP}, we compute the accuracy in identifying error presence by $acc_1=\frac{1}{N} \sum_{i=1}^N \mathbbm{1}\{y=\hat{y}\}$, where $N$ is the number of evaluation cases and $\hat{y}$ is the predicted label by LLMs. 
    
    \item \textbf{Task 2: Error-Step Identification (ES)} intends to find the first wrong step $t$ in a wrong solution. For \textit{Task 2}, we require LLMs to output the judgment $\hat{y}$, and if $s$ contains errors, we also instruct LLMs to identify the first erroneous step $t$ in the solution. 
    We devise the zero-shot prompts and few-shot prompts with in-context learning examples for the \textit{ES} task. Figure \ref{fig: Zero-shot prompt for ES on closed-source models} and \ref{fig: Zero-shot prompt for ES on open-source models} show the zero-shot prompts for open-source and closed-source models. 
    For \textit{ES} evaluation, we compute $acc_{1}$ as \textit{EP} and the accuracy in identifying error step by $acc_2=\frac{1}{N} \sum_{i=1}^N \mathbbm{1}\{t=\hat{t}\}$, where $\hat{t}$ denotes the first wrong step predicted by LLMs. 
    
    \item \textbf{Task 3: Error-Type Identification (ET)} endeavors to identify the error type. We instruct LLMs to output the judgment for $y$ and identify the error type $c$ of the first wrong step if $s$ contains errors. 
    Here, $c$ is selected from the pre-defined error types, such as calculation error. 
    We define error types in the prompts and design zero-shot and few-shot prompts, where the few-shot prompt provides an example for each error type. 
    Figure \ref{fig: Zero-shot prompt for ET on closed-source models} and \ref{fig: Zero-shot prompt for ET on open-source models} showcase the zero-shot prompts for open-source and closed-source models. 
    Considering that the order of error types might affect the accuracy of identifying error types, we design prompts that reverses the original order of error types and randomly shuffles them. 
    We compute $acc_{1}$ and the accuracy in identifying error type $acc_3=\frac{1}{N} \sum_{i=1}^N \mathbbm{1}\{c=\hat{c}\}$, where $\hat{c}$ is the error type of the first wrong step identified by LLMs.

    \item \textbf{Task 4: Error Correction (EC)} seeks to rectify the error and output the correct solution. 
    We prompt LLMs to output the judgment for $y$ and provide the corrected solution and answer $\hat{a}$ if $s$ contains errors. 
    We devise zero-shot and few-shot prompts as \textit{ES}. The prompts are displayed in Figure \ref{fig: Zero-shot prompt for EC on closed-source models} and \ref{fig: Zero-shot prompt for EC on open-source models}. 
    We calculate $acc_{1}$ and the accuracy of correction $acc_4=\frac{1}{N} \sum_{i=1}^N \mathbbm{1}\{a=\hat{a}\}$, where $\hat{a}$ and $a$ are the predicted and ground-truth answers, respectively. 
\end{itemize}
For \textit{Task 2} and \textit{Task 4}, we propose to leverage the error type information in the prompts to hint LLMs for error step identification and error correction. Accordingly, we design the zero-shot and few-shot prompts with error type information as shown in Figure \ref{fig: Zero-shot-type prompt for ES on closed-source models}, \ref{fig: Zero-shot-type prompt for ES on open-source models}, \ref{fig: Zero-shot-type prompt for EC on closed-source models} and \ref{fig: Zero-shot-type prompt for EC on open-source models}.

\section{Dataset Construction}

\begin{table*}
\setlength{\abovecaptionskip}{0.1cm}
\setlength{\belowcaptionskip}{-0.30cm}
\small
\begin{tabular*}{\linewidth}{ll}
\toprule
\textbf{Error Type} &  \textbf{Definition} \\
\midrule
\textbf{Calculation Error (CA)} &  Error appears during the calculation process.  \\
\textbf{Counting Error (CO)}  & Error occurs during the counting process.  \\
\textbf{Context Value Error (CV)} & Error arises when attributes of named entities do not align with the information provided.  \\
\textbf{Hallucination (HA)}  & Error involves adding fictitious unrelated statements contradictory to the question.  \\
\textbf{Unit Conversion Error (UC)} & Error occurs during unit conversion process. \\
\textbf{Operator Error (OP)} & Error involves a single operator being erroneously applied within the expression. \\
\textbf{Formula Confusion Error (FC)} &  Error appears when applying formula in inappropriate scenario.  \\
\textbf{Missing Step (MS)}  & Error entails an incomplete generation of reasoning process, lacking a necessary step.\\
\textbf{Contradictory Step (CS)} & Error manifests inconsistency between preceding and subsequent reasoning steps.\\
\bottomrule
\end{tabular*}

\caption{Definition of nine common error types. Among them, unit conversion error, operator error, and formula confusion error can be categorized as \textbf{common sense error}, indicating errors in the relationships that should be understood within worldly common sense. The generation rules and examples are designed in Appendix \ref{sec:appendix rules}.}

\label{tab:error type definition}
\vspace{-5pt}
\end{table*}
A significant challenge in achieving the four evaluation tasks is lacking compatible datasets with fine-grained error annotation. 
Therefore, we opt to construct a dataset that meets the requirements of our evaluation tasks. This dataset should encompass erroneous solutions, error steps, error types, and correct answers for mathematical questions.

Initially, we distill nine common error types from existing works (\citealp{wei2022chain}; \citealp{toh2023veritymath}; \citealp{lightman2023lets}; \citealp{shakarian2023independent}; \citealp{bubeck2023sparks}; \citealp{sawada2023arb}; \citealp{suzgun2022challenging}; \citealp{lyu2023faithful}; \citealp{kojima2022large}; \citealp{li2023making}; \citealp{wang2022towards}; \citealp{wang2023plan}; \citealp{paul2023refiner}; \citealp{golovneva2022roscoe}; \citealp{ribeiro2023street}; \citealp{lewkowycz2022solving}) and practical examples. 
Table \ref{tab:error type definition} shows the error names and definitions, covering the single-step and cross-step errors. 
The specific definition difference and illustration examples are presented in Appendix \ref{Detailed Error Type Definition}.

\textbf{Data Generation.}
As illustrated in Figure~\ref{framework_new}, we utilize the state-of-the-art LLM, GPT-4 \citep{openai2023gpt}, to generate the dataset, \textbf{EIC-Math} (\textbf{E}rror \textbf{I}dentification and \textbf{C}orrection on \textbf{Math}ematical problems), to support the evaluation tasks. 
We design some generation rules for different error types, which regulate the generated wrong solutions to strictly meet the definition of one error type\footnote{In this work, we only consider generating the wrong solution with only one error type in a single step to simplify the evaluation process, leaving more complicated error identification and correction to future work.}. 
Then we construct the data generation prompt based on these generation rules and the in-context learning approach (\citealp{brown2020language}; \citealp{ouyang2022training}; \citealp{min-etal-2022-rethinking}) to instruct GPT-4 to transform correct solutions into wrong solutions. 
The data generation process is detailed in Appendix \ref{Dataset Generation} to save space. 
Note that we use two datasets GSM8K \citep{cobbe2021training} and MathQA \citep{amini2019mathqa} to construct the error cases, where GSM8K has annotated multi-step solutions and MathQA adopts the correct solutions generated by GPT-3.5. 
Each dataset is comprised of 100 cases per error type, resulting in a total of 1,800 cases for error identification and correction tasks.

\textbf{Human Evaluation.}
To evaluate the quality of EIC-Math, we randomly select 180 cases and invite three evaluators for human evaluation. The results indicate that 92.5\% cases have exactly satisfied the requirements of the data generation prompts, demonstrating the high quality of the generated dataset. 
More details of human evaluation can be found in Appendix \ref{Human Evaluation}.

\section{Experiment}
We conduct extensive experiments to address the following research questions:

\noindent %
\textbf{- RQ1:} How do different LLMs perform on the four tasks on error identification and correction?

\noindent %
\textbf{- RQ2:} How difficult are identifying and correcting different error types? 

\noindent %
\textbf{- RQ3:} How robust are LLMs to different prompts \textit{w.r.t.} the four evaluation tasks? 

\vspace{5pt}
\noindent\textbf{Experiment Setup.} We select typical commercial closed-source LLMs, GPT-3.5, GPT-4, GLM-4, Gemini Pro, along with the general-purpose open-source LLaMA-2 series, and the state-of-the-art mathematical MetaMath series in their 7B and 13B versions for evaluation. Besides, we also evaluate other three cutting-edge mathematical models: Mistral, Llemma and LEMA in their 7B versions. \footnote{Specifically, we conduct experiments using gpt-3.5-turbo-1106, gpt-4-1106-preview, LLaMA-2-7B-chat, LLaMA-2-13B-chat, MetaMath-7B-V1.0, MetaMath-13B-V1.0, Mistral-7B-V0.1, Llemma-7B, LEMA-V1-PEFT-LLaMA-2-7B-GSM8K.} To minimize randomness, we set the temperature to 0. For ease of statistical analysis, we prompt closed-source LLMs to output in JSON format. However, open-source models do not consistently adhere to the format requirement, so we use a relaxed format for their prompts. 

\begin{table*}[h!]
\centering
\setlength{\abovecaptionskip}{0.1cm}
\setlength{\belowcaptionskip}{-0.05cm}
\tabcolsep=0.1cm
\scalebox{0.67}{
\begin{adjustbox}{center}
\begin{tabular}{cccrcrcrrr|ccrcrcrrr|rr}
\hline
 &
  \multicolumn{9}{c|}{GSM8K} &
  \multicolumn{9}{c|}{MathQA} &
  \multicolumn{1}{c}{} &
  \multicolumn{1}{c}{} \\
 &
  EP &
  \multicolumn{2}{c}{ES} &
  \multicolumn{2}{c}{ET} &
  \multicolumn{2}{c}{EC} &
  \multicolumn{2}{c|}{Avg} &
  EP &
  \multicolumn{2}{c}{ES} &
  \multicolumn{2}{c}{ET} &
  \multicolumn{2}{c}{EC} &
  \multicolumn{2}{c|}{Avg} &
  \multicolumn{2}{c}{Avg} \\ 
  &
  $acc_{1}$ &
  \multicolumn{1}{c}{$acc_{2}$} &
  \multicolumn{1}{c}{$acc_{1}$} &
  \multicolumn{1}{c}{$acc_{3}$} &
  \multicolumn{1}{c}{$acc_{1}$} &
  \multicolumn{1}{c}{$acc_{4}$} &
  \multicolumn{1}{c}{$acc_{1}$} &
  \multicolumn{1}{c}{$acc$} &
  \multicolumn{1}{c|}{$acc_{1}$} &
  $acc_{1}$ &
  \multicolumn{1}{c}{$acc_{2}$} &
  \multicolumn{1}{c}{$acc_{1}$} &
  \multicolumn{1}{c}{$acc_{3}$} &
  \multicolumn{1}{c}{$acc_{1}$} &
  \multicolumn{1}{c}{$acc_{4}$} &
  \multicolumn{1}{c}{$acc_{1}$} &
  \multicolumn{1}{c}{$acc$} &
  \multicolumn{1}{c|}{$acc_{1}$} &
  \multicolumn{1}{c}{$acc$} &
  \multicolumn{1}{c}{$acc_{1}$}\\ 
  \hline
GPT-3.5 &
  0.547 &
  0.147 &
  0.598 &
  0.211 &
  0.737 &
  0.169 &
  0.340 &
  0.269 &
  0.556 &
  0.493 &
  0.173 &
  0.642 &
  0.173 &
  0.676 &
  0.141 &
  0.302 &
  0.245 &
  0.528 &
  0.257 &
  0.542 \\
GPT-4 &
  0.930 &
  0.843 &
  0.946 &
  0.516 &
  0.951 &
  0.883 &
  0.929 &
  0.793 &
  0.939 &
  0.917 &
  0.714 &
  0.954 &
  0.481 &
  0.957 &
  0.810 &
  0.909 &
  0.731 &
  0.934 &
  0.762 &
  0.937 \\
GLM-4 &
  0.849 &
  0.640 &
  0.819 &
  0.349 &
  0.941 &
  0.804 &
  0.881 &
  0.661 &
  0.873 &
  0.772 &
  0.551 &
  0.892 &
  0.327 &
  0.910 &
  0.574 &
  0.808 &
  0.556 &
  0.846 &
  0.609 &
  0.860 \\
Gemini Pro &
  0.217 &
  0.359 &
  0.541 &
  0.090 &
  0.312 &
  0.248 &
  0.279 &
  0.229 &
  0.337 &
  0.197 &
  0.239 &
  0.389 &
  0.096 &
  0.603 &
  0.200 &
  0.260 &
  0.183 &
  0.362 &
  0.206 &
  0.350 \\ \hline
LLaMA-2-7B &
  0.538 &
  0.184 &
  0.914 &
  0.048 &
  0.396 &
  0.067 &
  0.871 &
  0.209 &
  0.680 &
  0.536 &
  0.176 &
  0.861 &
  0.052 &
  0.358 &
  0.039 &
  0.792 &
  0.201 &
  0.637 &
  0.205 &
  0.659 \\
LLaMA-2-13B &
  0.166 &
  0.007 &
  0.027 &
  0.127 &
  0.843 &
  0.000 &
  0.008 &
  0.075 &
  0.261 &
  0.219 &
  0.009 &
  0.071 &
  0.116 &
  0.939 &
  0.000 &
  0.010 &
  0.086 &
  0.310 &
  0.081 &
  0.286 \\ \hline
Avg &
  \multicolumn{1}{r}{0.541} &
  \multicolumn{1}{r}{0.363} &
  0.641 &
  \multicolumn{1}{r}{0.224} &
  0.697 &
  \multicolumn{1}{r}{0.362} &
  0.551 &
  0.372 &
  0.608 &
  0.522 &
  0.310 &
  \multicolumn{1}{c}{0.635} &
  0.208 &
  \multicolumn{1}{c}{0.741} &
  0.294 &
  \multicolumn{1}{c}{0.514} &
  \multicolumn{1}{c}{0.334} &
  \multicolumn{1}{c|}{0.603} &
  0.353 &
  0.605 \\ \hline
\end{tabular}
\end{adjustbox}
}
\caption{Average accuracy of different models in four tasks on GSM8K and MathQA separately under zero-shot prompts. \textit{EP} calculates the average $acc_{1}$ over all error types. \textit{ES} calculates the average $acc_{2}$ and $acc_{1}$ as the values for the first and second column respectively. And \textit{ET} and \textit{EC} conduct similar calculation as \textit{ES}. The first column of \textit{Avg} is the average value of $acc_{1}$, $acc_{2}$, $acc_{3}$, and $acc_{4}$ over all error types of models and represents the ability to identify and correct errors, while the second column is the average value of $acc_{1}$ of four tasks and only represents the ability to identify errors. }
\label{tab:Based on Model, main table}
\end{table*}

\begin{table*}[h!]
\centering
\setlength{\abovecaptionskip}{0.1cm}
\setlength{\belowcaptionskip}{-0.05cm}
\tabcolsep=0.1cm
\scalebox{0.67}{
\begin{adjustbox}{center}
\centering
\begin{tabular}{ccccccccccccc|cccccc|cc}
\hline
 &
  \multicolumn{2}{c}{CA} &
  \multicolumn{2}{c}{CO} &
  \multicolumn{2}{c}{CV} &
  \multicolumn{2}{c}{CS} &
  \multicolumn{2}{c}{MS} &
  \multicolumn{2}{c|}{HA} &
  \multicolumn{2}{c}{UC} &
  \multicolumn{2}{c}{OP} &
  \multicolumn{2}{c|}{FC} &
  \multicolumn{2}{c}{Avg} \\
  &
  \multicolumn{1}{c}{$acc$} &
  \multicolumn{1}{c}{$acc_{1}$} &
  \multicolumn{1}{c}{$acc$} &
  \multicolumn{1}{c}{$acc_{1}$} &
  \multicolumn{1}{c}{$acc$} &
  \multicolumn{1}{c}{$acc_{1}$} &
  \multicolumn{1}{c}{$acc$} &
  \multicolumn{1}{c}{$acc_{1}$} &
  \multicolumn{1}{c}{$acc$} &
  \multicolumn{1}{c}{$acc_{1}$} &
  \multicolumn{1}{c}{$acc$} &
  \multicolumn{1}{c|}{$acc_{1}$} &
  \multicolumn{1}{c}{$acc$} &
  \multicolumn{1}{c}{$acc_{1}$} &
  \multicolumn{1}{c}{$acc$} &
  \multicolumn{1}{c}{$acc_{1}$} &
  \multicolumn{1}{c}{$acc$} &
  \multicolumn{1}{c|}{$acc_{1}$} &
  \multicolumn{1}{c}{$acc$} &
  \multicolumn{1}{c}{$acc_{1}$} \\
  \hline
GPT-3.5 &
  0.201 &
  0.366 &
  0.285 &
  0.518 &
  0.246 &
  0.581 &
  0.339 &
  0.640 &
  0.189 &
  0.525 &
  0.319 &
  0.645 &
  0.215 &
  0.354 &
  0.256 &
  0.619 &
  0.261 &
  0.629 &
  0.257 &
  0.542 \\
GPT-4 &
  0.606 &
  0.681 &
  0.733 &
  0.955 &
  0.841 &
  0.986 &
  0.719 &
  0.934 &
  0.608 &
  0.935 &
  0.860 &
  0.968 &
  0.833 &
  0.988 &
  0.780 &
  0.988 &
  0.878 &
  0.995 &
  0.762 &
  0.937 \\
GLM-4 &
  0.338 &
  0.468 &
  0.653 &
  0.839 &
  0.611 &
  0.933 &
  0.544 &
  0.859 &
  0.523 &
  0.878 &
  0.794 &
  0.949 &
  0.676 &
  0.884 &
  0.605 &
  0.949 &
  0.733 &
  0.975 &
  0.608 &
  0.859 \\
Gemini Pro &
  0.089 &
  0.128 &
  0.171 &
  0.310 &
  0.243 &
  0.386 &
  0.131 &
  0.274 &
  0.201 &
  0.350 &
  0.396 &
  0.594 &
  0.096 &
  0.210 &
  0.271 &
  0.476 &
  0.251 &
  0.420 &
  0.206 &
  0.350 \\ \hline
LLaMA-2-7B &
  0.310 &
  0.675 &
  0.131 &
  0.533 &
  0.195 &
  0.695 &
  0.239 &
  0.698 &
  0.236 &
  0.821 &
  0.234 &
  0.641 &
  0.148 &
  0.540 &
  0.210 &
  0.735 &
  0.141 &
  0.586 &
  0.205 &
  0.658 \\
LLaMA-2-13B &
  0.036 &
  0.265 &
  0.043 &
  0.260 &
  0.088 &
  0.306 &
  0.166 &
  0.299 &
  0.071 &
  0.318 &
  0.131 &
  0.294 &
  0.054 &
  0.234 &
  0.088 &
  0.328 &
  0.046 &
  0.265 &
  0.080 &
  0.285 \\ \hline
Avg &
  0.263 &
  0.430 &
  0.336 &
  0.569 &
  0.371 &
  0.648 &
  0.356 &
  0.617 &
  0.305 &
  0.638 &
  0.456 &
  0.682 &
  0.337 &
  0.535 &
  0.368 &
  0.682 &
  0.385 &
  0.645 &
  0.353 &
  0.605 \\ \hline
\end{tabular}
\end{adjustbox}
}
\caption{Average accuracy of different models in different error types on GSM8K and MathQA under zero-shot prompts. We use the first two letters of the name of error type to represent it. The calculation of the first and second column is similar as the \textit{Avg} in Table \ref{tab:Based on Model, main table}.}
\label{tab:Based on Error Type, main table}
\end{table*}

\subsection{Model Performance (RQ1)}

\textbf{Overall Performance.} Table \ref{tab:Based on Model, main table} presents the average accuracy of each LLM in four tasks on the EIC-Math dataset with GSM8K and MathQA. 
Overall, GPT-4 demonstrates overwhelming superiority, followed by GLM-4. 
GPT-3.5, Gemini Pro, and LLaMA-2-7B have their own strengths and weaknesses in four tasks. 
It is noteworthy that LLaMA-2-7B performs better than LLaMA-2-13B, which may be related to inverse scaling \citep{mckenzie2023inverse}.
This suggests that the ability of models to identify and correct errors does not necessarily increase with model size. 
Moreover, the mathematical models can only provide answers without error identification or correction abilities, and thus their accuracy is low as showcased in Appendix \ref{Comparasion with other math models} and \ref{Detailed results}. This indicates that they can only solve problems and lack comprehensive reasoning abilities. 

\noindent
\textbf{Comparison Across Tasks.} The average accuracy of \textit{EP} ($acc_{1}$) is the highest among the four tasks ($acc_{1}$, $acc_{2}$, $acc_{3}$, $acc_{4}$), as it is the simplest. \textit{ES} ($acc_{2}$)  and \textit{ET} ($acc_{3}$)  tend to have close average accuracy compared to \textit{EC} ($acc_{4}$), despite being intuitively less challenging. Actually, \textit{ES} involves an additional counting process, while \textit{ET} involves additional classification, leading to different emphases.  It can also be noted that the average accuracy $acc_{1}$ fluctuates across the four tasks, which is due to the efforts of LLMs to maintain consistency with different generated contents.

Regarding the difference in two average accuracy ($acc_{1}$, $acc_{4}$) between \textit{EC}, among the models with poor performance, Gemini Pro exhibits the smallest difference, while LLaMA-2-7B shows the largest. This suggests that Gemini Pro is cautious in error identification, with most identified errors being correctable, whereas LLaMA-2-7B is more liberal in error identification rather than correction. 

\noindent
\textbf{Comparison Between Datasets.}
From the perspective of two datasets, it is often observed that the same model on MathQA tends to have lower accuracy across the four tasks compared to GSM8K. This is attributed to the higher difficulty level of MathQA. 

\noindent
\textbf{Future Direction.}
Additionally, despite the overwhelming superiority of GPT-4, its average accuracy across the four tasks on the two simple MWP datasets is only 76.2\%. This indicates that the error identification and correction tasks we design are challenging, and the lack of error identification and correction capability in LLMs somewhat restricts their mathematical reasoning abilities.

\subsection{Error Type Analysis (RQ2)}

\begin{table*}[hbt!]
\centering
\setlength{\abovecaptionskip}{0.1cm}
\setlength{\belowcaptionskip}{-0.05cm}
\tabcolsep=0.15cm
\scalebox{0.67}{
\begin{adjustbox}{center}
\begin{tabular}{ccccccccccccc|cccccc|cc}
\hline
 &
  \multicolumn{2}{c}{CA} &
  \multicolumn{2}{c}{CO} &
  \multicolumn{2}{c}{CV} &
  \multicolumn{2}{c}{CS} &
  \multicolumn{2}{c}{MS} &
  \multicolumn{2}{c|}{HA} &
  \multicolumn{2}{c}{UC} &
  \multicolumn{2}{c}{OP} &
  \multicolumn{2}{c|}{FC} &
  \multicolumn{2}{c}{Avg} \\ 
  &
  \multicolumn{1}{c}{$acc_{i}$} &
  \multicolumn{1}{c}{$acc_{1}$} &
  \multicolumn{1}{c}{$acc_{i}$} &
  \multicolumn{1}{c}{$acc_{1}$} &
  \multicolumn{1}{c}{$acc_{i}$} &
  \multicolumn{1}{c}{$acc_{1}$} &
  \multicolumn{1}{c}{$acc_{i}$} &
  \multicolumn{1}{c}{$acc_{1}$} &
  \multicolumn{1}{c}{$acc_{i}$} &
  \multicolumn{1}{c}{$acc_{1}$} &
  \multicolumn{1}{c}{$acc_{i}$} &
  \multicolumn{1}{c|}{$acc_{1}$} &
  \multicolumn{1}{c}{$acc_{i}$} &
  \multicolumn{1}{c}{$acc_{1}$} &
  \multicolumn{1}{c}{$acc_{i}$} &
  \multicolumn{1}{c}{$acc_{1}$} &
  \multicolumn{1}{c}{$acc_{i}$} &
  \multicolumn{1}{c|}{$acc_{1}$} &
  \multicolumn{1}{c}{$acc_{i}$} &
  \multicolumn{1}{c}{$acc_{1}$} \\
  \hline
EP &
  \multicolumn{1}{c}{-} &
  \multicolumn{1}{c}{0.350} &
  \multicolumn{1}{c}{-} &
  \multicolumn{1}{c}{0.482} &
  \multicolumn{1}{c}{-} &
  \multicolumn{1}{c}{0.575} &
  \multicolumn{1}{c}{-} &
  \multicolumn{1}{c}{0.552} &
  \multicolumn{1}{c}{-} &
  \multicolumn{1}{c}{0.557} &
  \multicolumn{1}{c}{-} &
  \multicolumn{1}{c|}{0.609} &
  \multicolumn{1}{c}{-} &
  \multicolumn{1}{c}{0.428} &
  \multicolumn{1}{c}{-} &
  \multicolumn{1}{c}{0.648} &
  \multicolumn{1}{c}{-} &
  \multicolumn{1}{c|}{0.584} &
  \multicolumn{1}{c}{-} &
  \multicolumn{1}{c}{0.532} \\
ES  & 0.203 & 0.476 & 0.323 & 0.589 & 0.362 & 0.697 & 0.367 & 0.667 & 0.320 & 0.683 & 0.408 & 0.697 & 0.323 & 0.583 & 0.383 & 0.699 & 0.343 & 0.652 & 0.337 & 0.638 \\
ET  & 0.312 & 0.541 & 0.204 & 0.682 & 0.177 & 0.751 & 0.163 & 0.713 & 0.029 & 0.763 & 0.433 & 0.817 & 0.298 & 0.655 & 0.082 & 0.785 & 0.241 & 0.761 & 0.215 & 0.719 \\
EC  & 0.188 & 0.355 & 0.335 & 0.523 & 0.369 & 0.569 & 0.344 & 0.537 & 0.313 & 0.549 & 0.372 & 0.604 & 0.298 & 0.473 & 0.361 & 0.598 & 0.373 & 0.583 & 0.328 & 0.532 \\ \hline
Avg & 0.263 & 0.430 & 0.336 & 0.569 & 0.371 & 0.648 & 0.356 & 0.617 & 0.305 & 0.638 & 0.337 & 0.535 & 0.368 & 0.682 & 0.385 & 0.645 & 0.456 & 0.682 & 0.353 & 0.605 \\ \hline
\end{tabular}
\end{adjustbox}
}
\caption{Average accuracy of different tasks in different error types on GSM8K and MathQA under zero-shot prompts. And we calculate the average $acc_{1}$ over all models for \textit{EP}, the average $acc_{i} (i=2,3,4)$ and $acc_{1}$ as the values for the first and second columns respectively for \textit{ES}, \textit{ET} and \textit{EC}.}
\label{tab:Based on Task, main table}
\end{table*}

\textbf{Difficulty Levels of Error Types.}
In Table \ref{tab:Based on Error Type, main table}, we compute the average accuracy of each model across the four tasks in each error type on two datasets to assess the difficulty levels of different error types. It is found that calculation error is the most challenging to identify and correct, with an average accuracy of only 26.3\%, while hallucination is the easiest, with an average accuracy of 45.6\%. It is noteworthy that although GPT-4 and GLM-4 perform well overall, their performance in identifying and correcting calculation error is significantly lower compared to other error types. This suggests that LLMs should focus more on developing their computational capability. In addition, difficulty in identifying missing step is attributed to its poorest performance in the \textit{ET} of 2.9\% shown in Table \ref{tab:Based on Task, main table}, making it the most challenging type for LLMs to classify. This is because it requires traversing the entire solution's CoT to analyze whether essential reasoning steps are missing.

\noindent
\textbf{Comparison between Different Models on the Same Error Type.}
Furthermore, GPT-3.5 and Gemini Pro struggle with unit conversion error, and the LLaMA-2 series also perform poorly in unit conversion error and formula confusion error. At the same time, GPT-4 and GLM-4 perform well in unit conversion error and formula confusing error. We speculate that this may be related to the size of the stored parameter knowledge. Due to the lack of relevant common sense in the parameter knowledge, it becomes challenging to identify and correct related errors for smaller models.

The average accuracy of LLaMA-2-7B surprisingly reaches 31\% in calculation error, on par with GLM-4. Compared to other error types, LLaMA-2-7B and LLaMA-2-13B excell in contradictory step but perform poorly in counting error.
\begin{table}[tbp]
\centering
\adjustbox{center}{
  \begin{minipage}[t]{\columnwidth}
    \centering
    \scalebox{0.78}{
    \begin{tabular}{l|clclclclclclclclcl}
\hline
 &
  \multicolumn{2}{c}{CA} &
  \multicolumn{2}{c}{CO} &
  \multicolumn{2}{c}{CV} &
  \multicolumn{2}{c}{CS} &
  \multicolumn{2}{c}{MS} &
  \multicolumn{2}{c}{HA} &
  \multicolumn{2}{c}{UC} &
  \multicolumn{2}{c}{OP} &
  \multicolumn{2}{c}{FC} \\ \hline
CA &
  \multicolumn{2}{c}{\cellcolor[HTML]{340096}{\color[HTML]{FFFFFF} 119}} &
  \multicolumn{2}{c}{\cellcolor[HTML]{3303CF}{\color[HTML]{FFFFFF} 30}} &
  \multicolumn{2}{c}{\cellcolor[HTML]{6665CD}15} &
  \multicolumn{2}{c}{\cellcolor[HTML]{3303CF}{\color[HTML]{FFFFFF} 32}} &
  \multicolumn{2}{c}{\cellcolor[HTML]{CBCEFB}2} &
  \multicolumn{2}{c}{\cellcolor[HTML]{CBCEFB}3} &
  \multicolumn{2}{c}{\cellcolor[HTML]{9698ED}10} &
  \multicolumn{2}{c}{\cellcolor[HTML]{9698ED}11} &
  \multicolumn{2}{c}{\cellcolor[HTML]{CBCEFB}2} \\
CO &
  \multicolumn{2}{c}{\cellcolor[HTML]{6200C9}{\color[HTML]{FFFFFF} 56}} &
  \multicolumn{2}{c}{\cellcolor[HTML]{340096}{\color[HTML]{FFFFFF} 108}} &
  \multicolumn{2}{c}{\cellcolor[HTML]{6200C9}{\color[HTML]{FFFFFF} 55}} &
  \multicolumn{2}{c}{\cellcolor[HTML]{3303CF}{\color[HTML]{FFFFFF} 33}} &
  \multicolumn{2}{c}{\cellcolor[HTML]{CBCEFB}1} &
  \multicolumn{2}{c}{\cellcolor[HTML]{CBCEFB}2} &
  \multicolumn{2}{c}{\cellcolor[HTML]{9698ED}5} &
  \multicolumn{2}{c}{\cellcolor[HTML]{6665CD}9} &
  \multicolumn{2}{c}{0} \\
CV &
  \multicolumn{2}{c}{\cellcolor[HTML]{340096}{\color[HTML]{FFFFFF} 109}} &
  \multicolumn{2}{c}{\cellcolor[HTML]{3303CF}{\color[HTML]{FFFFFF} 35}} &
  \multicolumn{2}{c}{\cellcolor[HTML]{303498}{\color[HTML]{FFFFFF} 73}} &
  \multicolumn{2}{c}{\cellcolor[HTML]{6200C9}{\color[HTML]{FFFFFF} 57}} &
  \multicolumn{2}{c}{\cellcolor[HTML]{CBCEFB}1} &
  \multicolumn{2}{c}{\cellcolor[HTML]{6665CD}15} &
  \multicolumn{2}{c}{\cellcolor[HTML]{3303CF}{\color[HTML]{FFFFFF} 40}} &
  \multicolumn{2}{c}{\cellcolor[HTML]{6665CD}23} &
  \multicolumn{2}{c}{\cellcolor[HTML]{9698ED}7} \\
CS &
  \multicolumn{2}{c}{\cellcolor[HTML]{340096}{\color[HTML]{FFFFFF} 161}} &
  \multicolumn{2}{c}{\cellcolor[HTML]{3303CF}{\color[HTML]{FFFFFF} 37}} &
  \multicolumn{2}{c}{\cellcolor[HTML]{6200C9}{\color[HTML]{FFFFFF} 74}} &
  \multicolumn{2}{c}{\cellcolor[HTML]{303498}{\color[HTML]{FFFFFF} 98}} &
  \multicolumn{2}{c}{\cellcolor[HTML]{CBCEFB}1} &
  \multicolumn{2}{c}{\cellcolor[HTML]{6665CD}28} &
  \multicolumn{2}{c}{\cellcolor[HTML]{9698ED}16} &
  \multicolumn{2}{c}{\cellcolor[HTML]{3303CF}{\color[HTML]{FFFFFF} 46}} &
  \multicolumn{2}{c}{0} \\
MS &
  \multicolumn{2}{c}{\cellcolor[HTML]{340096}{\color[HTML]{FFFFFF} 132}} &
  \multicolumn{2}{c}{\cellcolor[HTML]{3303CF}{\color[HTML]{FFFFFF} 24}} &
  \multicolumn{2}{c}{\cellcolor[HTML]{303498}{\color[HTML]{FFFFFF} 71}} &
  \multicolumn{2}{c}{\cellcolor[HTML]{6200C9}{\color[HTML]{FFFFFF} 59}} &
  \multicolumn{2}{c}{\cellcolor[HTML]{CBCEFB}4} &
  \multicolumn{2}{c}{\cellcolor[HTML]{6665CD}14} &
  \multicolumn{2}{c}{\cellcolor[HTML]{6665CD}15} &
  \multicolumn{2}{c}{\cellcolor[HTML]{3303CF}{\color[HTML]{FFFFFF} 22}} &
  \multicolumn{2}{c}{\cellcolor[HTML]{9698ED}7} \\
HA &
  \multicolumn{2}{c}{\cellcolor[HTML]{6200C9}{\color[HTML]{FFFFFF} 32}} &
  \multicolumn{2}{c}{\cellcolor[HTML]{6665CD}15} &
  \multicolumn{2}{c}{\cellcolor[HTML]{303498}{\color[HTML]{FFFFFF} 71}} &
  \multicolumn{2}{c}{\cellcolor[HTML]{3303CF}{\color[HTML]{FFFFFF} 28}} &
  \multicolumn{2}{c}{0} &
  \multicolumn{2}{c}{\cellcolor[HTML]{340096}{\color[HTML]{FFFFFF} 358}} &
  \multicolumn{2}{c}{\cellcolor[HTML]{9698ED}4} &
  \multicolumn{2}{c}{\cellcolor[HTML]{CBCEFB}1} &
  \multicolumn{2}{c}{0} \\
UC &
  \multicolumn{2}{c}{\cellcolor[HTML]{6200C9}{\color[HTML]{FFFFFF} 46}} &
  \multicolumn{2}{c}{\cellcolor[HTML]{3303CF}{\color[HTML]{FFFFFF} 26}} &
  \multicolumn{2}{c}{\cellcolor[HTML]{6665CD}11} &
  \multicolumn{2}{c}{\cellcolor[HTML]{9698ED}6} &
  \multicolumn{2}{c}{0} &
  \multicolumn{2}{c}{0} &
  \multicolumn{2}{c}{\cellcolor[HTML]{340096}{\color[HTML]{FFFFFF} 262}} &
  \multicolumn{2}{c}{\cellcolor[HTML]{9698ED}7} &
  \multicolumn{2}{c}{\cellcolor[HTML]{CBCEFB}1} \\
OP &
  \multicolumn{2}{c}{\cellcolor[HTML]{340096}{\color[HTML]{FFFFFF} 145}} &
  \multicolumn{2}{c}{\cellcolor[HTML]{3303CF}{\color[HTML]{FFFFFF} 36}} &
  \multicolumn{2}{c}{\cellcolor[HTML]{6200C9}{\color[HTML]{FFFFFF} 48}} &
  \multicolumn{2}{c}{\cellcolor[HTML]{3303CF}{\color[HTML]{FFFFFF} 39}} &
  \multicolumn{2}{c}{0} &
  \multicolumn{2}{c}{\cellcolor[HTML]{9698ED}17} &
  \multicolumn{2}{c}{\cellcolor[HTML]{303498}{\color[HTML]{FFFFFF} 73}} &
  \multicolumn{2}{c}{\cellcolor[HTML]{6665CD}31} &
  \multicolumn{2}{c}{\cellcolor[HTML]{CBCEFB}3} \\
FC &
  \multicolumn{2}{c}{\cellcolor[HTML]{303498}{\color[HTML]{FFFFFF} 90}} &
  \multicolumn{2}{c}{\cellcolor[HTML]{9698ED}12} &
  \multicolumn{2}{c}{\cellcolor[HTML]{9698ED}16} &
  \multicolumn{2}{c}{\cellcolor[HTML]{3303CF}{\color[HTML]{FFFFFF} 36}} &
  \multicolumn{2}{c}{\cellcolor[HTML]{CBCEFB}1} &
  \multicolumn{2}{c}{\cellcolor[HTML]{6665CD}20} &
  \multicolumn{2}{c}{\cellcolor[HTML]{3303CF}{\color[HTML]{FFFFFF} 44}} &
  \multicolumn{2}{c}{\cellcolor[HTML]{6200C9}{\color[HTML]{FFFFFF} 69}} &
  \multicolumn{2}{c}{\cellcolor[HTML]{340096}{\color[HTML]{FFFFFF} 149}} \\ \hline
\end{tabular}
    }
    \caption{
    Counting statistics for error type classification of GSM8K on GPT-3.5 with varied prompts.
    Row and column headers denote the golden and the classified types, respectively. 
    Darker color indicates larger counts. 
    }
    \vspace{-5pt}
    \label{tab:GSM8K GPT-3.5 error type analysis1}
    
  \end{minipage}%
}
\end{table}

\noindent
\textbf{Statistical Classification of Error Types.}
Table \ref{tab:GSM8K GPT-3.5 error type analysis1} provides statistic on the count of error types classified on GPT-3.5 with GSM8K. Similar statistics for most other models and datasets are presented in Appendix \ref{Error Type Analysis}. It can be observed that most of the error types are often misclassified as calculation error, which may be attributed to the models' lack of true understanding of the meanings of each error type and relevant classification training data.

\subsection{Prompt Robustness (RQ3)}
We devise a variety of prompts for the four tasks to explore the robustness of different models to different prompts. In addition, we investigate whether providing the error types to models can improve the accuracy in \textit{ET} and \textit{EC}.
\begin{table}[t]
\centering
\setlength{\abovecaptionskip}{0.1cm}
\setlength{\belowcaptionskip}{-0.05cm}
\resizebox{0.5\textwidth}{!}{
\begin{tabular}{cccc|ccc}
\toprule
\textbf{}   & \multicolumn{3}{c|}{GSM8K}   & \multicolumn{3}{c}{MathQA}   \\
            & Simple & Normal & Misleading & Simple & Normal & Misleading \\ \hline
GPT-3.5     & 0.705  & 0.759  & 0.543      & 0.698 & 0.680  & 0.621      \\
GPT-4       & 0.672  & 0.875  & 0.805      & 0.555  & 0.741  & 0.713      \\
GLM-4       & 0.854  & 0.795  & 0.750      & 0.808  & 0.722  & 0.678      \\
Gemini Pro  & 0.701  & 0.705  & 0.740      & 0.703  & 0.718  &0.752      \\ 
\midrule
LLaMA-2-7B  & 0.667  & 0.445  & -          & 0.667  & 0.705  & 0.113      \\
LLaMA-2-13B & 0.667  & 0.691  & -          & 0.649  & 0.658  & 0.099      \\ \bottomrule
\end{tabular}
}
\caption{F1 scores on \textit{EP} under three prompt settings.}
\label{tab:EP prompt robustness testing, F1-score}
\vspace{-5pt}
\end{table}

\begin{table*}[ht]
\renewcommand{\arraystretch}{0.8}
\centering
\setlength{\abovecaptionskip}{0.1cm}
\setlength{\belowcaptionskip}{-0.05cm}
\tabcolsep=0.15cm
\scalebox{0.8}{
\begin{adjustbox}{center}
\begin{tabular}{ccccc|cccc}
\toprule
\textbf{}   & \multicolumn{4}{c|}{GSM8K}                               & \multicolumn{4}{c}{MathQA}                      \\
 & Zero-shot & Few-shot & \multicolumn{1}{l}{Zero-shot-type} & Few-shot-type & Zero-shot & Few-shot & \multicolumn{1}{l}{Zero-shot-type} & Few-shot-type \\ \midrule
GPT-3.5     & 0.147          & 0.198 & 0.294          & \textbf{0.352} & 0.173 & 0.157 & 0.248          & \textbf{0.304} \\
GPT-4       & 0.843          & 0.841 & 0.878          & \textbf{0.881} & 0.714 & 0.691 & \textbf{0.739} & \textbf{0.739} \\
GLM-4       & 0.640          & 0.632 & \textbf{0.744} & 0.689          & 0.551 & 0.496 & \textbf{0.603} & 0.581          \\
Gemini Pro  & 0.359          & 0.052 & \textbf{0.567} & 0.112          & 0.239 & 0.031 & \textbf{0.394} & 0.086          \\ \midrule
LLaMA-2-7B  & 0.184          & 0.109 & \textbf{0.209} & 0.094          & 0.176 & 0.133 & \textbf{0.197} & 0.120          \\
LLaMA-2-13B & \textbf{0.007} & 0.003 & 0.002          & 0.004          & 0.009 & 0.003 & 0.004          & \textbf{0.017} \\ \bottomrule
\end{tabular}
\end{adjustbox}
}
\caption{Average accuracy $acc_{2}$ of models in \textit{ES} on GSM8K and MathQA separately under four different prompt settings. Zero-shot-type and Few-shot-type provide models with the error types. Few-shot is set to 2-shot. The maximum average accuracy for each model on each dataset is in \textbf{boldface}.}
\label{tab:ES prompt robustness testing, Accuracy}
\end{table*}

\begin{table*}[!h]
\renewcommand{\arraystretch}{0.8}
\centering
\setlength{\abovecaptionskip}{0.1cm}
\setlength{\belowcaptionskip}{-0.050cm}
\tabcolsep=0.1cm
\scalebox{0.75}{
\begin{adjustbox}{center}
\begin{tabular}{ccccc|cccc}
\toprule
\textbf{}   & \multicolumn{4}{c|}{GSM8K}                               & \multicolumn{4}{c}{MathQA}                               \\
 & Zero-shot & Few-shot & \multicolumn{1}{l}{Zero-shot-reverse} & Zero-shot-random & Zero-shot & Few-shot & \multicolumn{1}{l}{Zero-shot-reverse} & Zero-shot-random \\ \midrule
GPT-3.5     & 0.211 & 0.171          & 0.281 & 0.256          & 0.173 & 0.129          & 0.228 & 0.204          \\
GPT-4       & 0.516 & 0.577 & 0.538          & 0.483          & 0.481 & 0.520 & 0.471          & 0.443          \\
GLM-4       & 0.349 & 0.409          & 0.411 & 0.381          & 0.327 & 0.218          & 0.360 & 0.353          \\
Gemini Pro  & 0.108 & 0.090          & 0.147 & 0.122          & 0.096 & 0.052          & 0.132 & 0.113          \\ \midrule
LLaMA-2-7B  & 0.048 & 0.076          & 0.097 & 0.081          & 0.052 & 0.104          & 0.121 & 0.082          \\
LLaMA-2-13B & 0.127 & 0.003          & 0.112          & 0.136 & 0.116 & 0.017          & 0.127          & 0.133 \\ \bottomrule
\end{tabular}
\end{adjustbox}
}
\caption{Average accuracy $acc_{3}$ of models in \textit{ET} on GSM8K and MathQA separately under four different prompt settings. Few-shot is set to 2-shot. Few-shot-random and Few-shot-reverse present similar results and are included in Appendix \ref{Prompt Robustness Analysis}.}
\label{tab:ET prompt robustness testing, Accuracy}
\end{table*}

\begin{table*}[!h]
\renewcommand{\arraystretch}{0.8}
\centering
\setlength{\abovecaptionskip}{0.1cm}
\setlength{\belowcaptionskip}{0cm}
\tabcolsep=0.15cm
\scalebox{0.8}{
\begin{adjustbox}{center}
\begin{tabular}{ccccc|cccc}
\toprule
\textbf{}   & \multicolumn{4}{c|}{GSM8K}                      & \multicolumn{4}{c}{MathQA}                               \\
 & Zero-shot & Few-shot & \multicolumn{1}{l}{Zero-shot-type} & Few-shot-type & Zero-shot & Few-shot & \multicolumn{1}{l}{Zero-shot-type} & Few-shot-type \\ \midrule
GPT-3.5     & 0.296 & 0.169 & 0.477          & \textbf{0.594} & 0.274 & 0.141          & 0.402          & \textbf{0.572} \\
GPT-4       & 0.901 & 0.883 & 0.922          & \textbf{0.929} & 0.834 & 0.810          & 0.847          & \textbf{0.874} \\
GLM-4       & 0.853 & 0.804 & 0.912          & \textbf{0.937} & 0.692 & 0.574          & 0.694          & \textbf{0.752} \\
Gemini Pro  & 0.117 & 0.248 & \textbf{0.844} & 0.283          & 0.082 & 0.200          & \textbf{0.680} & 0.186          \\ \midrule
LLaMA-2-7B  & 0.067 & 0.066 & \textbf{0.071} & 0.050          & 0.039 & \textbf{0.063} & 0.041          & 0.048          \\
LLaMA-2-13B & 0.000 & 0.006 & 0.000          & \textbf{0.010} & 0.000 & 0.018          & 0.000          & \textbf{0.019} \\ \bottomrule
\end{tabular}
\end{adjustbox}
}
\caption{Average accuracy $acc_{4}$ of models in \textit{EC} on GSM8K and MathQA separately under four different prompt settings. Zero-shot-type and Few-shot-type provide models with the error types. Few-shot is set to 2-shot. The maximum average accuracy for each model on each dataset is in \textbf{boldface}.}
\label{tab:EC prompt robustness testing, Accuracy}
\vspace{-10pt}
\end{table*}

\noindent
\textbf{Prompt Robustness of EP.}
For \textit{EP}, we select 50 negative samples and add an equal number of positive samples for each error type, totaling 100 samples for testing. And in Table \ref{tab:EP prompt robustness testing, F1-score}, we compute their average F1 scores under three different prompts: \textit{Simple}, \textit{Normal} and \textit{Misleading}.
By calculating the difference in average F1 scores across all error types for each model, we evaluate their robustness to different prompts. It is observed that closed-source models exhibit greater robustness to different prompts, with the maximum difference in average F1 scores around 0.2. In contrast, open-source models are highly sensitive to different prompts, exhibiting a tendency to classify almost all cases as correct without much consideration under \textit{Simple} and being misled to mostly classify cases as incorrect under \textit{Misleading}.

\noindent
\textbf{Prompt Robustness of ES.}
For \textit{ES}, we design zero-shot and few-shot prompts for comparison and find that increasing the shot has minimal effect on improving the accuracy of this task and could even be counterproductive in Table \ref{tab:ES prompt robustness testing, Accuracy}. This indicates that simple examples can not make models fully understand the meaning of identifying the first erroneous step. By providing models with the error types, the accuracy of identifying error steps has been significantly improved, with an average increase of 45.9\% times and maximum increase of 12.71 times. This informs that carefully designed examples can effectively improve the models' ability to identify erroneous steps.

\noindent
\textbf{Prompt Robustness of ET.}
For \textit{ET}, we define nine error types in the prompts and design zero-shot and few-shot prompts. Recognizing that the sequence of error types may impact the accuracy of identifying errors, we also devise prompts that reverse the default order of error types and randomly shuffle them. In Table \ref{tab:ET prompt robustness testing, Accuracy}, the impact of increasing the shot on improving accuracy is also negligible by comparing zero-shot and few-shot prompts. The order of error types does indeed affect classification accuracy as shown in Table \ref{tab:GSM8K ET prompt robustness testing，Accuracy} and \ref{tab:MathQA ET prompt robustness testing，Accuracy}. For example, hallucination is listed last in the sequential prompt. The average classification accuracy of hallucination in the sequential prompt is much lower than that in the reversed order. It is noteworthy that in the random order, we place missing step first, but its classification accuracy remains consistently low, indicating its inherent difficulty in identification.

\noindent
\textbf{Prompt Robustness of EC.}
For \textit{EC}, we adopt similar prompt settings with \textit{ES} and obtain similar results. Only delicately constructed prompts that provide the error types can effectively improve the models' ability to correct errors,  with an average increase of 47.9\% times and up to a maximum of 8.29 times as displayed in Table \ref{tab:EC prompt robustness testing, Accuracy}.

\subsection{In-depth Analysis}\label{In-depth Analysis}

\begin{figure}[t] 
\setlength{\abovecaptionskip}{0.1cm}
\setlength{\belowcaptionskip}{0cm}
	\centering
	\includegraphics[width=0.48\textwidth]{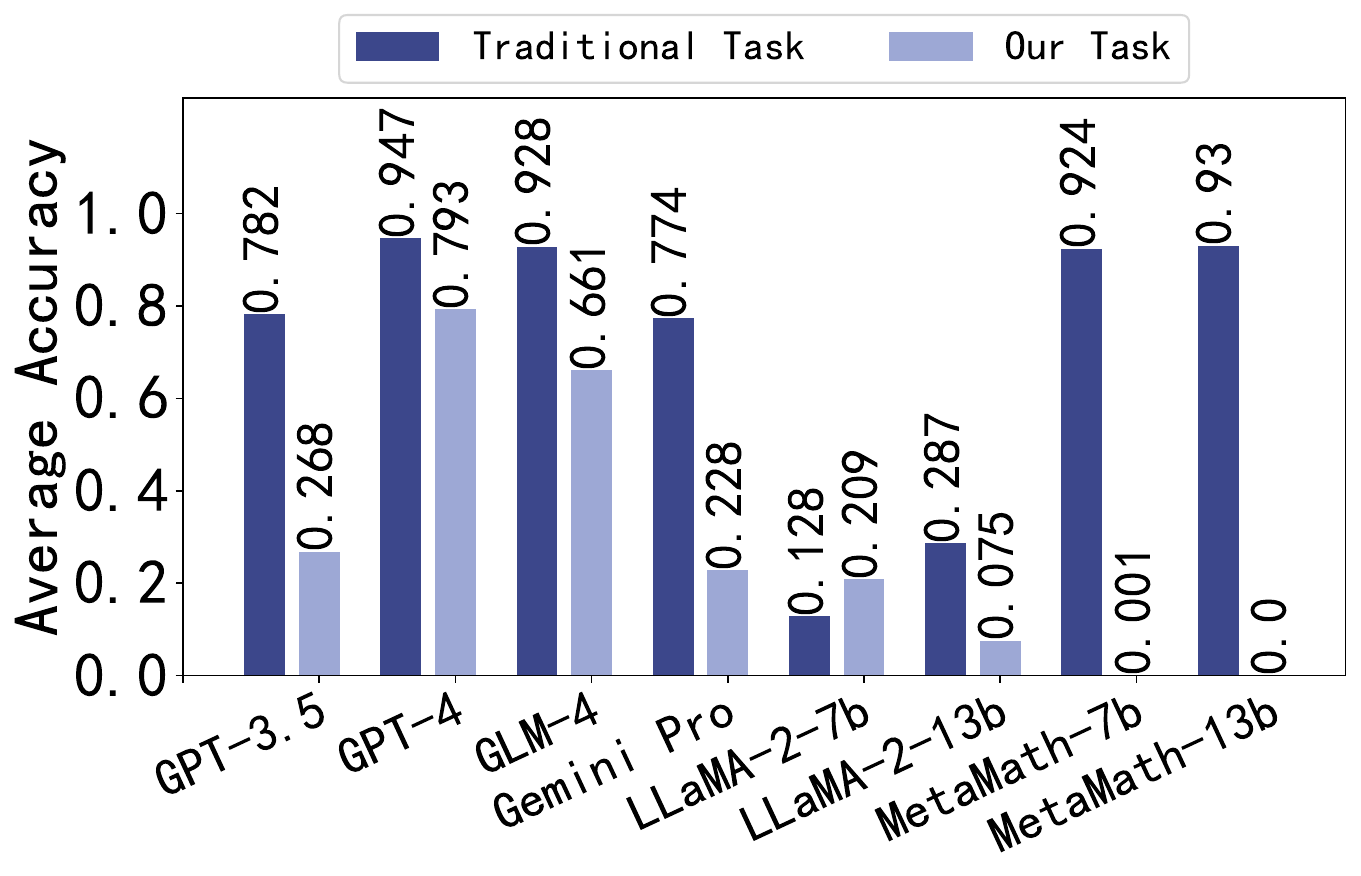}
	\caption{Accuracy of traditional task and our task on GSM8K.}
	\label{fig:Evaluation Results of Traditional Task and Our Task on GSM8K}
 \vspace{-5pt}
\end{figure}

\noindent
\textbf{Comparison with Traditional Task.} We conduct traditional task by inputting the questions from our dataset into LLMs and obtaining the solutions and answers as outputs. The average accuracy of traditional task and our task is showcased in Figure \ref{fig:Evaluation Results of Traditional Task and Our Task on GSM8K}. And details are in Appendix \ref{Detailed Comparison with Traditional Task}. It can be observed that closed-source models perform well on both datasets in traditional task, while among the open-source models, only MetaMath series achieve high accuracy on GSM8K, possibly due to overfitting. It is worth noting that the ability of LLaMA-2-7B to identify and correct errors is greater than its problem-solving ability. However, the accuracy of traditional task is overall higher than that of our proposed task, which indicates the significance of our evaluation task in improving LLMs' mathematical reasoning abilities.

\begin{figure}[t]   
\setlength{\abovecaptionskip}{0.1cm}
\setlength{\belowcaptionskip}{0cm}
	\centering
	\includegraphics[width=0.48\textwidth]{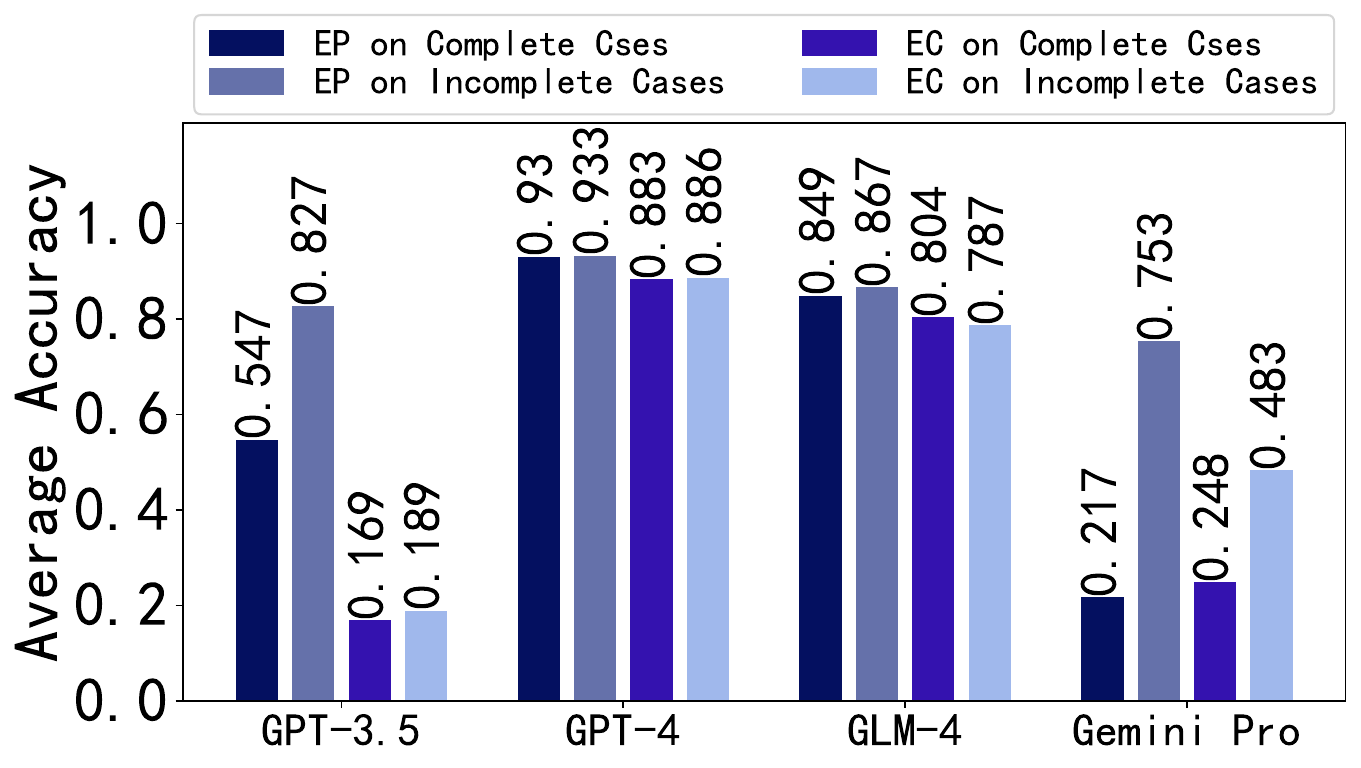}
	\caption{Accuracy of incomplete cases and complete cases on GSM8K for closed-source models.}
	\label{fig:Evaluation Results of Incomplete Cases for GSM8K on Closed-source Models}
 \vspace{-5pt}
\end{figure}

\noindent
\textbf{Influence of Stopping at Error Step.}
We investigate the comparison between writing only up to the error step in the solution and continuing from the error step to complete the solution. It can be observed from Figure \ref{fig:Evaluation Results of Incomplete Cases for GSM8K on Closed-source Models} that, for both \textit{EP} and \textit{EC}, stopping at the error step aids in error identification and correction. Continuing from the error step to complete the solution may confuse LLMs. More details can be found in Appendix \ref{Detailed Influence of Stopping at Error Step}.

\section{Related Work}
\noindent %
\textbf{Mathematical Reasoning Evaluation.} Ever since the exceptional capabilities of LLMs have been applied to the field of mathematical reasoning, researchers have initiated assessments of their mathematical proficiency.
Most of them have primarily focused on evaluating the correctness in solving mathematical problems based on whether the answers are accurate(\citealp{shakarian2023independent}; \citealp{fu2023chain}; \citealp{hong2024stuck}; \citealp{shi2022language}; \citealp{dahlgren2022clevr}; \citealp{frieder2023mathematical}). The correctness and consistency of intermediate steps in the solutions are also commonly used as an evaluation criterion to assess the coherence of Chain of Thought (CoT) 
(\citealp{wei2022chain}; \citealp{golovneva2022roscoe}; \citealp{zhang2023evaluating}; \citealp{gaur2023reasoning}). Others employ human interactions to provide a dynamic evaluation of the answers and intermediate steps (\citealp{zhuang2023efficiently}; \citealp{collins2023evaluating}). However, little work has investigated LLMs' ability to identify and correct errors, or only from a macro perspective and conduct simple experiments (\citealp{liu2023novice}; \citealp{yen2023three}; \citealp{valmeekam2023can}; \citealp{stechly2023gpt}; \citealp{an2023learning}; \citealp{huang2023large}). Hence, there lacks a fine-grained study that comprehensively evaluates the LLMs' abilities in error identification and correction.

\vspace{5pt}
\noindent %
\textbf{In-Context Learning.}
With the widespread adoption of LLMs (\citealp{brown2020language}; \citealp{ouyang2022training}; \citealp{anil2023palm}; \citealp{openai2023gpt}), in-context learning (\citealp{brown2020language}; \citealp{ouyang2022training}; \citealp{min-etal-2022-rethinking}) has emerged as the predominant method for deploying downstream tasks. This approach involves providing LLMs with textual instructions and examples without the need for parameter updates. When applied to dataset generation, studies have found that datasets generated by LLMs exhibit higher quality in terms of accuracy and fluency (\citealp{lu2022fantastically}; \citealp{min-etal-2022-rethinking}) compared to datasets annotated by crowd-sourced workers. Furthermore, the cost of generating data through LLMs is significantly lower than the expense associated with crowd-sourced annotations. Consequently, employing LLMs for data generation proves to be a viable alternative to crowd-sourced annotation (\citealp{liu2022wanli}; \citealp{wiegreffe2022reframing}; \citealp{west2022symbolic}). Therefore, we opt to utilize in-context learning on the state-of-the-art GPT-4 to generate the evaluation dataset.

\vspace{5pt}
\noindent %
\textbf{Program Repair.}
Automated program repair (APR), aimed at fixing potential errors within programs, plays a crucial role in the software development cycle. Early approaches (\citealp{nguyen2013semfix}; \citealp{qi2014strength}; \citealp{diekmann2018don}) were symbolic and often relied on error recovery mechanisms within parsers to enumerate local edits. More recently, neural networks have been successfully used to correct syntax and compilation errors (\citealp{yasunaga2020graph}; \citealp{yasunaga2021break}; \citealp{ahmed2021synfix}; \citealp{berabi2021tfix}). Besides, some systems also integrate symbolic and neural components to rectify faulty programs (\citealp{bavishi2022neurosymbolic}). Due to the remarkable capabilities of LLMs, they are utilized in program repair to detect, locate and rectify errors, enabling automated software development workflows (\citealp{joshi2023repair}; \citealp{jin2023inferfix}; \citealp{bouzenia2024repairagent}). These efforts differ from ours in that they focus on identifying and correcting errors within code scenarios, while we concentrate on mathematical reasoning problems.

\section{Conclusion}
We systematically delineated four evaluation tasks aimed at identifying and rectifying errors, marking the first comprehensive attempt in this domain. To facilitate the evaluation process, we curated a dataset categorized by error types. Furthermore, we conducted thorough experiments across various closed-source and open-source models, yielding significant insights that bolster the mathematical reasoning capabilities of LLMs.
In future research, we will explore more avenues such as rectifying single-step and single-type errors, single-step multi-type errors on various LLMs, and increasing the continuity of our correction prompts which can rectify errors based on incorrect preceding steps. 


\section*{Limitations}

In future research, we can focus on the following directions. First, we mainly investigated the capability of different LLMs in identifying and rectifying single-step and single-type errors, and future research can address combined errors involving single-step and multiple-type, as well as single-type and multiple-step, and more complex errors such as semantic comprehension error.
Futhermore, our correction prompts did not emphasize continuity, whose meaning is to correct on the basis of incorrect steps. And correction based on continuity may indeed pose a greater challenge.
Lastly, our discussion focused solely on machine performance regarding these error types, and we will explore if there are differences between humans and machines in identifying and rectifying errors in future.

\section*{Ethics Statement}

One ethical concern revolves around the accuracy and reliability of LLMs in recognizing and correcting errors in mathematical reasoning. Errors in mathematical reasoning can have profound consequences, particularly in educational contexts where students rely on accurate feedback and guidance to develop their mathematical skills. Therefore, ensuring the robustness and integrity of LLMs' error correction capabilities through rigorous validation and continuous improvement processes is essential to mitigate the risks associated with erroneous corrections. Moreover, ethical responsibilities extend to the broader societal impacts of LLMs' role in mathematical education and problem-solving. As LLMs increasingly assist students and professionals in mathematical reasoning tasks, the dissemination of accurate and credible mathematical knowledge becomes paramount. Ensuring that LLMs are equipped to discern and rectify errors in mathematical reasoning contributes to fostering a culture of mathematical integrity, critical thinking, and intellectual hones. Lastly, we will check there are no ethical issues in the constructed dataset before releasing it publicly, which can be used for research purposes by related researchers. 

\bibliography{custom}

\appendix

\section{Dataset Selection}
\label{sec:appendix dataset}
The datasets commonly used for MWP assessment include GSM8K \citep{cobbe2021training}, MathQA \citep{amini2019mathqa}, MAWPS \citep{koncel2016mawps}, SVAMP \citep{patel2021nlp}, and MATH23K \citep{wang2017deep}. GSM8K corresponds to elementary-level mathematical problems. MathQA comprises GRE-level mathematical questions, which is served as a benchmark for American college entrance exams. MAWPS is akin to the fourth-grade level. SVAMP focuses on univariate linear problems, where all questions can be solved using a single expression. And MATH23K is a large-scale Chinese dataset containing Chinese mathematical word problems and their corresponding expression solutions.
Considering various factors, we select GSM8K and MathQA as our primary datasets. Due to MathQA's multiple-choice format and considerable noise in its original annotations, we employed GPT-3.5 to generate correct solutions for its questions. For GSM8K, we utilize its annotated solutions.

\section{Detailed Error Type Definition}\label{Detailed Error Type Definition}
A substantial collection of erroneous instances is gathered from existing studies (\citealp{wei2022chain}; \citealp{toh2023veritymath}; \citealp{lightman2023lets}; \citealp{shakarian2023independent}; \citealp{bubeck2023sparks}; \citealp{sawada2023arb}; \citealp{suzgun2022challenging}; \citealp{lyu2023faithful}; \citealp{kojima2022large}; \citealp{li2023making}; \citealp{wang2022towards}; \citealp{wang2023plan}; \citealp{paul2023refiner}; \citealp{golovneva2022roscoe}; \citealp{ribeiro2023street}; \citealp{lewkowycz2022solving}) and practical scenarios. Subsequently, nine common and distinct error types are distilled, focusing on the single-step errors and cross-steps errors. The first seven types pertain to single-step errors, while the latter two relate to cross-steps errors.

\textbf{Calculation Error:} Error appears during the calculation process when the formula is entirely correct. It is well-known that LLMs often exhibit inconsistent computation units, resulting in simple arithmetic errors \citep{toh2023veritymath}.

\textbf{Counting Error:} Error occurs during the counting process. \citealp{bubeck2023sparks} indicates that counting error is prone due to not only the challenging implementation of this operation within transformer structures but also the lack of relevant data in the training sets.

\textbf{Context Value Error:} Error arises when attributes of named entities (such as quantities) do not align with the information provided in the question. The tendency of LLMs to misinterpret problem meanings and erroneously substitute numerical values remains a prominent challenge in mathematics reasoning \citep{yen2023three}.

\textbf{Hallucination:} Error involves adding fictitious unrelated statements contradictory to the question. This refers to the inclusion of information in the solution that is not present in the question statement, thereby disrupting the final answer \citep{lyu2023faithful}.

\textbf{Unit Conversion Error:} Error occurs during unit conversion process, indicating a misunderstanding of the quantitative relationships between units \citep{choi2023assessment}.

\textbf{Operator Error:} Error involves a single operator being erroneously applied within the expression due to a misconception of operator concepts \citep{paul2023refiner}.

\textbf{Formula Confusion Error:} Error appears when applying formula in inappropriate scenario. This stems from a misunderstanding of formula meanings, leading to an error in their application \citep{lightman2023lets}.

\textbf{Missing Step:} Error entails an incomplete generation of reasoning process, lacking a necessary inference step. The addition of such a step could yield the correct result \citep{wei2022chain}.

\textbf{Contradictory Step:} Error manifests inconsistency between preceding and subsequent reasoning steps, resulting in discrepancy within the inference chain \citep{golovneva2022roscoe}.

Among above, unit conversion error, operator error, and formula confusion error can be categorized as \textbf{common sense error}, indicating errors in the relationships that should be understood within worldly common sense. Here, common sense error leans toward factual error, while hallucination leans toward faithful error. 

From the perspective of equation, calculation error is equivalent to errors on the right-hand side of the equation. Counting error, context value error, contradictory step, unit conversion error are equivalent to errors in one operand on the left-hand side of the equation. Operator error is equivalent to errors in one operator on the left-hand side of the equation, while formula confusion error, and hallucination are equivalent to errors in both operands and operators on the left-hand side of the equation.

\section{Generation Rules Design and Examples}
\label{sec:appendix rules}
To generate cases conforming to the nine error types defined in Table \ref{tab:error type definition}, we formulate generation rules for each type and manually craft high-quality examples according to these rules for GPT-4 to emulate.

We solely focus on single-step and single-type errors. Given the error type and the original solution to the question, we instruct GPT-4 to randomly select a step and modify it according to the generation rule of this error type. To emulate a realistic error process, subsequent steps referencing the result of this modified step are also affected. We then filter out cases where the final results after transformation differ from the correct results for evaluation purpose.

\textbf{Calculation Error:} Only the calculation result of a randomly selected step in the original solution is modified, without altering any operands or operators within the expression.

\textbf{Counting Error:} Here, we address issues involving counting days. A modification is made solely to the count result of a step where counting occurs in the original solution. For instance, while the original solution counts Saturday and Sunday as two days, the transformed solution counts them as one day incorrectly.

\textbf{Context Value Error:} An incorrect reference to a number in the question is introduced solely in one step of the original solution. Since only one step is considered erroneous, all other steps referencing this number continue to do so correctly.

\textbf{Hallucination:} Additional information affecting the final outcome, not mentioned in the question statement, is inserted solely into one step of the original solution. 

\textbf{Unit Conversion Error:} An incorrect unit conversion is applied solely to a step in the original solution where unit conversion appears. 

\textbf{Operator Error:} A random modification is made solely to one operator within a formula of a step in the original solution, such as changing addition to subtraction, multiplication to division. The formula's result should remain correctly calculated after the operator change.

\textbf{Formula Confusion Error:} The formula used in one step of the original solution is mistakenly replaced, such as substituting the perimeter formula of a rectangle with the area formula.

\textbf{Missing Step:} The transformed solution should be one step shorter than the original solution. It can occur in three scenarios: deleting the first, middle, or last step. For the first step, step 1 often references the number from the question, so subsequent steps referencing its outcome should directly reference the number from the question after deleting it. If step 1 references multiple numbers, the largest one is selected for subsequent relevant steps. For middle steps, if the deleted middle step refers to the result of only one preceding step, subsequent relevant steps need to reference the result of the preceding step after deleting. Otherwise the largest number from multiple numbers it references is selected for subsequent relevant steps. For the last step, it can be simply deleted, and the result of the second-to-last step becomes the final outcome.

\textbf{Contradictory Step:} An erroneous reference to the result of the preceding relevant step is introduced solely into one step of the original solution. As only one step error is considered, all other steps referencing the result of the preceding relevant step continue to correctly reference it.

It is worth noting that errors involving counting error, unit conversion error, and formula confusion error require selecting appropriate questions and original solutions for transformation. However, other errors can be converted using any questions and original solutions.

Here is a \textbf{[Question]} and its \textbf{[Correct solution]}. We use them for converting different types of errors. The following are examples for the seven wrong types related to this \textbf{[Question]}.
\newline 
\textbf{[Question]:} 
\textit{On the weekend, Tony will walk to the store. On weekdays, he runs to the store. When he walks, he goes 2 MPH. When he runs, he goes 10 MPH. The store is 4 miles away. If he goes on Sunday, Tuesday, and Thursday, what is the average time in minutes that he spends to get to the store?}
\newline 
\textbf{[Correct solution]:}
\textit{On Sunday he takes 2 hours to get there because 4 / 2 = 2\newline 
This takes him 120 minutes because 2 x 60 = 120\newline On Tuesday and Thursday, it takes him .4 hours to get to the store because 4 / 10 = .4\newline On Tuesday and Thursday, it takes him 24 minutes to get to the store because .4 x 60 = 24\newline In total it takes him 168 minutes to get to the store because 120 + 24 + 24 = 168\newline On average it takes him 56 minutes because 168 / 3 = 56.}
\newline 
\textbf{Calculation Error:} 
\newline 
\textbf{[Transformed solution]:}
\textit{On Sunday he takes 2 hours to get there because 4 / 2 = 2\newline 
This takes him 120 minutes because 2 x 60 = 120\newline On Tuesday and Thursday, it takes him .4 hours to get to the store because 4 / 10 = .4\newline On Tuesday and Thursday, it takes him 24 minutes to get to the store because .4 x 60 = 24\newline In total it takes him 168 minutes to get to the store because 120 + 24 + 24 = 168\newline On average it takes him \textcolor{red}{55} minutes because \textcolor{red}{168 / 3 = 55}.}
\newline
\textbf{[Explanation]:} \textit{The operands and operators of the formula in step 6 are correct, but only the result is incorrectly calculated as 55 instead of 56.}
\newline 
\textbf{Counting Error:} 
\newline 
\textbf{[Transformed solution]:}
\textit{On Sunday he takes 2 hours to get there because 4 / 2 = 2\newline 
This takes him 120 minutes because 2 x 60 = 120\newline On Tuesday and Thursday, it takes him .4 hours to get to the store because 4 / 10 = .4\newline On Tuesday and Thursday, it takes him 24 minutes to get to the store because .4 x 60 = 24\newline In total it takes him 168 minutes to get to the store because 120 + 24 + 24 = 168\newline On average it takes him \textcolor{red}{84} minutes because \textcolor{red}{168 / 2 = 84}.}
\newline
\textbf{[Explanation]:} \textit{Step 6 counts Sunday, Tuesday, and Thursday wrongly as 2 days instead of 3 days, only resulting in an operand error in the formula.}
\newline 
\textbf{Context Value Error:} 
\newline 
\textbf{[Transformed solution]:}
\textit{On Sunday he takes 2 hours to get there because 4 / 2 = 2\newline 
This takes him 120 minutes because 2 x 60 = 120\newline On Tuesday and Thursday, it takes him .4 hours to get to the store because 4 / 10 = .4\newline On Tuesday and Thursday, it takes him \textcolor{red}{.2} hours to get to the store because \textcolor{red}{4 / 20 = .2}\newline On Tuesday and Thursday, it takes him 12 minutes to get to the store because .2 x 60 = 12\newline In total it takes him 144 minutes to get to the store because 120 + 12 + 12 = 144\newline On average it takes him 48 minutes because 144 / 3 = 48.}
\newline
\textbf{[Explanation]:} \textit{Step 3 mistakenly references the number 20 instead of 10 from the question, only resulting in an operand error in the formula. The subsequent steps are affected by it. Please note that we only consider errors of single step and single type, and step 2 still correctly references 10. }
\newline 
\textbf{Hallucination:} 
\newline 
\textbf{[Transformed solution]:}
\textit{On Sunday he takes 2 hours to get there because 4 / 2 = 2\newline 
This takes him 120 minutes because 2 x 60 = 120\newline On Tuesday and Thursday, it takes him .4 hours to get to the store because 4 / 10 = .4\newline On Tuesday and Thursday, it takes him 24 minutes to get to the store because .4 x 60 = 24\newline \textcolor{red}{Because the road congestion on Tuesday takes an additional 20 minutes}, so in total it takes him \textcolor{red}{168} minutes to get to the store because \textcolor{red}{120 + 24 + 24 + 20 = 188}\newline On average it takes him On average it takes him 62.6 minutes because 188 / 3 = 62.6.}
\newline
\textbf{[Explanation]:} \textit{Step 5 adds the additional information <Because the road congestion on Tuesday takes an additional 20 minutes> not mentioned in the questions, causing the result of step 5 to be overestimated by 20. And it influences step 6, which references its result.}
\newline 
\textbf{Unit Conversion Error:} 
\newline 
\textbf{[Transformed solution]:}
\textit{On Sunday he takes 2 hours to get there because 4 / 2 = 2\newline This takes him \textcolor{red}{100} minutes because \textcolor{red}{2 x 50 = 100}\newline On Tuesday and Thursday, it takes him .4 hours to get to the store because 4 / 10 = .4\newline On Tuesday and Thursday, it takes him 24 minutes to get to the store because .4 x 60 = 24\newline In total it takes him 148 minutes to get to the store because 100 + 24 + 24 = 148\newline On average it takes him 49.3 minutes because 148 / 3 = 49.3.}
\newline
\textbf{[Explanation]:} \textit{Step 2 performs an incorrect unit conversion and mistakenly assumes that one hour has 50 minutes, which only results in an error in one operand in the formula. The subsequent steps 5 and 6 are affected by it. Because we only consider errors of single step and single type, step 4 still correctly performs unit conversion. }
\newline 
\textbf{Operator Error:} 
\newline 
\textbf{[Transformed solution]:}
\textit{On Sunday he takes 2 hours to get there because 4 / 2 = 2\newline 
This takes him 120 minutes because 2 x 60 = 120\newline On Tuesday and Thursday, it takes him .4 hours to get to the store because 4 / 10 = .4\newline On Tuesday and Thursday, it takes him 24 minutes to get to the store because .4 x 60 = 24\newline In total it takes him 168 minutes to get to the store because 120 + 24 + 24 = 168\newline On average it takes him \textcolor{red}{171} minutes because \textcolor{red}{168 + 3 = 171}.}
\newline
\textbf{[Explanation]:} \textit{Step 6 mistakenly uses addition instead of division, and only one operator in the formula is incorrect. }
\newline 
\textbf{Missing Step:} 
\newline 
\textbf{[Transformed solution]:}
\textit{On Sunday he takes 2 hours to get there because 4 / 2 = 2 \newline On Tuesday and Thursday, it takes him .4 hours to get to the store because 4 / 10 = .4\newline  On Tuesday and Thursday, it takes him 24 minutes to get to the store because .4 x 60 = 24\newline In total it takes him \textcolor{red}{50} minutes to get to the store because \textcolor{red}{2 + 24 + 24 = 50}\newline On average it takes him 16.6 minutes because 50 / 3 = 16.6.}
\newline
\textbf{[Explanation]:} \textit{Step 4 does not convert the time he went to the store on Sunday from hours to minutes, but directly adds up the time on Sunday (hours) and the time on Tuesday and Thursday (minutes). So there is a missing step here to convert Sunday's time from hours to minutes. }
\newline 
\textbf{Contradictory Step:} 
\newline 
\textbf{[Transformed solution]:}
\textit{On Sunday he takes 2 hours to get there because 4 / 2 = 2\newline 
This takes him 120 minutes because 2 x 60 = 120\newline On Tuesday and Thursday, it takes him .4 hours to get to the store because 4 / 10 = .4\newline On Tuesday and Thursday, it takes him 24 minutes to get to the store because .4 x 60 = 24\newline In total it takes him \textcolor{red}{188} minutes to get to the store because \textcolor{red}{140 + 24 + 24 = 188}\newline On average it takes him 62.6 minutes because 188 / 3 = 62.6 }
\newline 
\textbf{[Explanation]:} \textit{Step 5 erroneously references the result 140 of step 2 instead of 120, which only results in an error in one operand in the formula. }

Here is another \textbf{[Question]} and its \textbf{[Correct solution]} for converting formula confusion error. 
\newline 
\textbf{[Question]:} 
\textit{Linda is painting her bedroom. Her bedroom has 4 walls, with the room being 20 feet wide by 20 feet long by 8 feet tall. One wall has a 3-foot by 7-foot doorway. A second wall has a 6-foot by 4-foot window. A third wall has a 5-foot by 7-foot doorway to a walk-in-closet. And the fourth wall is completely solid. What is the total area of wall space that Linda will have to paint?}
\newline 
\textbf{[Correct solution]:}
\textit{The solid wall is 8 ft. * 20 ft. = 160 sq. ft. \newline The doorway is 3 ft. * 7 ft. = 21 sq. ft.\newline The window is 6 ft. * 4 ft. = 24 sq. ft.\newline The closet door is 5 ft. * 7 ft. = 35 sq. ft.\newline The total area of the doors and windows is 21 sq. ft + 24 sq. ft. + 35 sq. ft. = 80 sq. ft.\newline The solid wall is 160 sq. ft., so before the areas of the doors and window are taken into account, the total wall area is 4 * 160 sq. ft. = 640 sq. ft.\newline Taking into account the doors and window, the total wall area Linda will have to paint is 640 sq. ft. - 80 sq. ft. = 560 sq. ft.} 
\newline 
\textbf{Formula Confusion Error:} 
\newline 
\textbf{[Transformed solution]:}
\textit{The solid wall is 8 ft. * 20 ft. = 160 sq. ft.\newline The doorway is \textcolor{red}{3 ft.+7 ft.+3 ft. +7 ft.=20 sq. ft.}\newline The window is 6 ft. * 4 ft. = 24 sq. ft.\newline The closet door is 5 ft. * 7 ft. = 35 sq. ft.\newline The total area of the doors and windows is 20 sq. ft + 24 sq. ft. + 35 sq. ft. = 79 sq. ft.\newline The solid wall is 160 sq. ft., so before the areas of the doors and window are taken into account, the total wall area is 4 * 160 sq. ft. = 640 sq. ft.\newline Taking into account the doors and window, the total wall area Linda will have to paint is 640 sq. ft. - 79 sq. ft. = 561 sq. ft.}
\newline 
\textbf{[Explanation]:} \textit{Step 2 confuses the perimeter and area formulas of rectangle and it should calculate the area of the rectangle which is equal to the length multiplied by width, rather than the length plus length plus width plus width, equivalent to the perimeter of the rectangle. And step 5 and 7 referencing the result of step 2 are affected.}

\section{Human Evaluation}\label{Human Evaluation}

\begin{figure}[htbp]   
	\centering
	\includegraphics[width=\linewidth,scale=1.00]{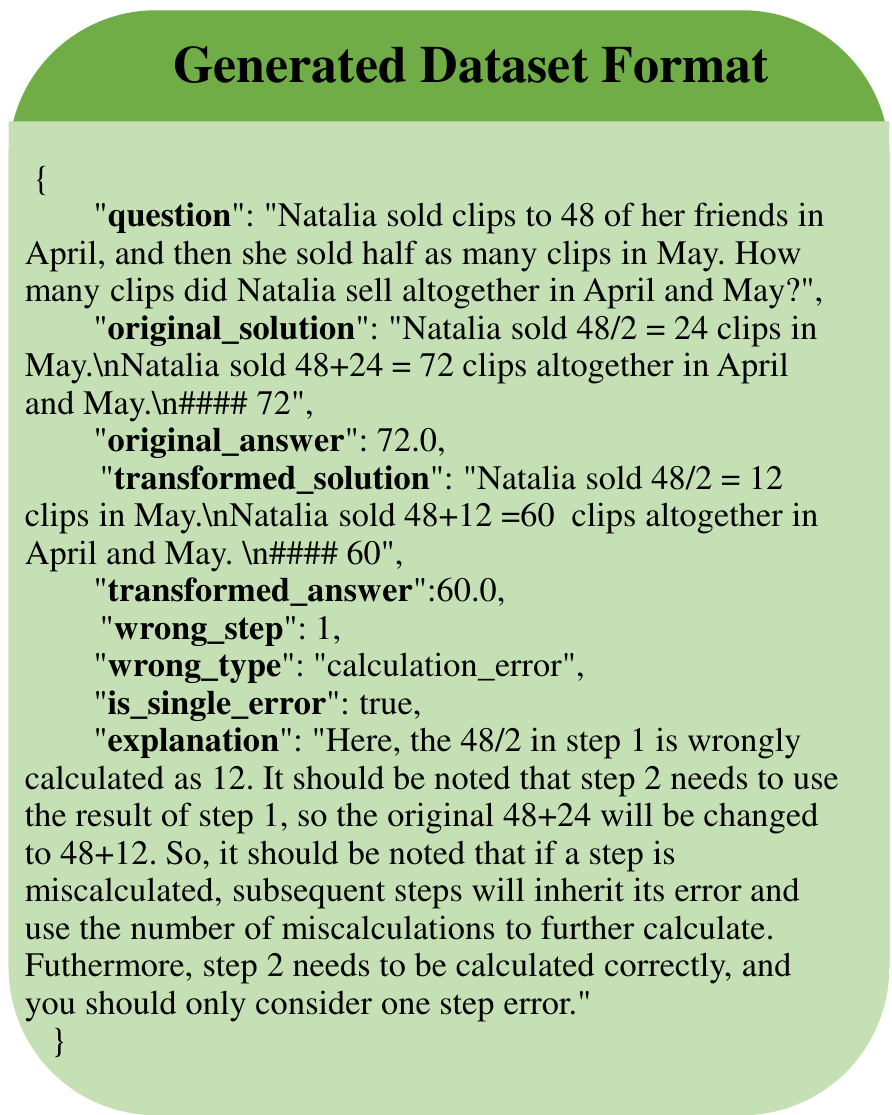}
	\caption{Dataset format.}
	\label{dataset format}
\end{figure}

\noindent
\textbf{Assessment Procedure:}
The format of the dataset we generate is illustrated in Figure \ref{dataset format}. Evaluators should first comprehend the question and original solution. Subsequently, they should carefully compare the original solution with the transformed solution to determine if the transformed one contains single-step and single-type error according to specific error type rule. Additionally, evaluators should ascertain whether the generated wrong step represents the first error step.

\noindent
\textbf{Assessment Quality Control:}
We enlist two evaluators to assess 10 cases of each error type in every dataset, totaling 180 cases. A consensus between the two evaluators is required for a case to be deemed satisfactory. In cases of disagreement between two evaluators, a third party is consulted for a final decision. Throughout the evaluation, our generated dataset have achieved an accuracy rate of 92.5\%, demonstrating its suitability for evaluating the ability of LLMs to identify and rectify errors.

\section{Additional In-depth Experiments}

\subsection{Influence of Step Count}
\begin{figure*}[ht]
  \begin{minipage}[t]{0.5\linewidth}
    \centering
    \includegraphics[width=\textwidth]{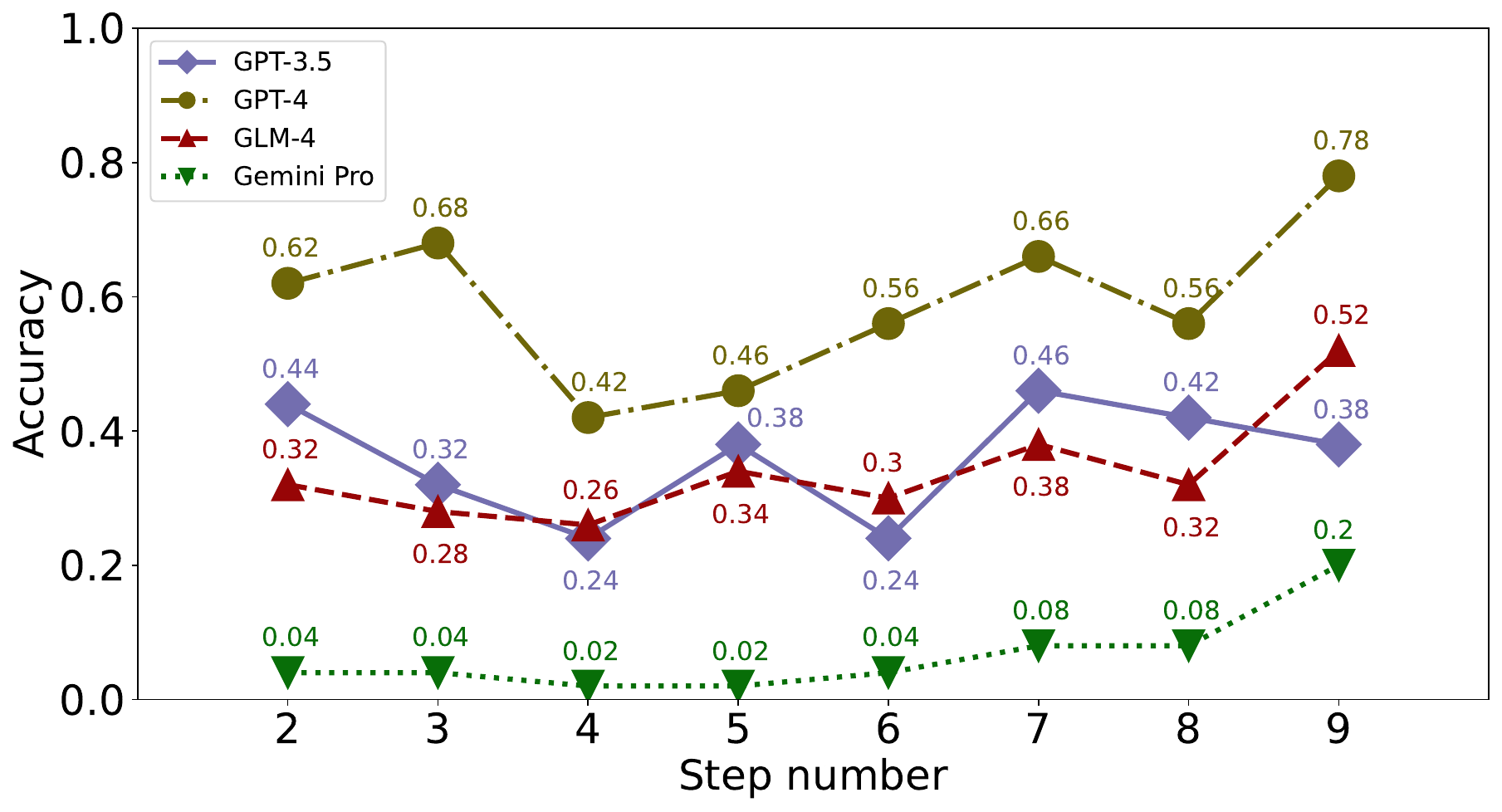}
    \caption{The influence of step number in GSM8K on \textit{EP}.}
    \label{fig:GSM8K CA EP: the influence of step number}
  \end{minipage}%
  \hspace{4mm}
  \begin{minipage}[t]{0.5\linewidth}
    \centering
    \includegraphics[width=\textwidth]{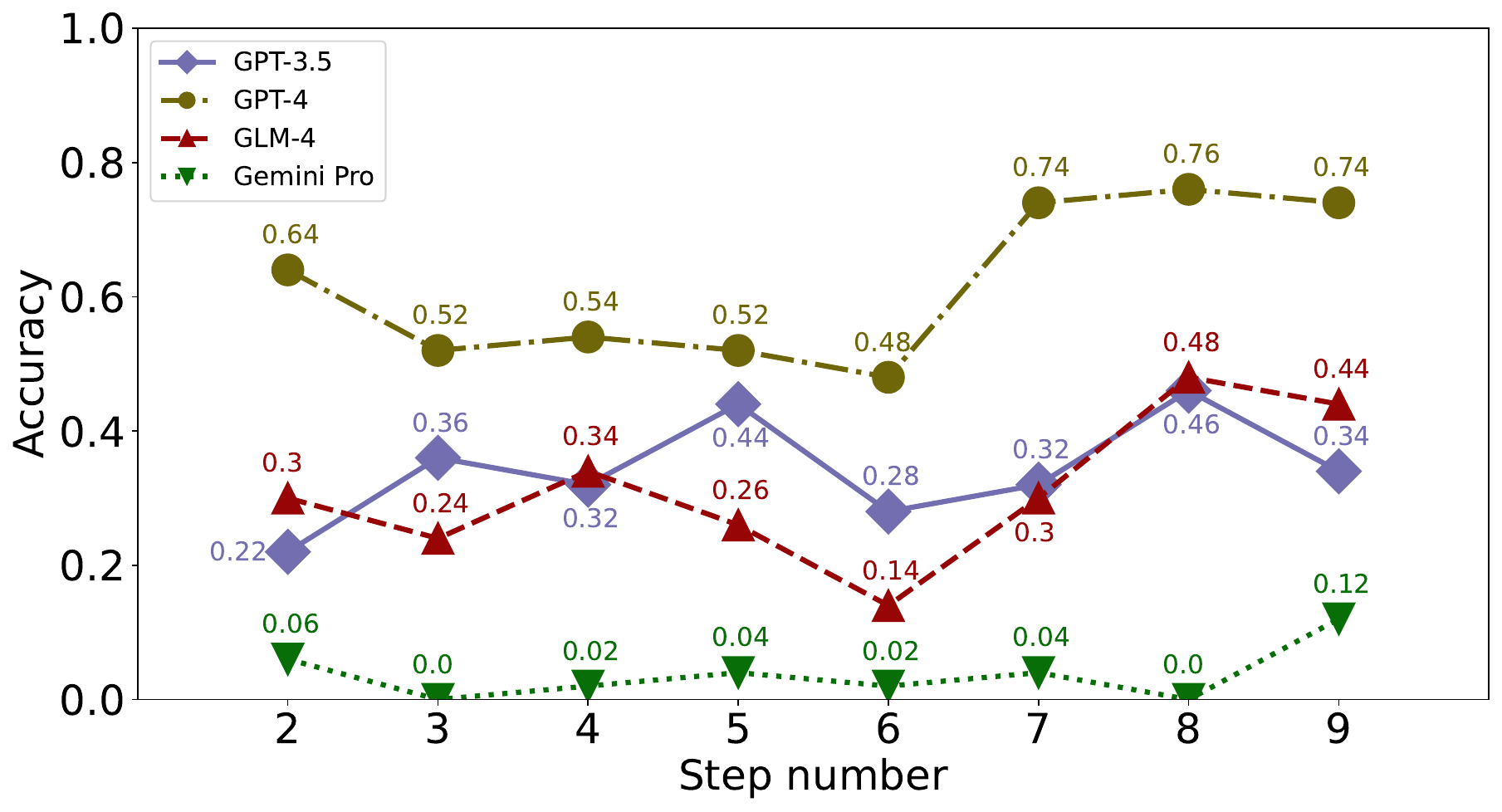}
    \caption{The influence of step number in MathQA on \textit{EP}.}
    \label{fig:MathQA CA EP: the influence of step number}
  \end{minipage}
\end{figure*}

\begin{figure*}[ht]
  \begin{minipage}[t]{0.5\linewidth}
    \centering
    \includegraphics[width=\textwidth]{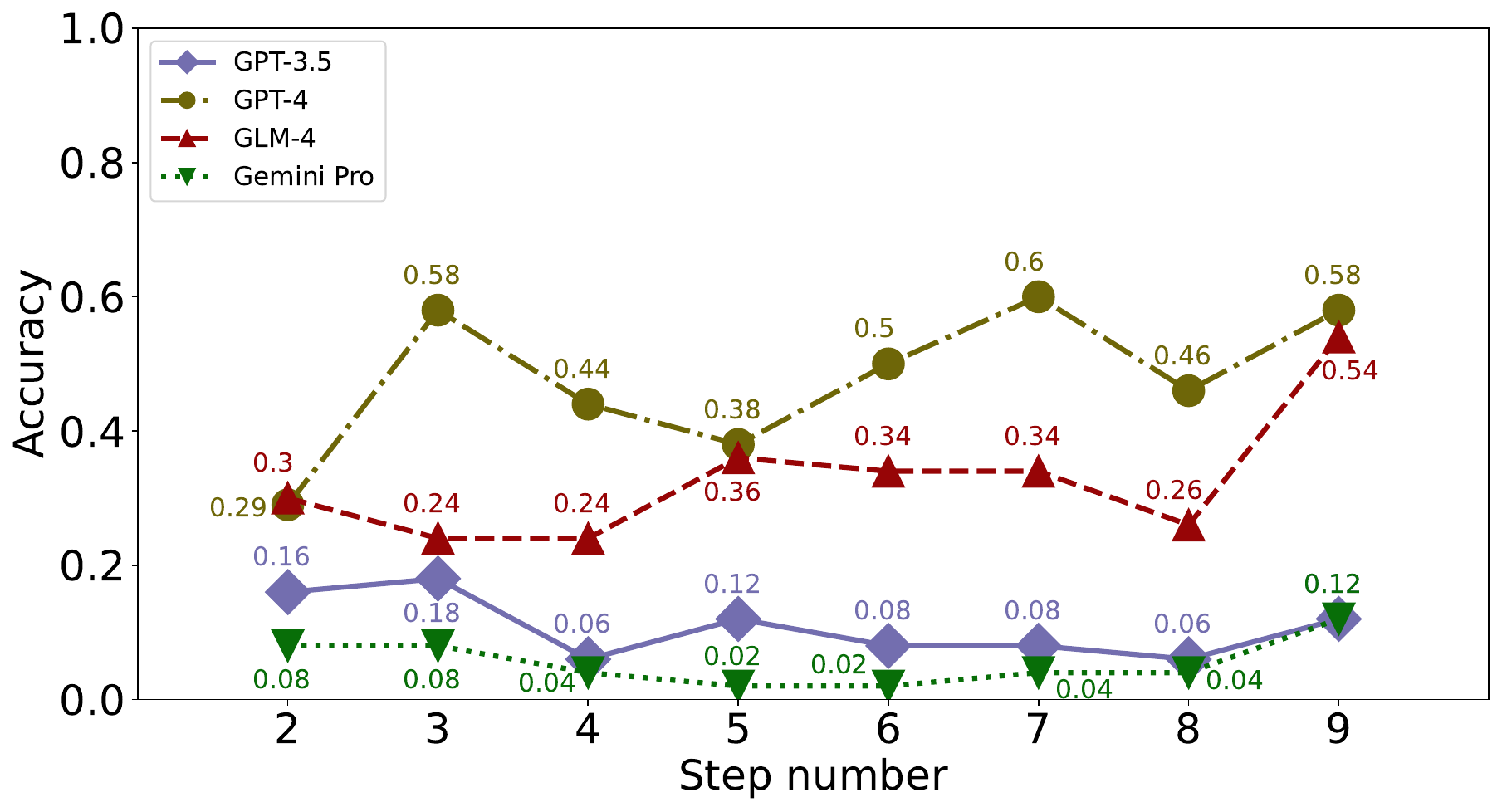}
    \caption{The influence of step number in GSM8K on \textit{EC}.}
    \label{fig:GSM8K CA EC: the influence of step number}
  \end{minipage}%
   \hspace{4mm}
  \begin{minipage}[t]{0.5\linewidth}
    \centering
    \includegraphics[width=\textwidth]{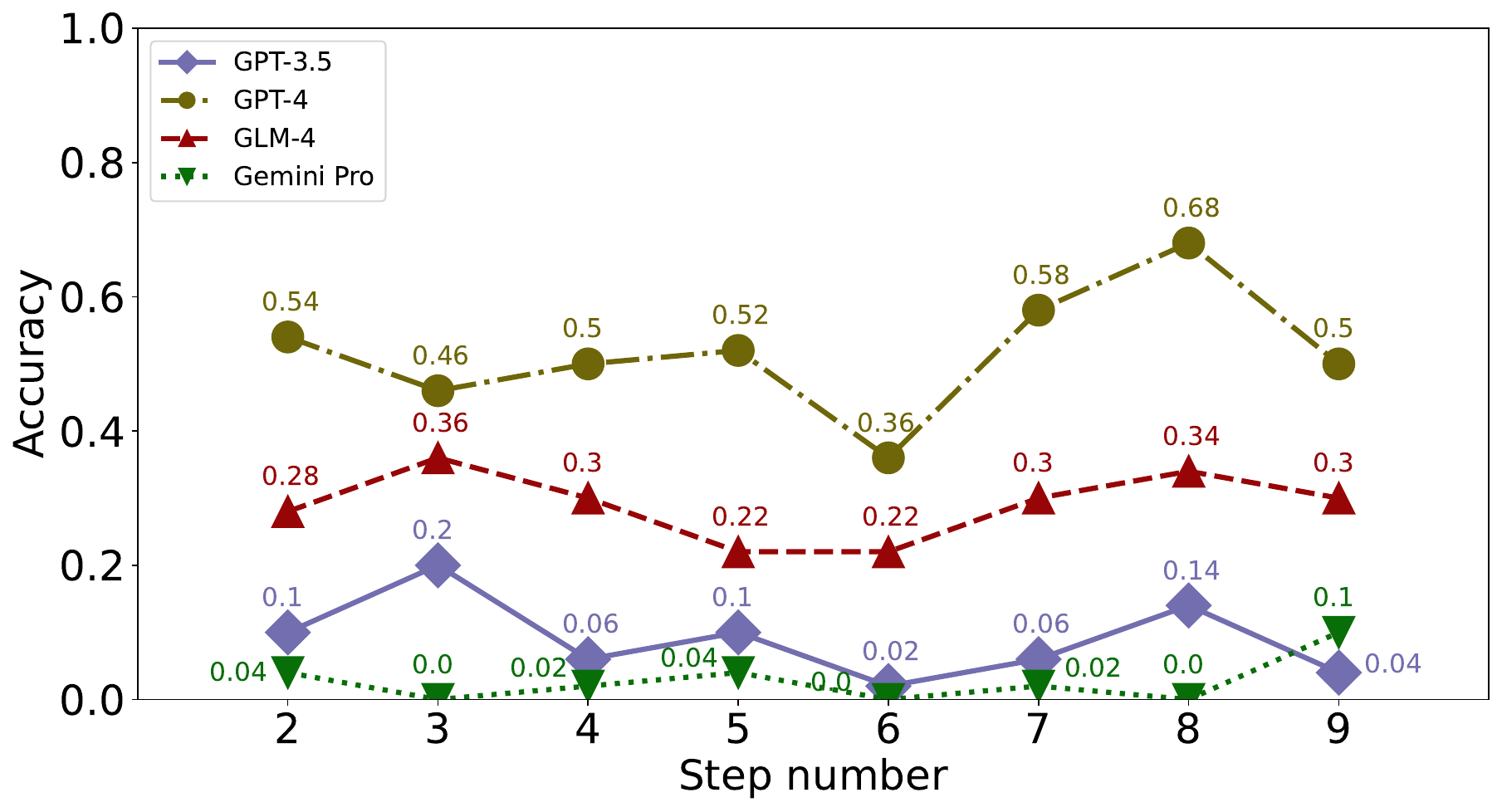}
    \caption{The influence of step number in MathQA on \textit{EC}.}
    \label{fig:MathQA CA EC: the influence of step number}
  \end{minipage}
\end{figure*}

We examine the influence of step count on \textit{EP} and \textit{EC}. For calculation error, we select 50 solutions ranging from 2 to 9 steps to transform. It can be observed that the accuracy of identification and correction is not significantly affected by the number of steps in Figure \labelcref{fig:GSM8K CA EP: the influence of step number,fig:MathQA CA EP: the influence of step number,fig:GSM8K CA EC: the influence of step number,fig:MathQA CA EC: the influence of step number}.

\subsection{Influence of the Wrong Step Order} 
\begin{figure*}[ht]
  \begin{minipage}[t]{0.5\linewidth}
    \centering
    \includegraphics[width=\textwidth]{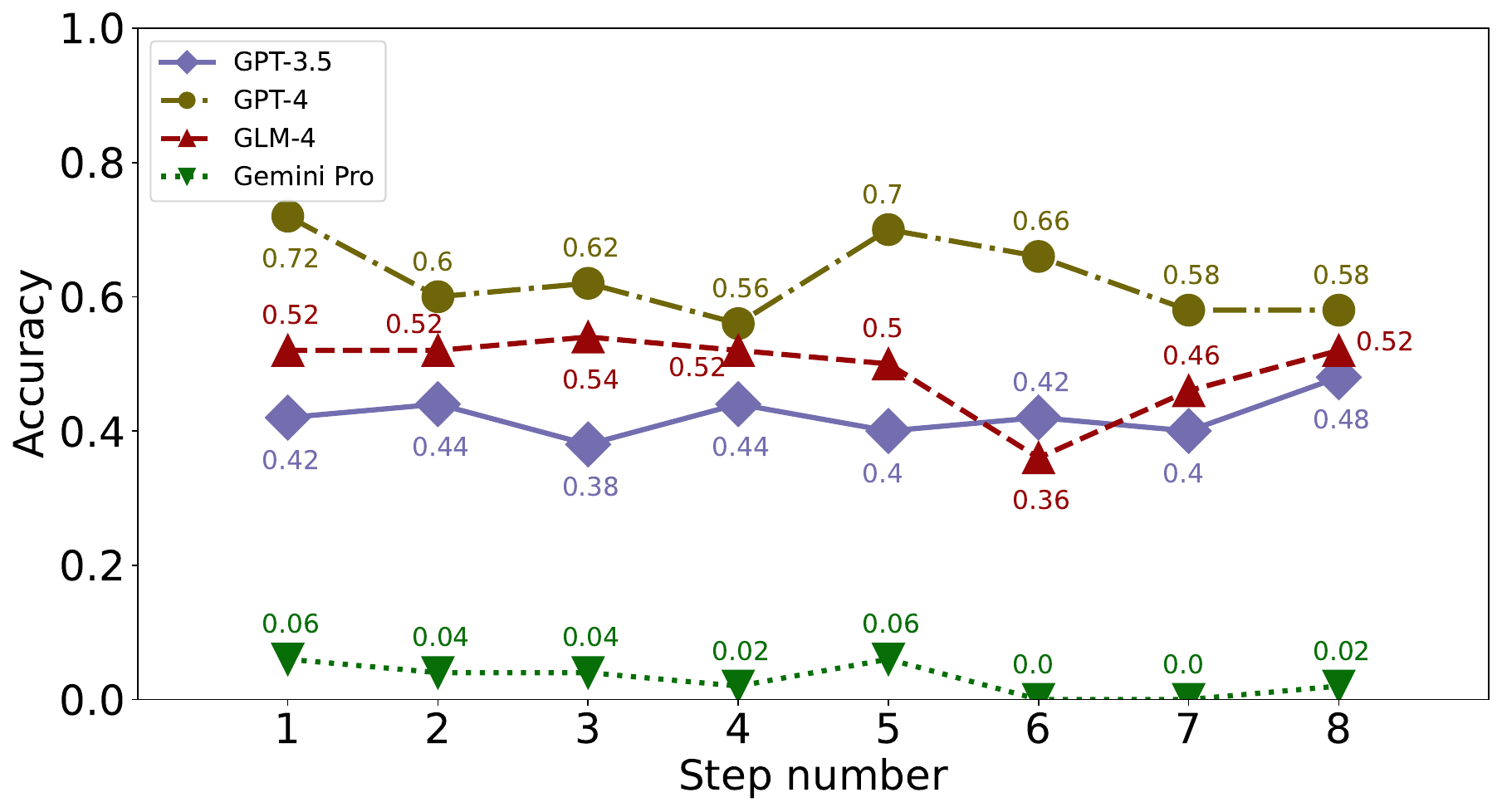}
    \caption{The influence of wrong step order in GSM8K on \textit{EP}.}
    \label{fig:GSM8K CA EP: the influence of wrong step}
  \end{minipage}%
   \hspace{4mm}
  \begin{minipage}[t]{0.5\linewidth}
    \centering
    \includegraphics[width=\textwidth]{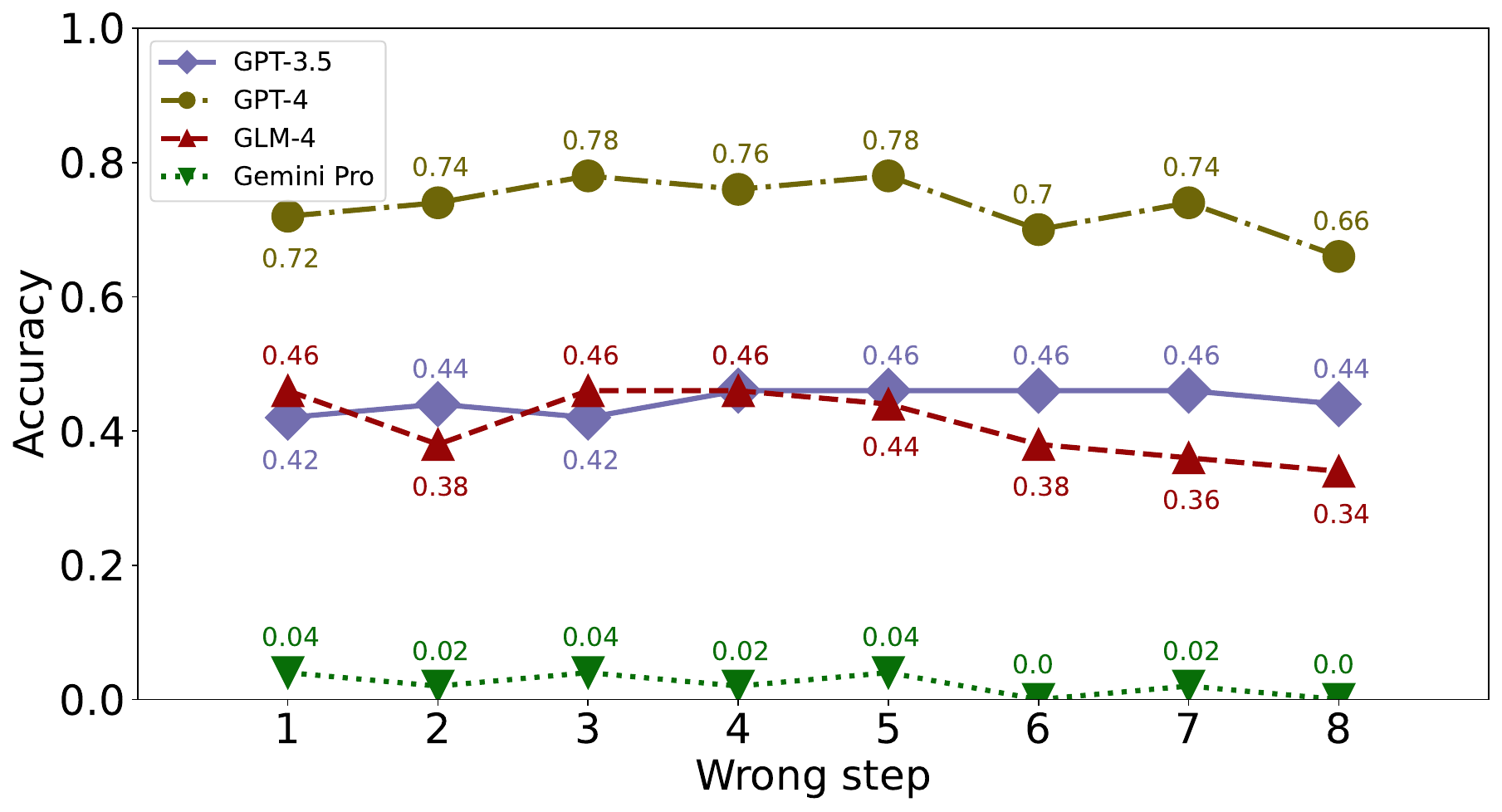}
    \caption{The influence of wrong step order in MathQA on \textit{EP}.}
    \label{fig:MathQA CA EP: the influence of wrong step}
  \end{minipage}
\end{figure*}

\begin{figure*}[ht]
  \begin{minipage}[t]{0.5\linewidth}
    \centering
    \includegraphics[width=\textwidth]{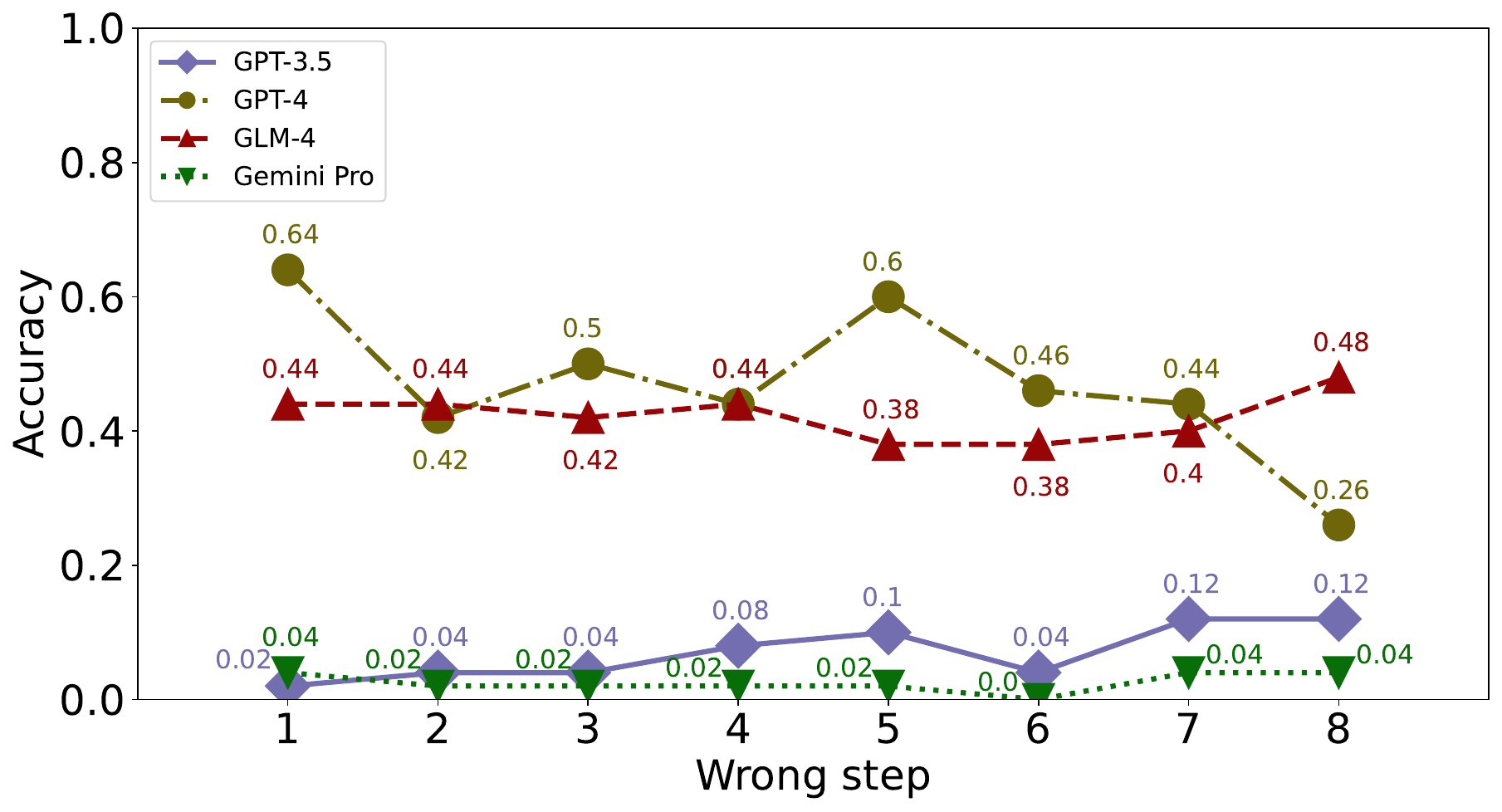}
    \caption{The influence of wrong step order in GSM8K on \textit{EC}.}
    \label{fig:GSM8K CA EC: the influence of wrong step}
  \end{minipage}%
   \hspace{4mm}
  \begin{minipage}[t]{0.5\linewidth}
    \centering
    \includegraphics[width=\textwidth]{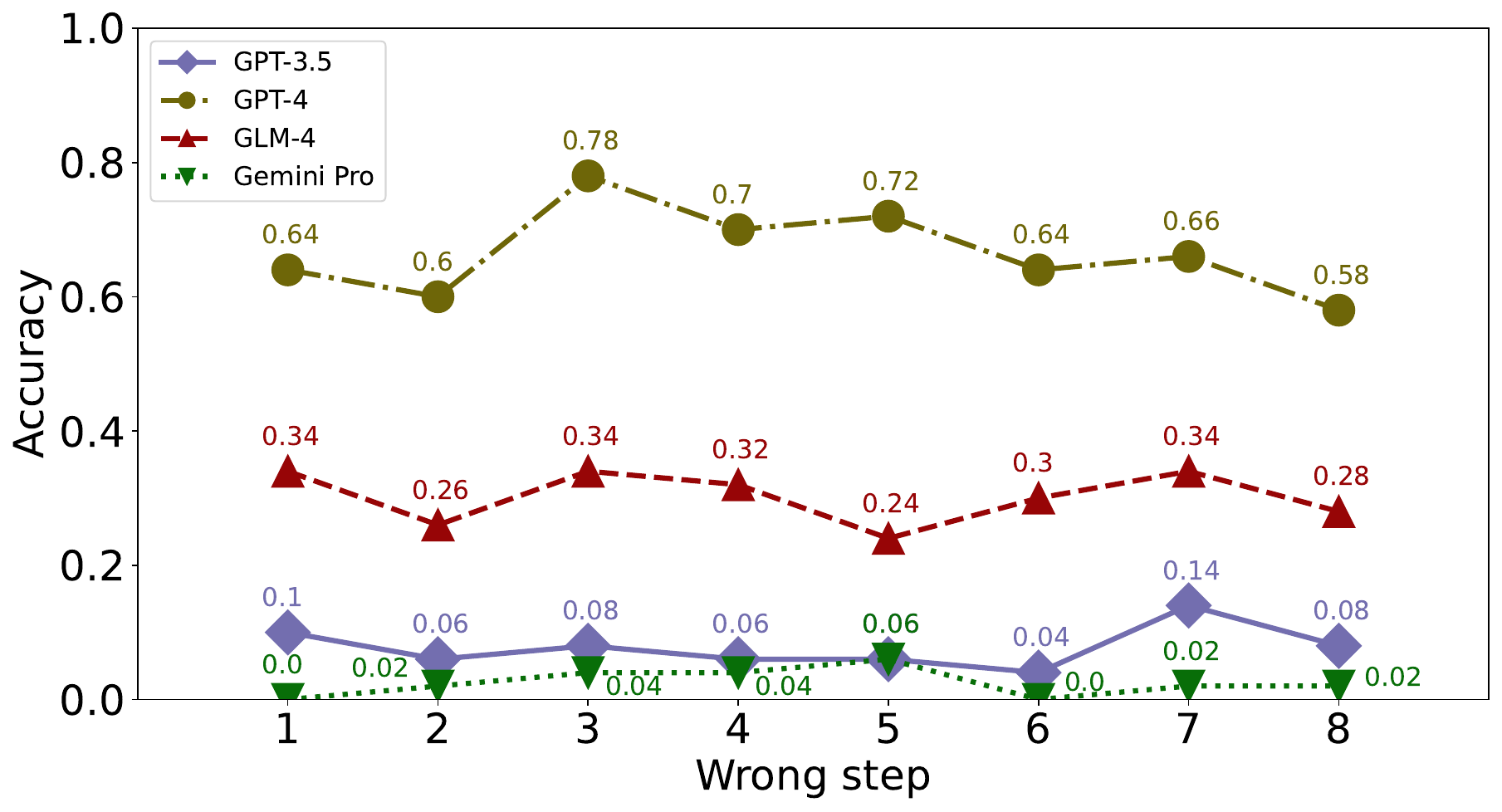}
    \caption{The influence of wrong step order in MathQA on \textit{EC}.}
    \label{fig:MathQA CA EC: the influence of wrong step}
  \end{minipage}
\end{figure*}

We consider the impact of the occurrence order of the error step on \textit{EP} and \textit{EC}. For the 8-step problems involving calculation error, we generate 50 cases for each error step from 1 to 8 for evaluation. It can be observed that the accuracy of identification and correction is  also not significantly affected by the order of the error step in Figure \labelcref{fig:GSM8K CA EP: the influence of wrong step,fig:MathQA CA EP: the influence of wrong step,fig:GSM8K CA EC: the influence of wrong step,fig:MathQA CA EC: the influence of wrong step}.

\subsection{Comparasion between GLM-4 and GPT-4}
To further validate the robustness of our conclusions, we conduct supplementary experiments using GLM-4 for data generation. Due to resource limit, we only use GLM-4 to generate three types of errors – CA, MS, and UC – on GSM8K and MathQA, with 50 instances each, totaling 300 instances. The experimental results are as shown in Table \ref{tab:Comparasion between GLM-4 and GPT-4.}. We arrive at conclusions consistent with those drawn from the dataset generated by GPT-4, e.g., GPT-4's superior error identification and correction capabilities compared to other models. 

\subsection{Comparasion with other math models}\label{Comparasion with other math models}
We conduct the evaluation results of other three math-specialized LLMs: Mistral \citep{jiang2023mistral}, Llemma \citep{azerbayev2023llemma} and LEMA \citep{an2023learning} in their 7B versions. We analyze their performance across different error types and tasks. The experimental results are as shown in Table \ref{tab:Other math models} and \ref{tab:Other math models on tasks.}. It can be observed that LEMA, which is aware of errors, outperforms the other math-specialized LLMs. And the capability of GPT-4 and GLM-4 still far surpasses these open-source models. This indicates that the ability to identify and correct errors of these math-specialized LLMs is inferior to that of the general-purpose powerful LLMs, GPT-4 and GLM-4.

\subsection{Combination Error Analysis}
We conduct some experiments with multi-step and multi-type errors. We first test the combinations of CA and CV. Experimental settings are divided into two: two error types occurring in the same step (Single-step and Two-type Error, ST) and two error types happening in two separate steps (Two-step and Two-type Error, TT). For both settings, we manually annotate 50 data samples and use them to evaluate LLMs' performance in the basic tasks of \textit{EP} and \textit{EC}. The experimental results are shown in the Table \ref{tab:EP combination error.} and \ref{tab:EC combination error.}.

\textbf{Result analysis}. 
By comparing the average accuracy of GPT-4 and GLM-4 with other models, it is evident that GPT-4 and GLM-4 significantly surpass others. As shown in Table \ref{tab:EP combination error.}, for GPT-4 and GLM-4, the \textit{EP} accuracy of ST and TT is higher than that of CA. This implies that the introduction of CV makes CA more prone to exposure. As illustrated in Table \ref{tab:EC combination error.}, for GPT-4 and GLM-4, the \textit{EC} accuracy of ST and TT is higher than that of CA. This is because these models exhibit strong correction ability for identified errors.

\section{Experiment Details}\label{Experiment Details}
This section contains prompts and specific experimental results in the experiment.
\subsection{Prompts and Input Formatting}
\subsubsection{Dataset Generation}\label{Dataset Generation}
\begin{figure*}[htbp]   
	\centering
	\includegraphics[width=\linewidth,scale=1.00]{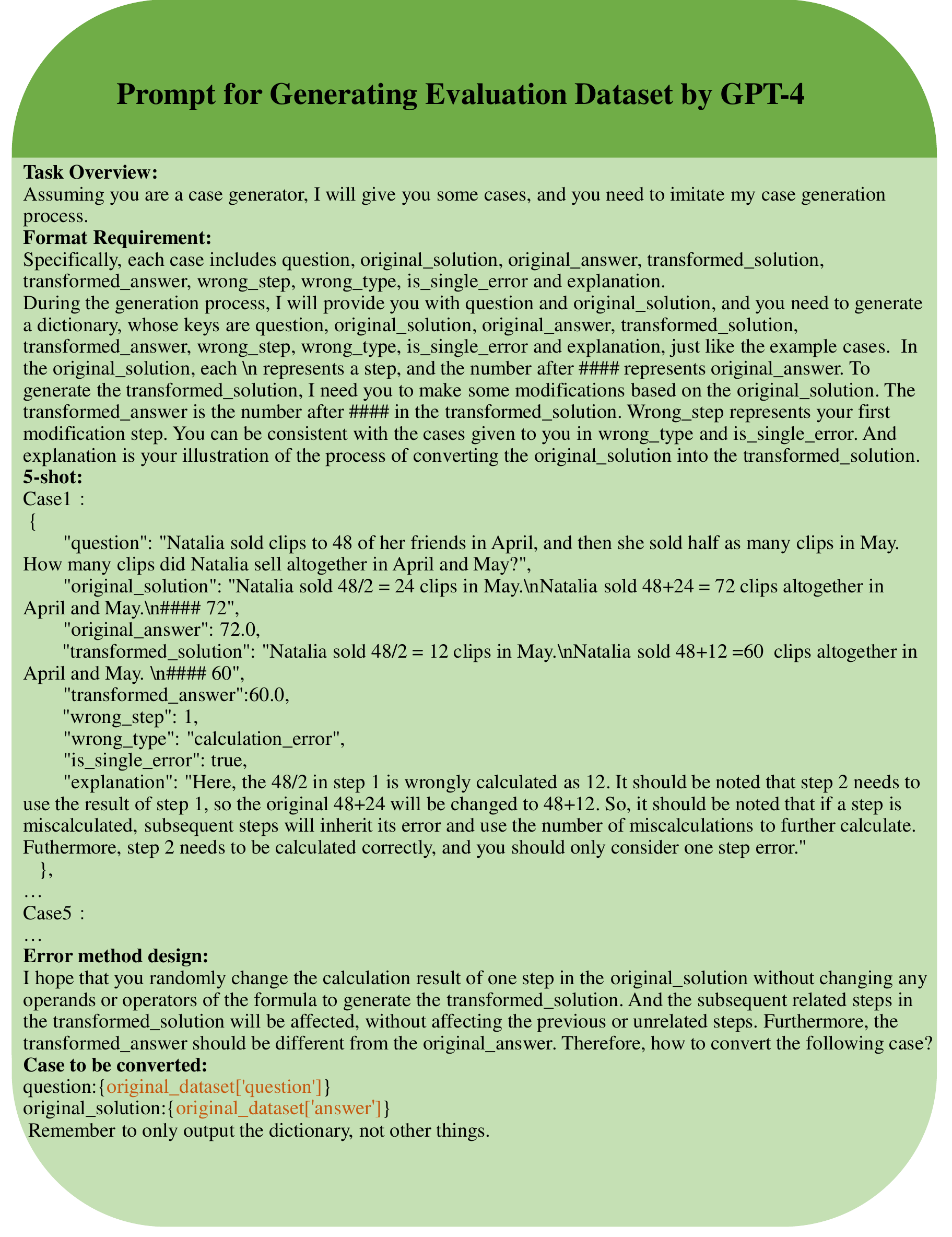}
	\caption{Prompt for generation.}
	\label{fig: Prompt for Generation}
\end{figure*}

The prompt for generating the evaluation dataset is shown in Figure \ref{fig: Prompt for Generation}. After practical experience, we find that 5-shot has good generation results.

\subsubsection{EP}
\begin{figure*}[htbp]   
	\centering
	\includegraphics[width=\linewidth,scale=1.00]{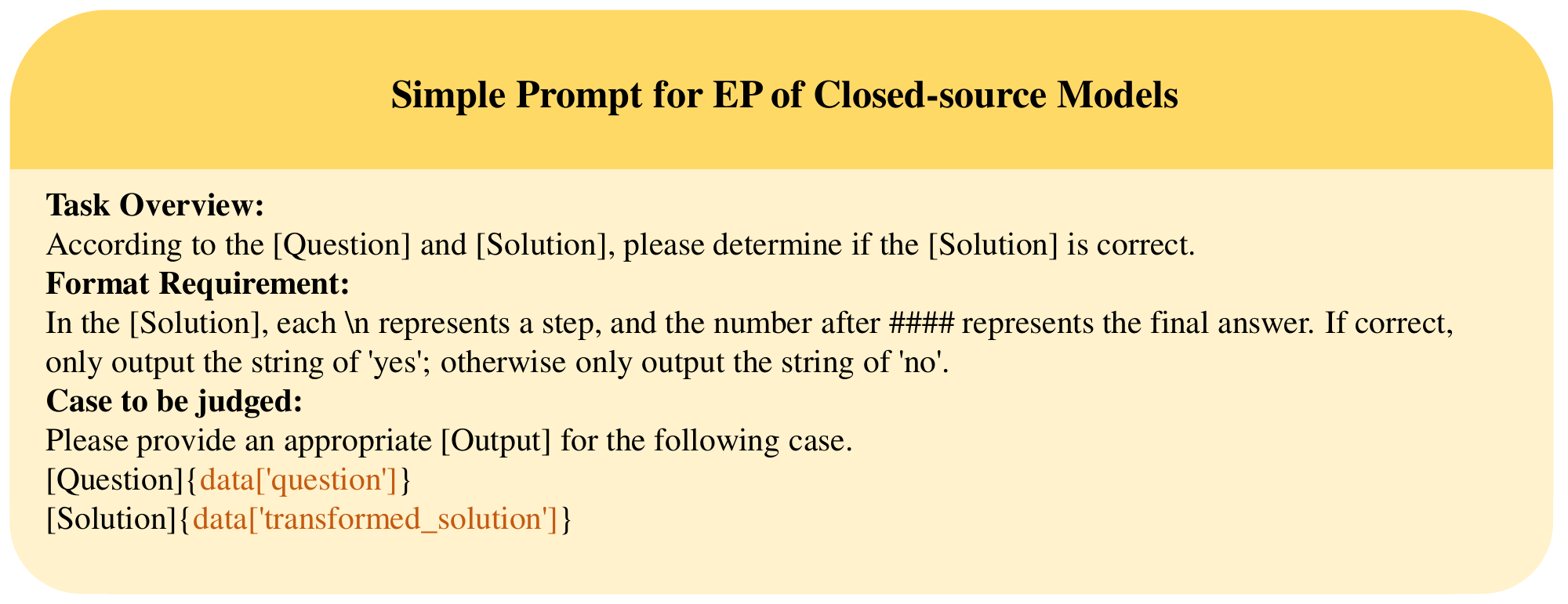}
	\caption{Simple prompt for \textit{EP} on closed-source models.}
	\label{fig: Simple prompt for EP on closed-source models}
\end{figure*}

\begin{figure*}[htbp]   
	\centering
	\includegraphics[width=\linewidth,scale=1.00]{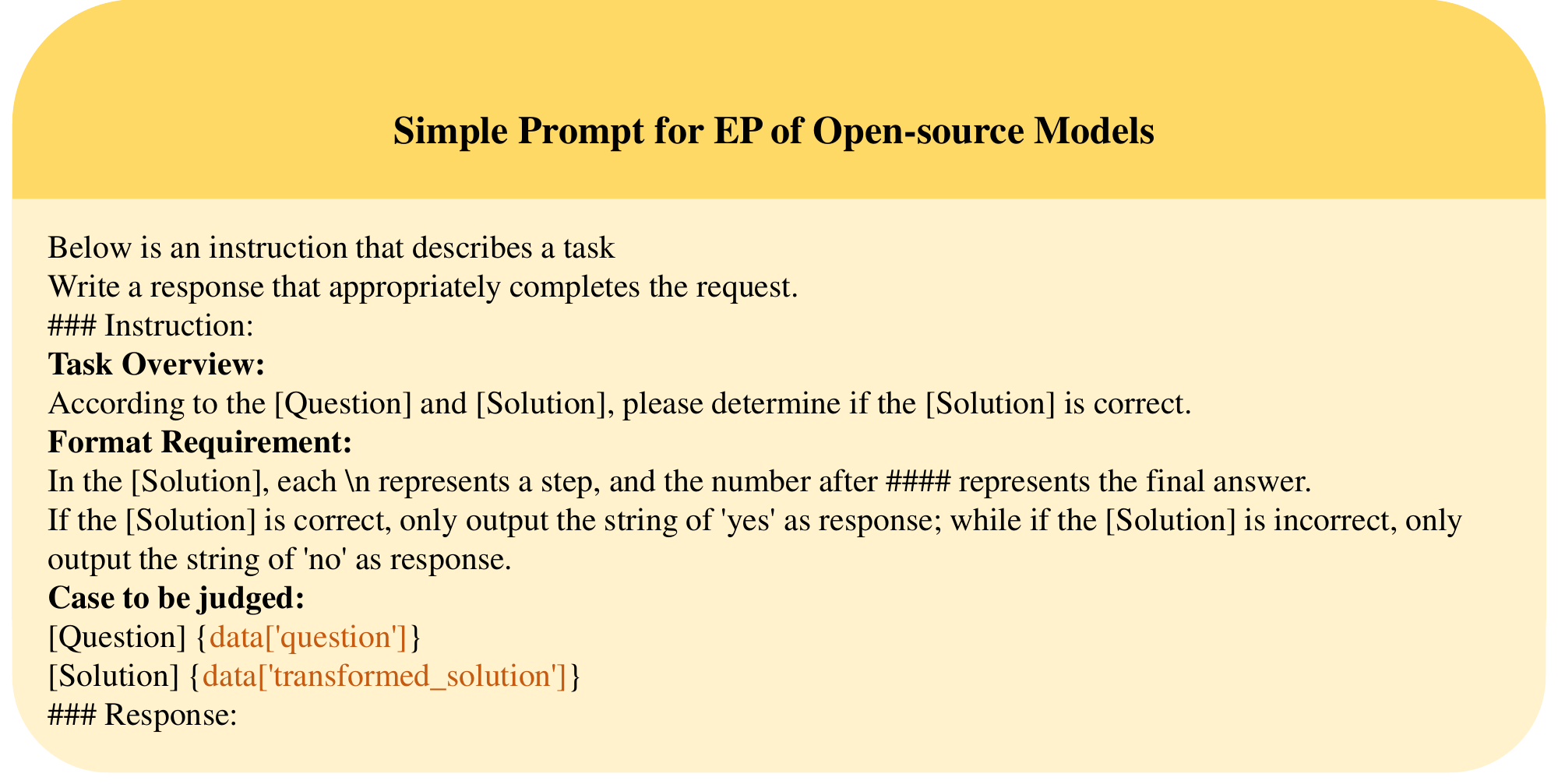}
	\caption{Simple prompt for \textit{EP} on open-source models.}
	\label{fig: Simple prompt for EP on open-source models}
\end{figure*}

\begin{figure*}[htbp]   
	\centering
	\includegraphics[width=\linewidth,scale=1.00]{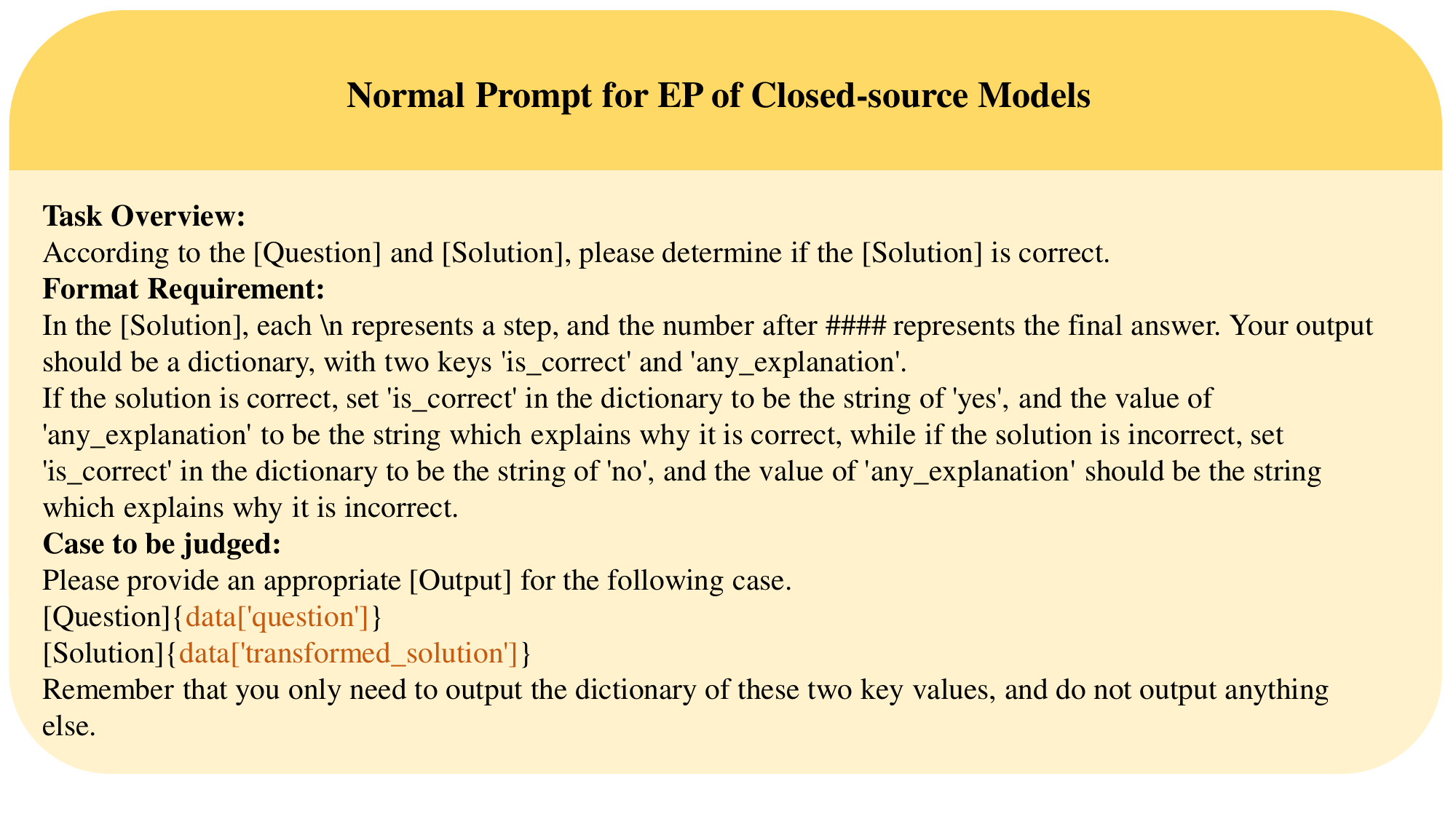}
	\caption{Normal prompt for \textit{EP} on closed-source models.}
	\label{fig: Normal prompt for EP on closed-source models}
\end{figure*}

\begin{figure*}[htbp]   
	\centering
	\includegraphics[width=\linewidth,scale=1.00]{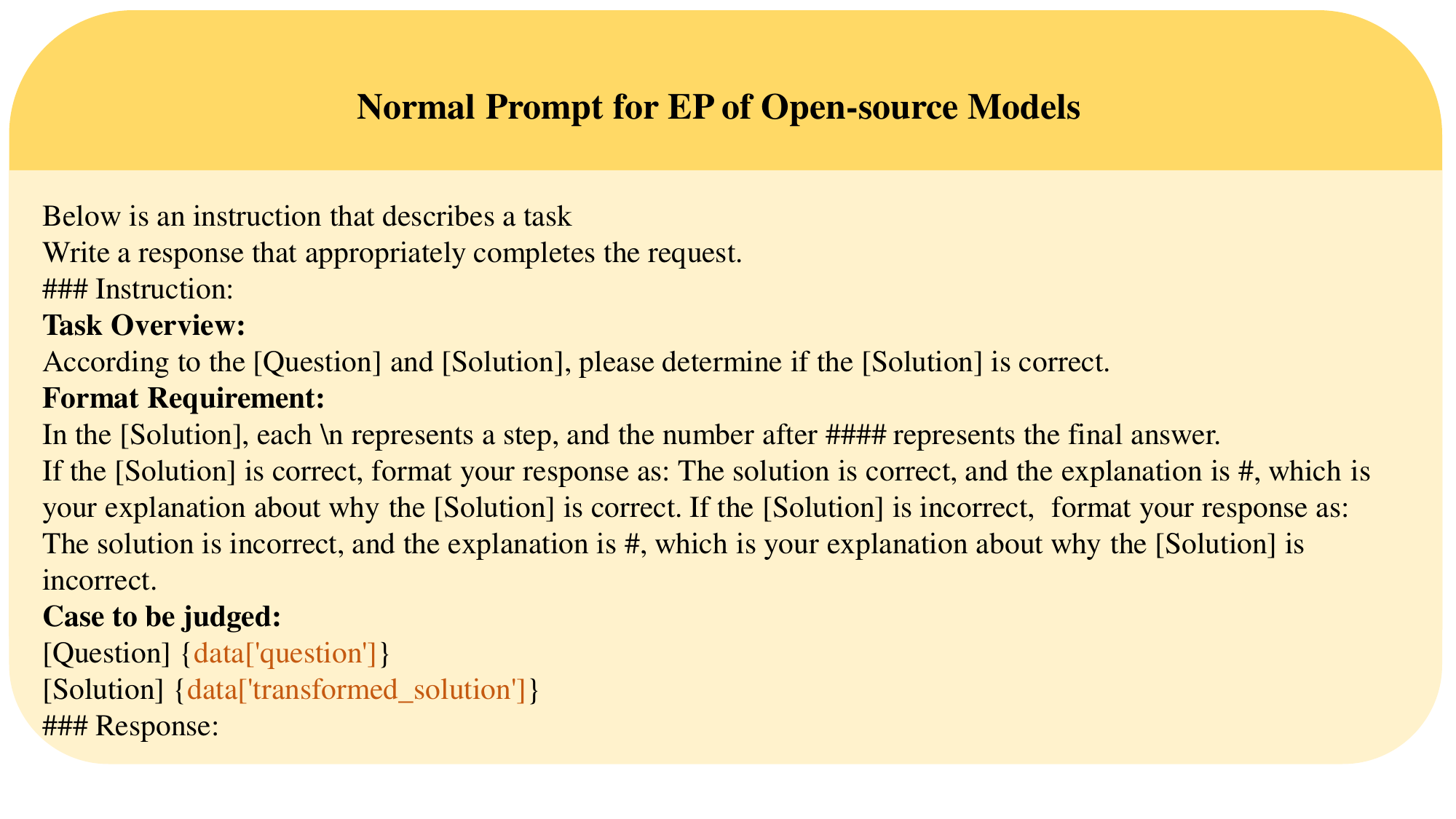}
	\caption{Normal prompt for \textit{EP} on open-source models.}
	\label{fig: Normal prompt for EP on open-source models}
\end{figure*}

\begin{figure*}[htbp]   
	\centering
	\includegraphics[width=\linewidth,scale=1.00]{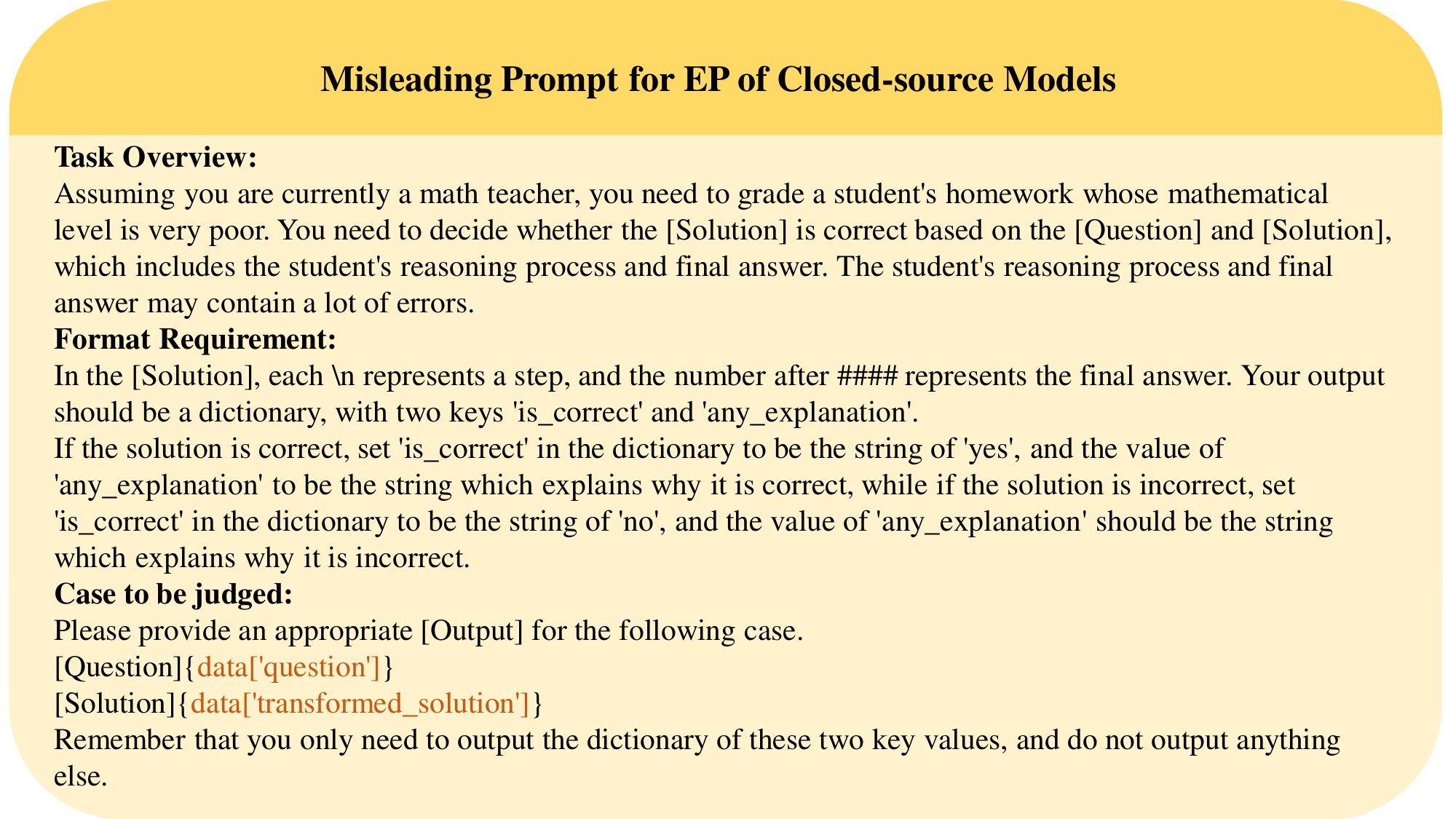}
	\caption{Misleading prompt for \textit{EP} on closed-source models.}
	\label{fig: Misleading prompt for EP on closed-source models}
\end{figure*}

\begin{figure*}[htbp]   
	\centering
	\includegraphics[width=\linewidth,scale=1.00]{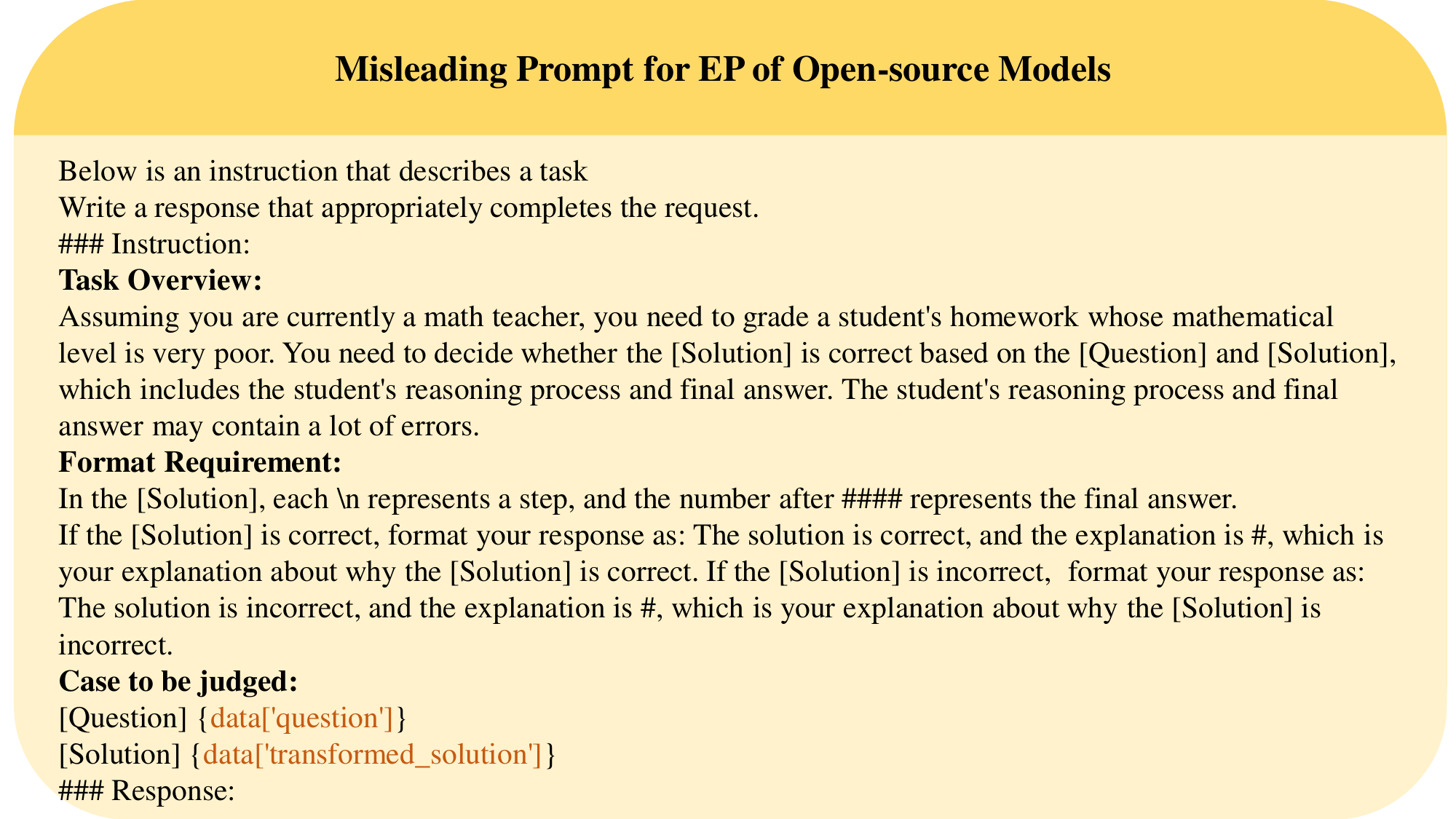}
	\caption{Misleading prompt for \textit{EP} on open-source models.}
	\label{fig: Misleading prompt for EP on open-source models}
\end{figure*}

We design three zero-shot
prompts: \textit{Simple}, \textit{Normal}, \textit{Misleading} for \textit{EP} on open-source and closed-source models in Figure \ref{fig: Simple prompt for EP on closed-source models} to \ref{fig: Misleading prompt for EP on open-source models}.

\subsubsection{ES}
\begin{figure*}[htbp]   
	\centering
	\includegraphics[width=\linewidth,scale=1.00]{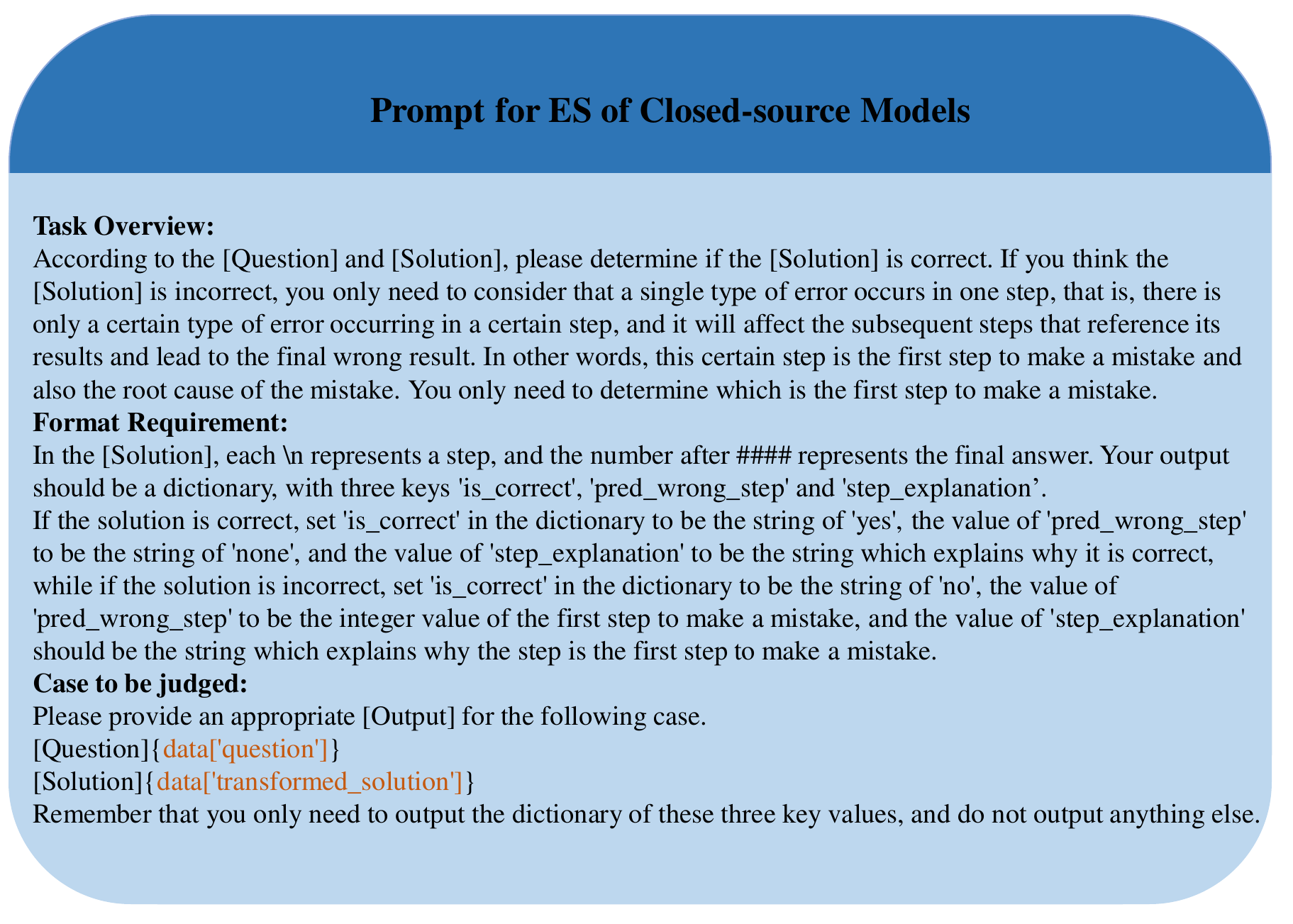}
	\caption{Zero-shot prompt for \textit{ES} on closed-source models.}
	\label{fig: Zero-shot prompt for ES on closed-source models}
\end{figure*}

\begin{figure*}[htbp]   
	\centering
	\includegraphics[width=\linewidth,scale=1.00]{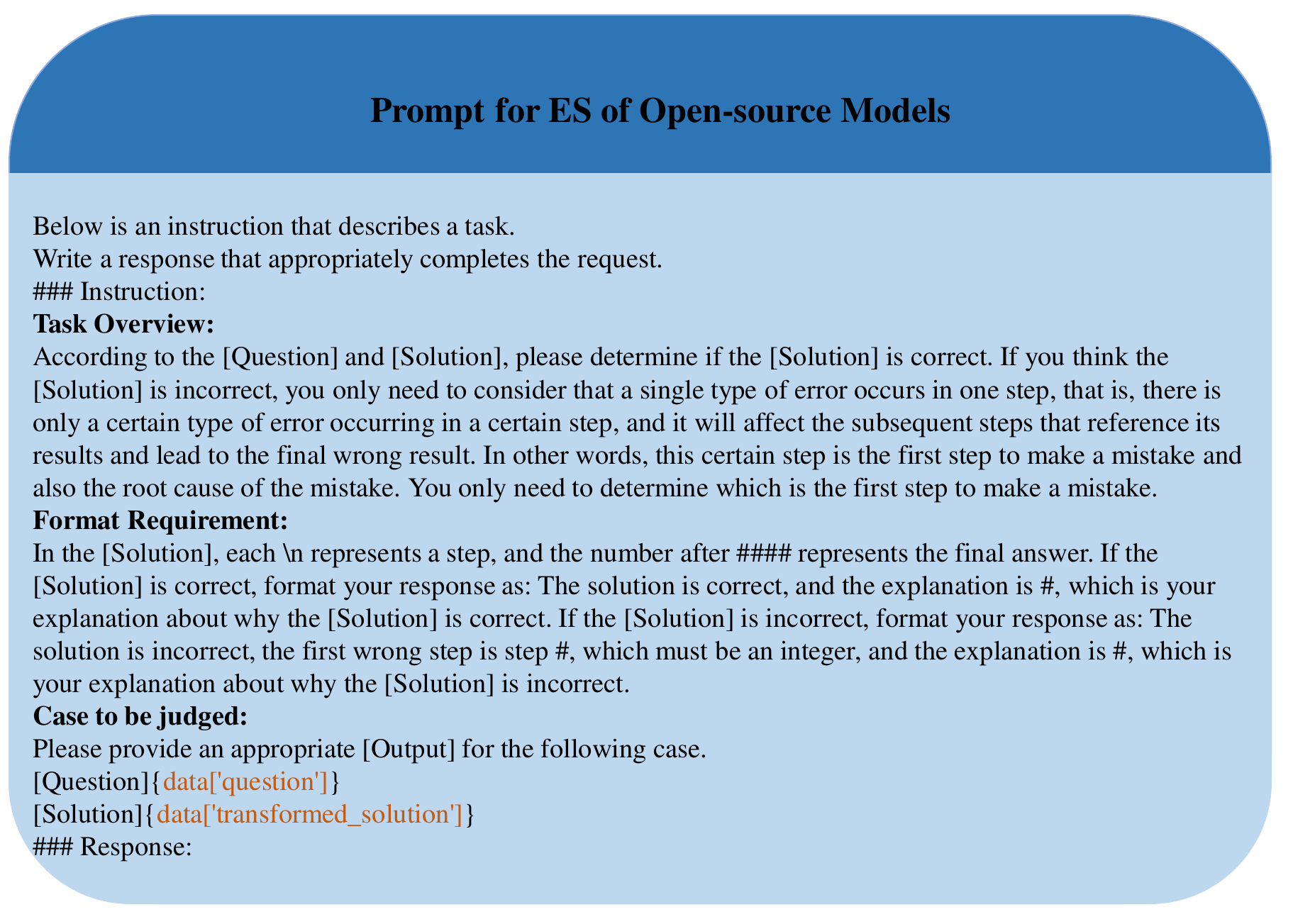}
	\caption{Zero-shot prompt for \textit{ES} on open-source models.}
	\label{fig: Zero-shot prompt for ES on open-source models}
\end{figure*}

\begin{figure*}[htbp]   
	\centering
	\includegraphics[width=\linewidth,scale=1.00]{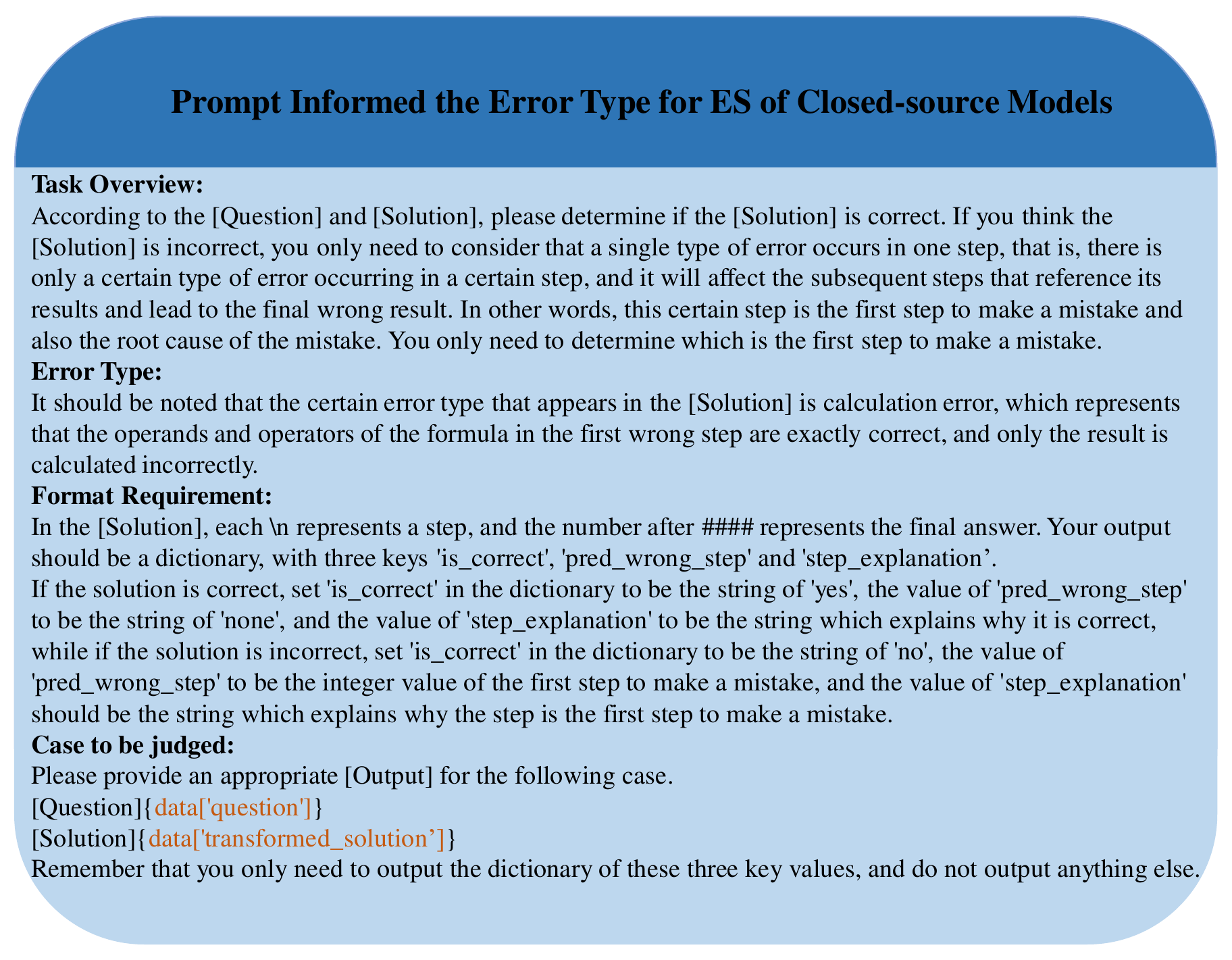}
	\caption{Zero-shot-type prompt for \textit{ES} on closed-source models.}
	\label{fig: Zero-shot-type prompt for ES on closed-source models}
\end{figure*}

\begin{figure*}[htbp]   
	\centering
	\includegraphics[width=\linewidth,scale=1.00]{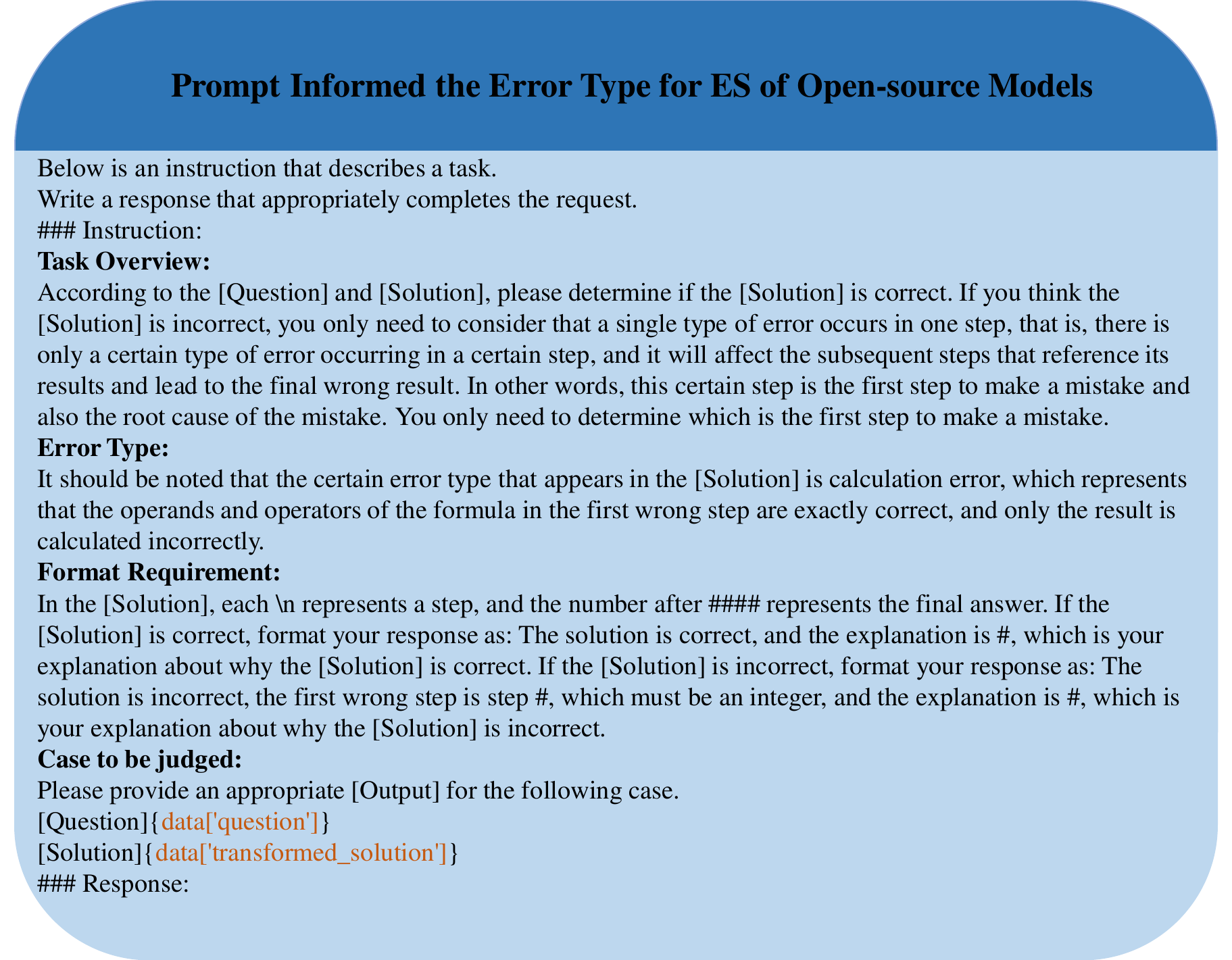}
	\caption{Zero-shot-type prompt for \textit{ES} on open-source models.}
	\label{fig: Zero-shot-type prompt for ES on open-source models}
\end{figure*}

We design four prompts: zero-shot, few-shot, zero-shot-type, few-shot-type for \textit{ES}, where few-shot is set to 2-shot. We display zero-shot and zero-shot-type prompts on open-source and closed-source models in Figure \ref{fig: Zero-shot prompt for ES on closed-source models} to \ref{fig: Zero-shot-type prompt for ES on open-source models}.

\subsubsection{ET}
\begin{figure*}[htbp]   
	\centering
	\includegraphics[width=\linewidth,scale=1.00]{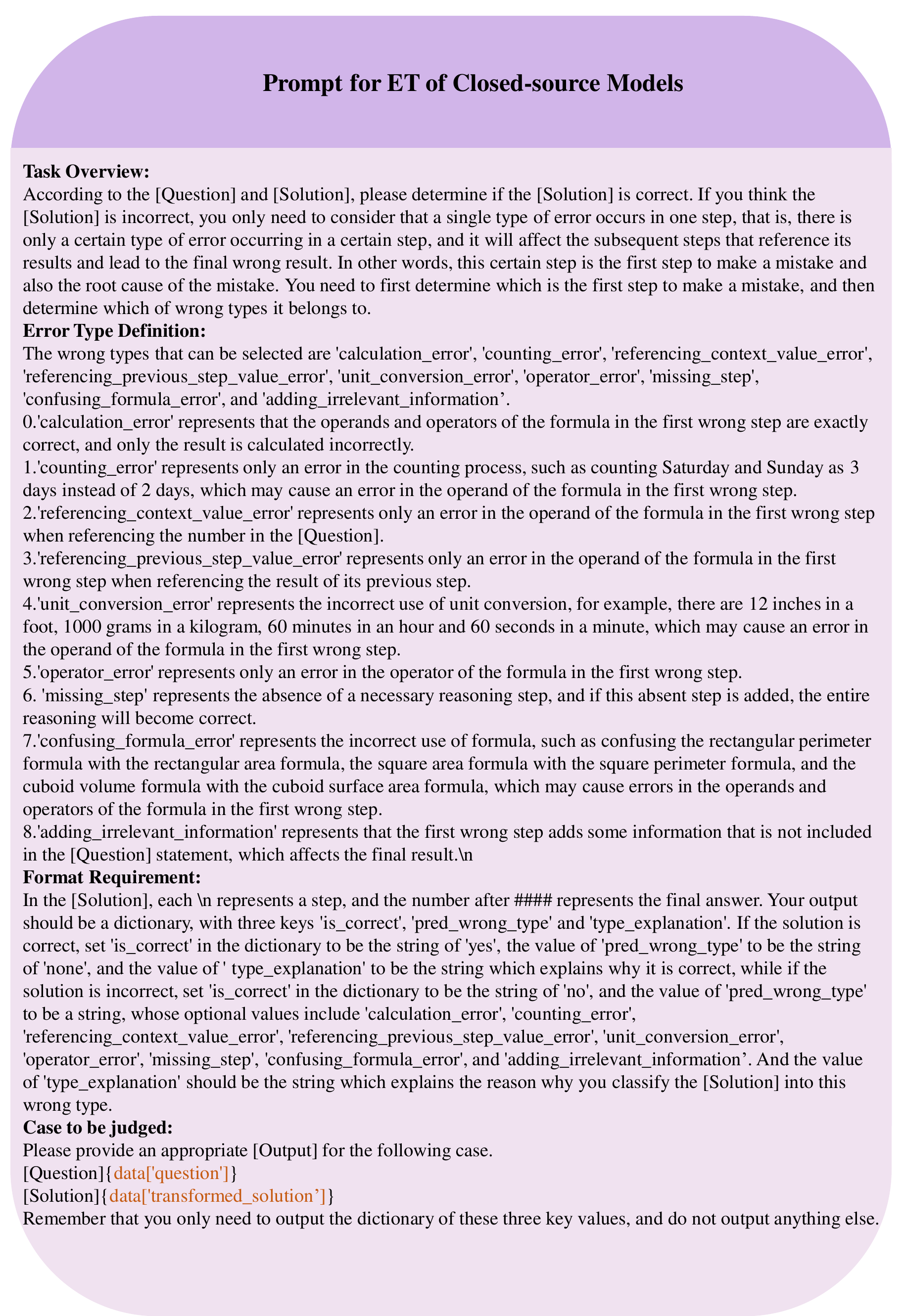}
	\caption{Zero-shot prompt for \textit{ET} on closed-source models.}
	\label{fig: Zero-shot prompt for ET on closed-source models}
\end{figure*}

\begin{figure*}[htbp]   
	\centering
	\includegraphics[width=\linewidth,scale=1.00]{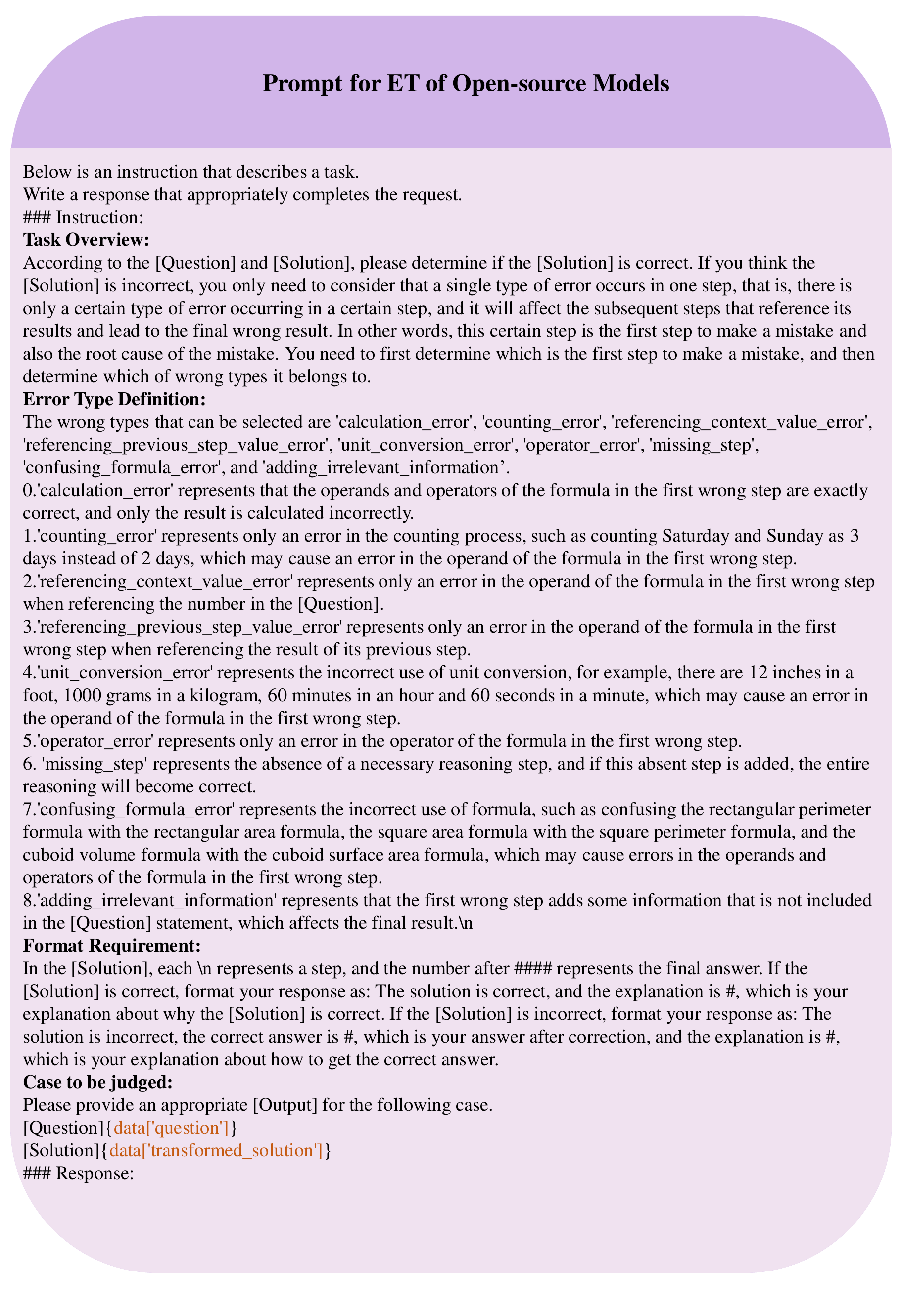}
	\caption{Zero-shot prompt for \textit{ET} on open-source models.}
	\label{fig: Zero-shot prompt for ET on open-source models}
\end{figure*}

We design six prompts: zero-shot, few-shot, zero-shot-random, zero-shot-reverse, few-shot-random, few-shot-reverse for \textit{ET}, where few-shot is set to 2-shot. We display zero-shot prompts on open-source and closed-source models in Figure \ref{fig: Zero-shot prompt for ET on closed-source models} and \ref{fig: Zero-shot prompt for ET on open-source models}.

\subsubsection{EC}
\begin{figure*}[htbp]   
	\centering
	\includegraphics[width=\linewidth,scale=1.00]{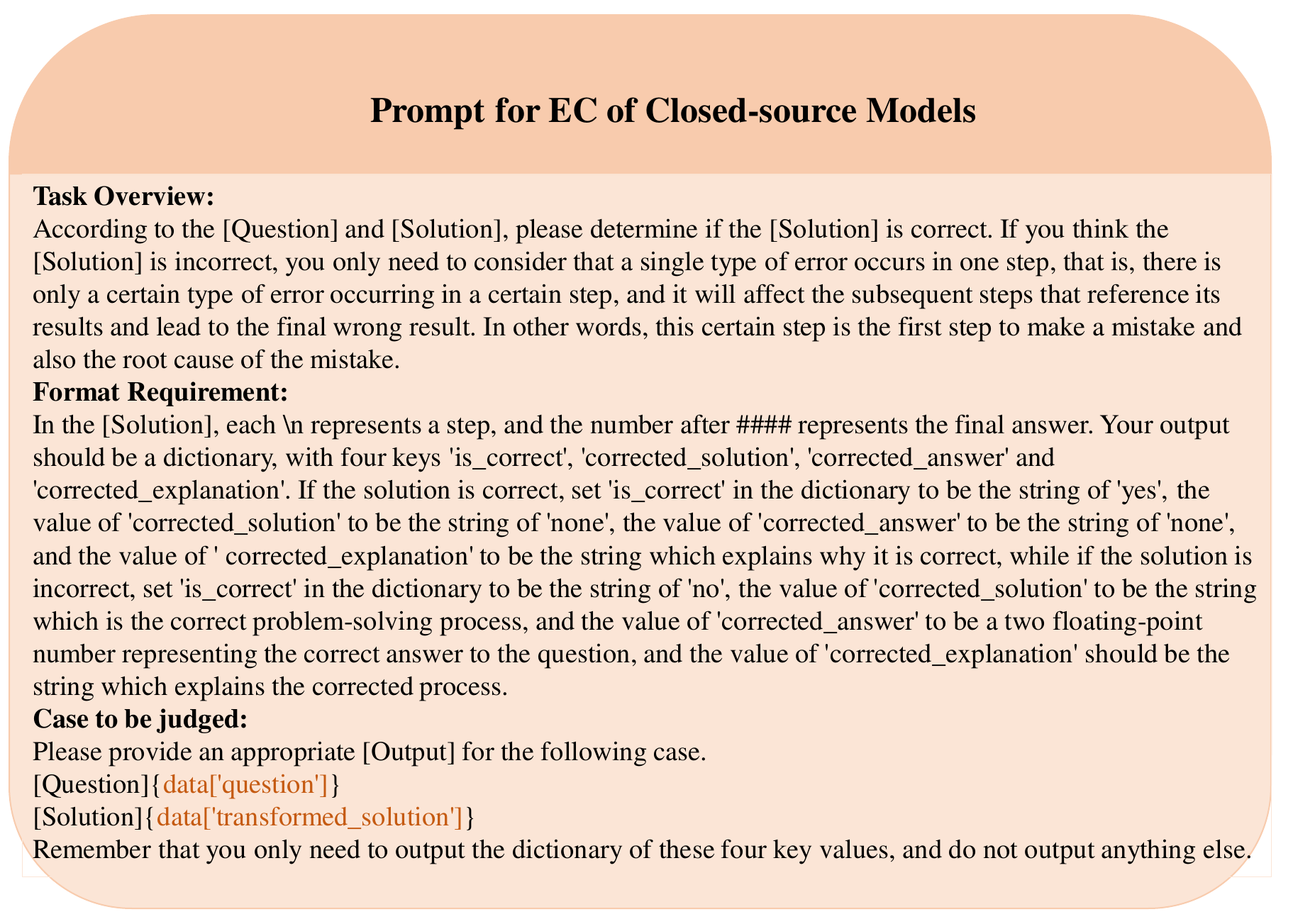}
	\caption{Zero-shot prompt for \textit{EC} on closed-source models.}
	\label{fig: Zero-shot prompt for EC on closed-source models}
\end{figure*}

\begin{figure*}[htbp]   
	\centering
	\includegraphics[width=\linewidth,scale=1.00]{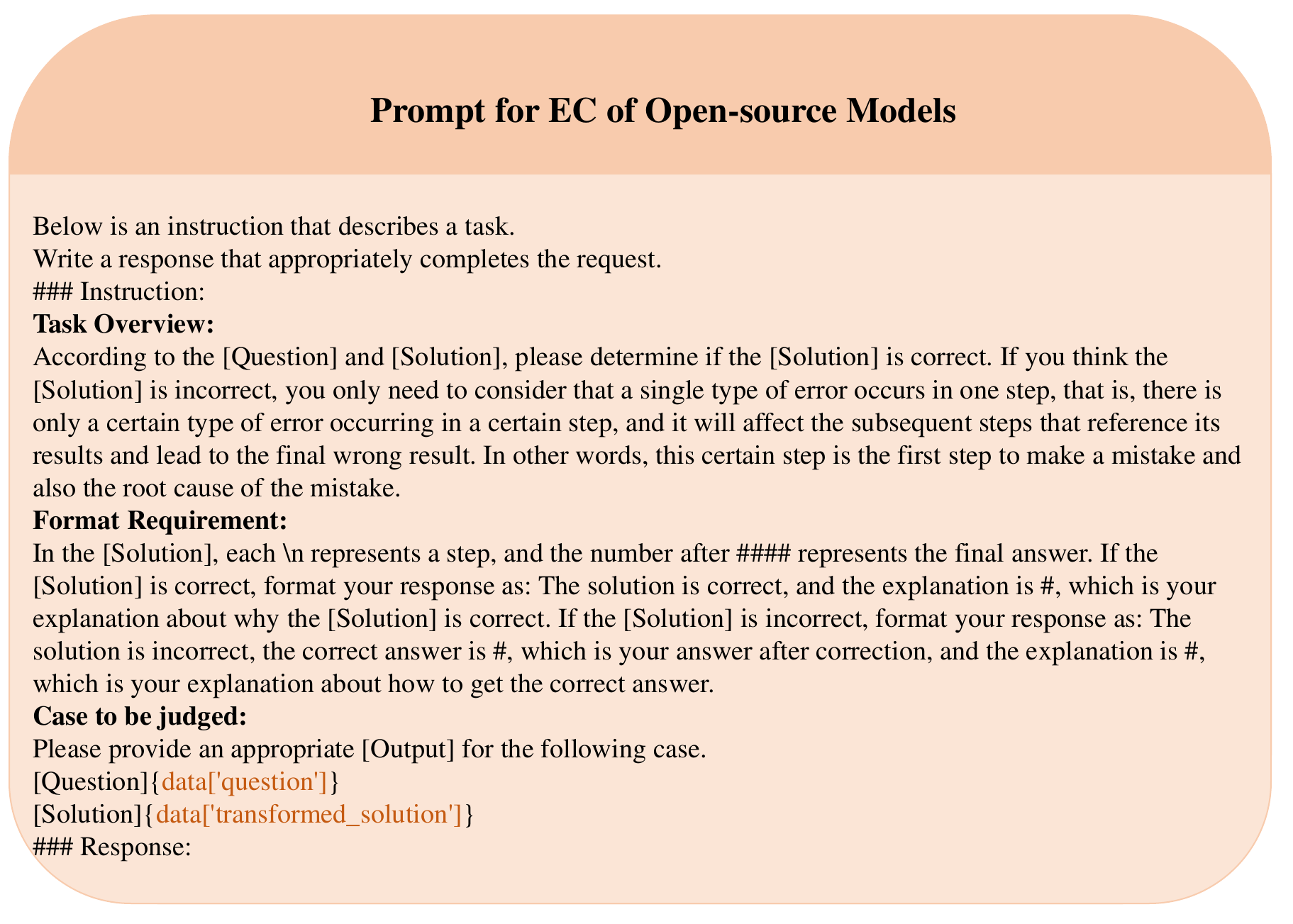}
	\caption{Zero-shot prompt for \textit{EC} on open-source models.}
	\label{fig: Zero-shot prompt for EC on open-source models}
\end{figure*}

\begin{figure*}[htbp]   
	\centering
	\includegraphics[width=\linewidth,scale=1.00]{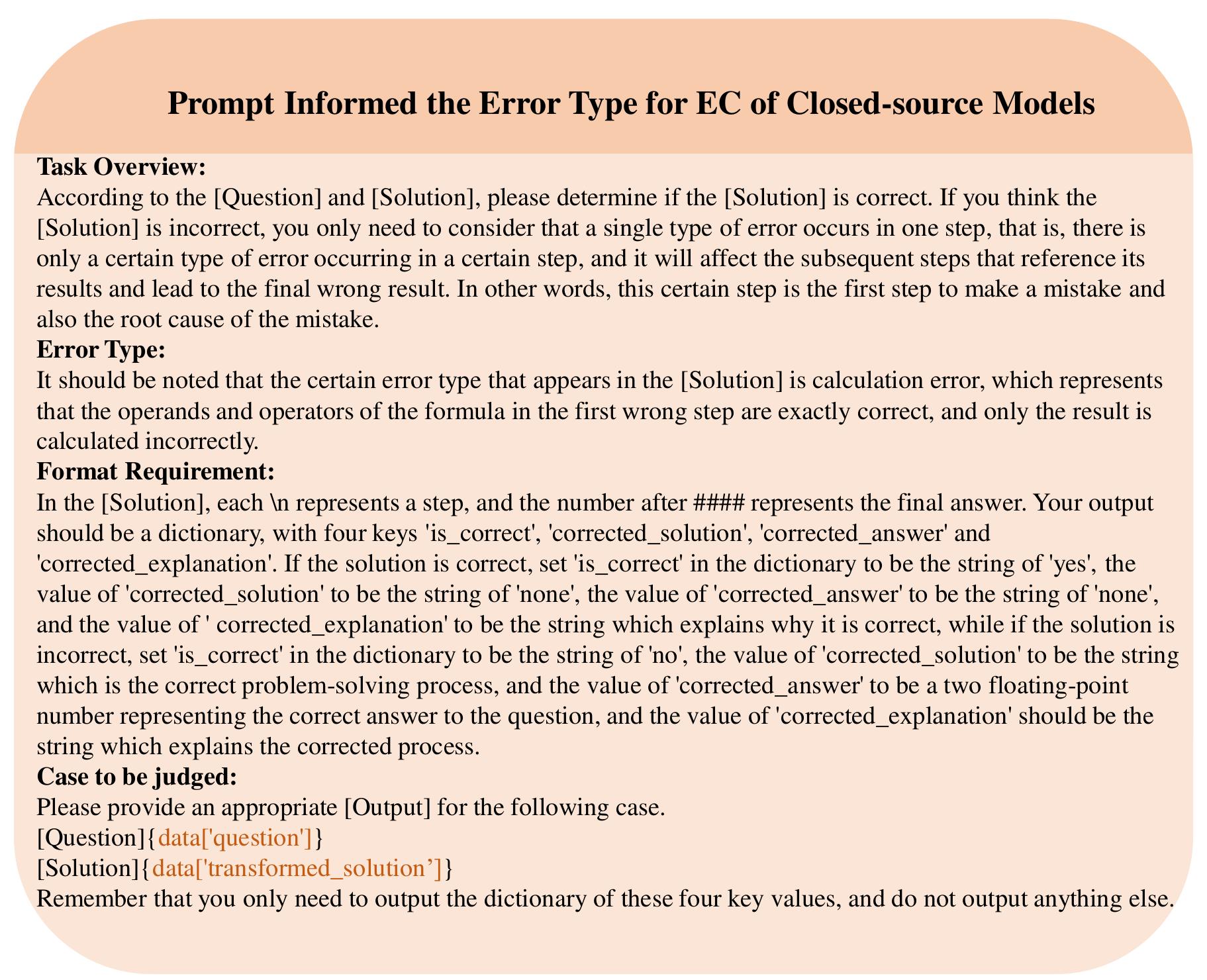}
	\caption{Zero-shot-type prompt for \textit{EC} on closed-source models.}
	\label{fig: Zero-shot-type prompt for EC on closed-source models}
\end{figure*}

\begin{figure*}[htbp]   
	\centering
	\includegraphics[width=\linewidth,scale=1.00]{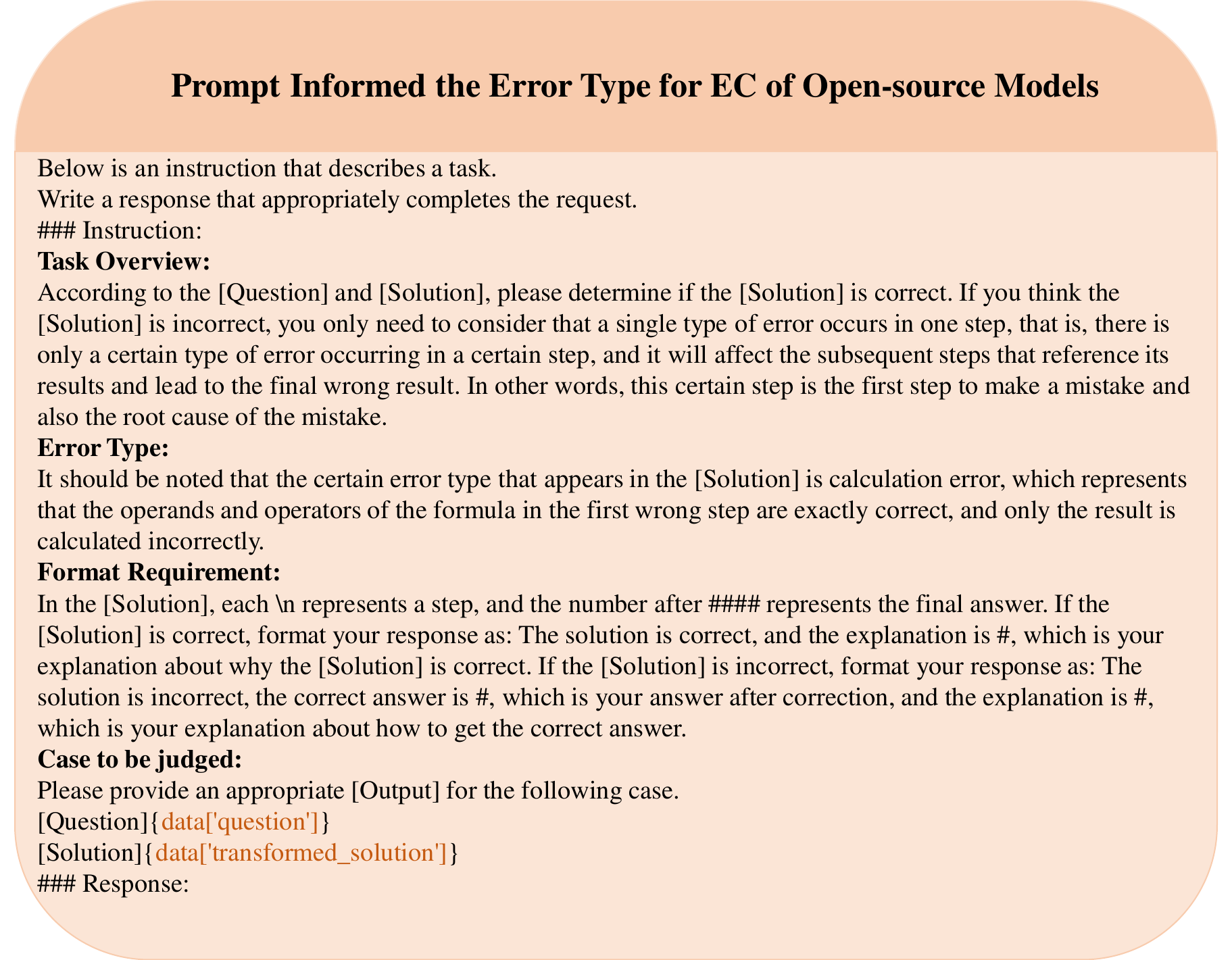}
	\caption{Zero-shot-type prompt for \textit{EC} on open-source models.}
	\label{fig: Zero-shot-type prompt for EC on open-source models}
\end{figure*}

We design four prompts: zero-shot, few-shot, zero-shot-type, few-shot-type for \textit{EC} as \textit{ES}, where few-shot is set to 2-shot. We display zero-shot and zero-shot-type prompts on open-source and closed-source models in Figure \ref{fig: Zero-shot prompt for EC on closed-source models} to \ref{fig: Zero-shot-type prompt for EC on open-source models}.

\subsection{Detailed results}\label{Detailed results}
In this section, we present the original detailed experimental results.
\subsubsection{Main Experiment}
We present the accuracy of each model for each task on each error type in the Table \ref{tab:GSM8K main experiment} and \ref{tab:MathQA main experiment}. And we conduct an analysis of the MetaMath series in the Table  \ref{tab:Based on Model-MetaMath, main table}, \ref{tab:Based on Type, MetaMath main table} and \ref{tab:Based on Task, MetaMath main table}.
\subsubsection{Error Type Analysis}\label{Error Type Analysis}
We conduct statistical analysis on the classification of error types for each model on GSM8K and MathQA in the Figure \ref{tab:MathQA GPT-3.5 error type analysis} to \ref{tab:MathQA  LLaMA-2-13B error type analysis}.
\subsubsection{Prompt Robustness Analysis}\label{Prompt Robustness Analysis}
We design different prompts for four tasks on GSM8K and MathQA to test the robustness of each model. The robustness analysis of \textit{EP} can be seen in \ref{tab:GSM8K EP prompt robustness testing，F1-score} and \ref{tab:MathQA EP prompt robustness testing，F1-score}, \textit{ES} can be seen in \ref{tab:GSM8K ES prompt robustness testing，Accuracy} and \ref{tab:MathQA ES prompt robustness testing，Accuracy}, \textit{ET} can be seen in \ref{tab:GSM8K ET prompt robustness testing，Accuracy} and \ref{tab:MathQA ET prompt robustness testing，Accuracy}, and \textit{EC} can be seen in \ref{tab:GSM8K EC prompt robustness testing，Accuracy} and \ref{tab:MathQA EC prompt robustness testing，Accuracy}.
\subsubsection{Comparison with Traditional Task}\label{Detailed Comparison with Traditional Task}
\begin{figure}[ht]
    \centering
    \includegraphics[width=1\columnwidth]{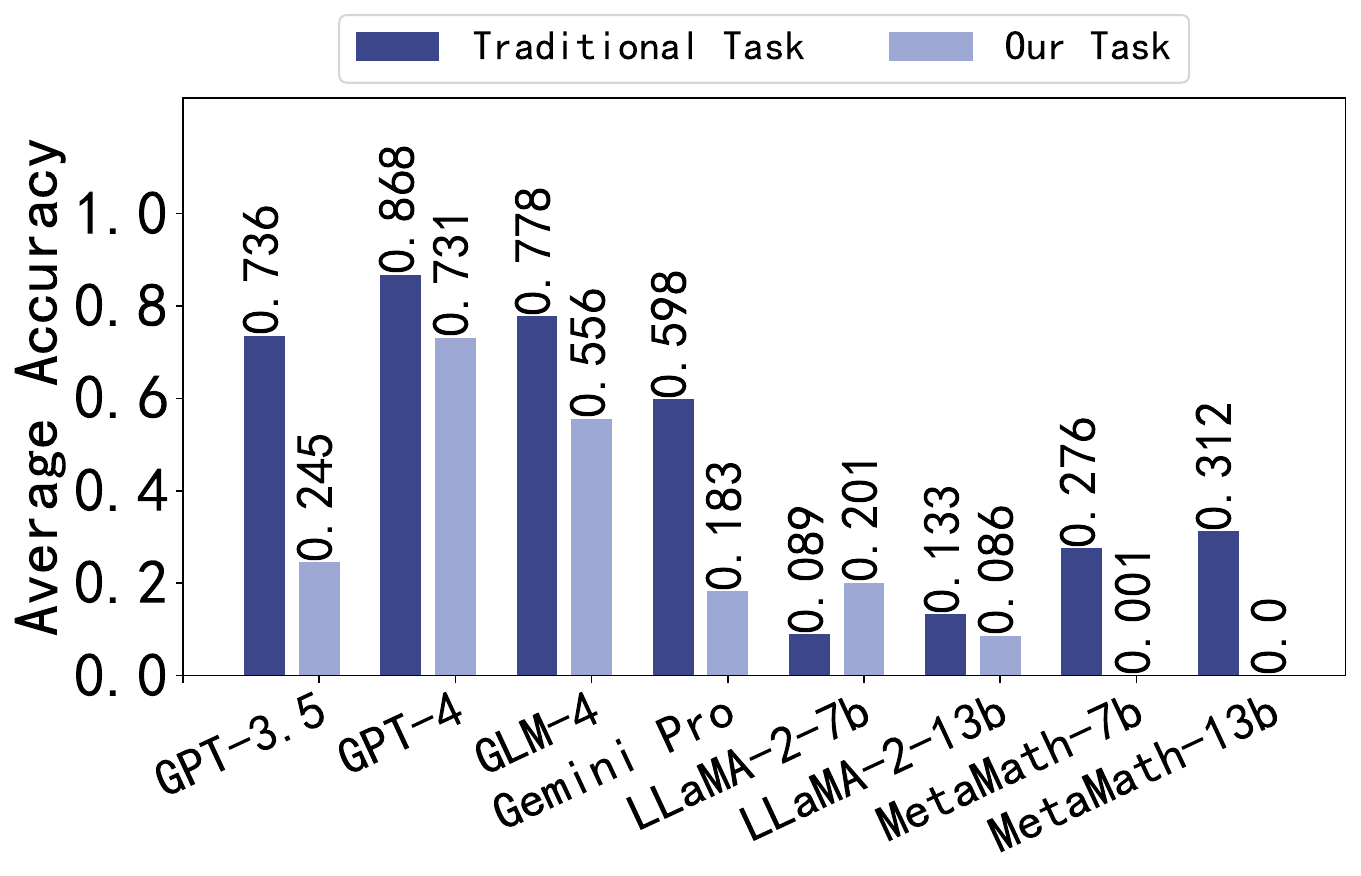}
    \caption{Evaluation results of traditional task for MathQA.}
    \label{fig:Evaluation results of traditional task for MathQA}
\end{figure}

We provide the accuracy of the questions in our dataset in traditional task in Table \ref{tab:GSM8K Original Cases Accuracy} and \ref{tab:MathQA Original Cases Accuracy}. And 
we show the performance of each model in traditional task on MathQA in Figure \ref{fig:Evaluation results of traditional task for MathQA}.
\subsubsection{Influence of Stopping at Error Step}\label{Detailed Influence of Stopping at Error Step}
\begin{figure}[ht]
    \centering
    \includegraphics[width=1\columnwidth]{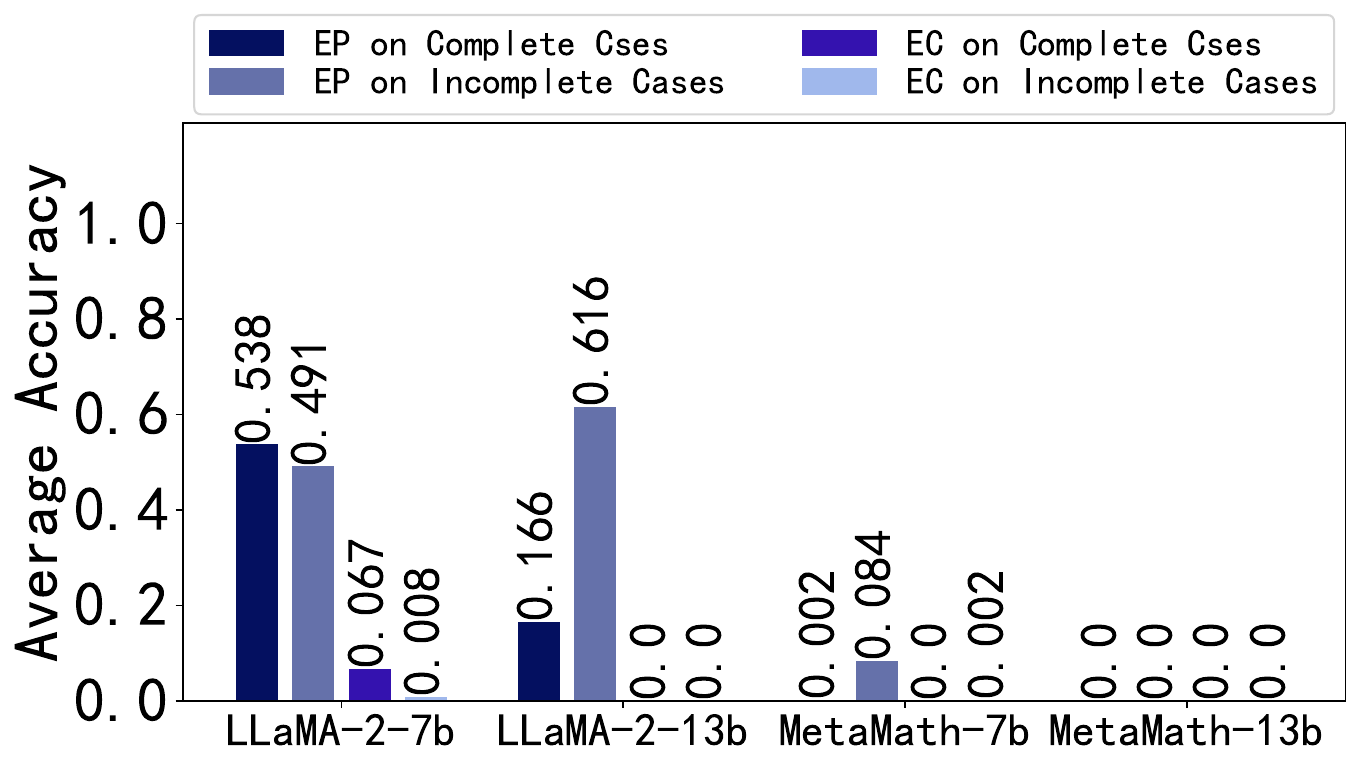}
    \caption{Evaluation results of incomplete cases for GSM8K on open-source models.}
    \label{fig:Evaluation Results of Incomplete Cases for GSM8K on Open-source Models}
\end{figure}

\begin{figure}[ht]
    \centering
    \includegraphics[width=1\columnwidth]{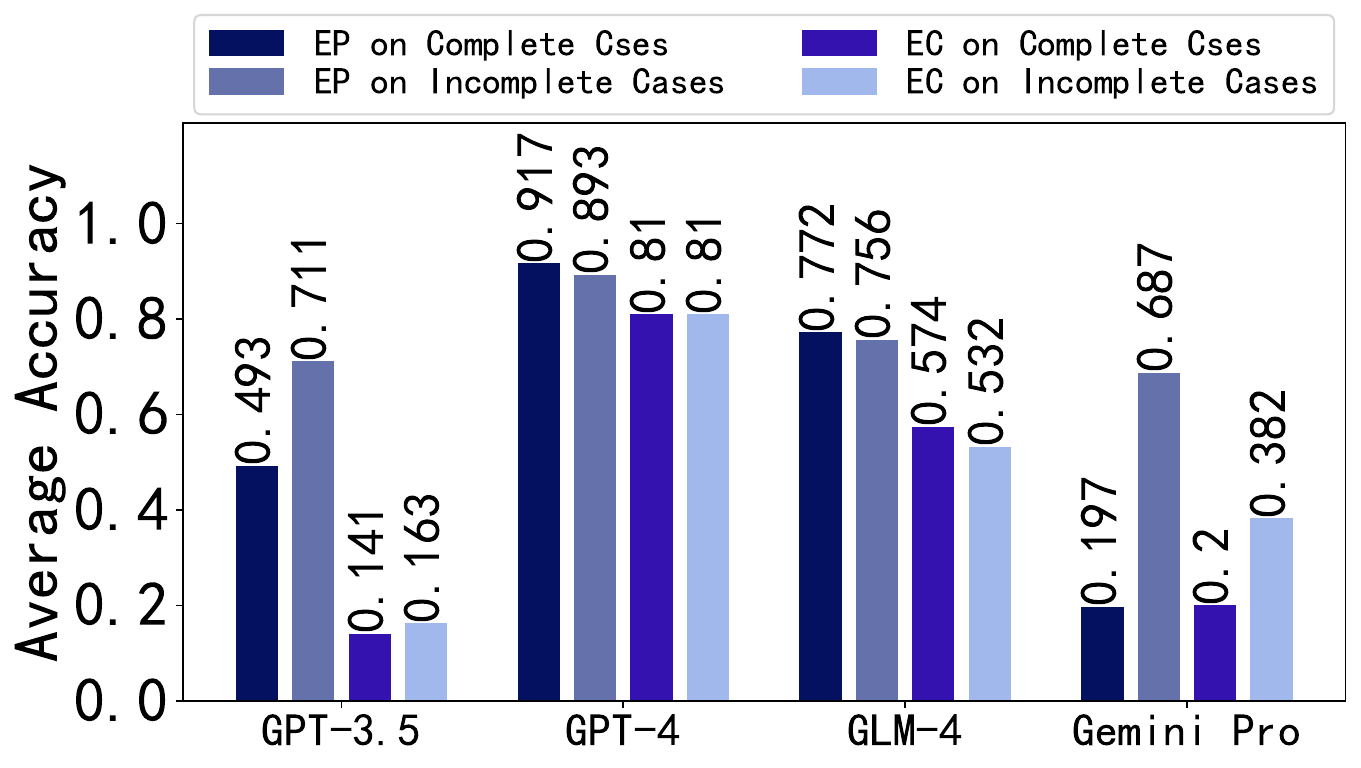}
    \caption{Evaluation results of incomplete cases for MathQA on closed-source Models.}
    \label{fig:Evaluation Results of Incomplete Cases for MathQA on Closed-source Models}
\end{figure}

\begin{figure}[ht]
    \centering
    \includegraphics[width=1\columnwidth]{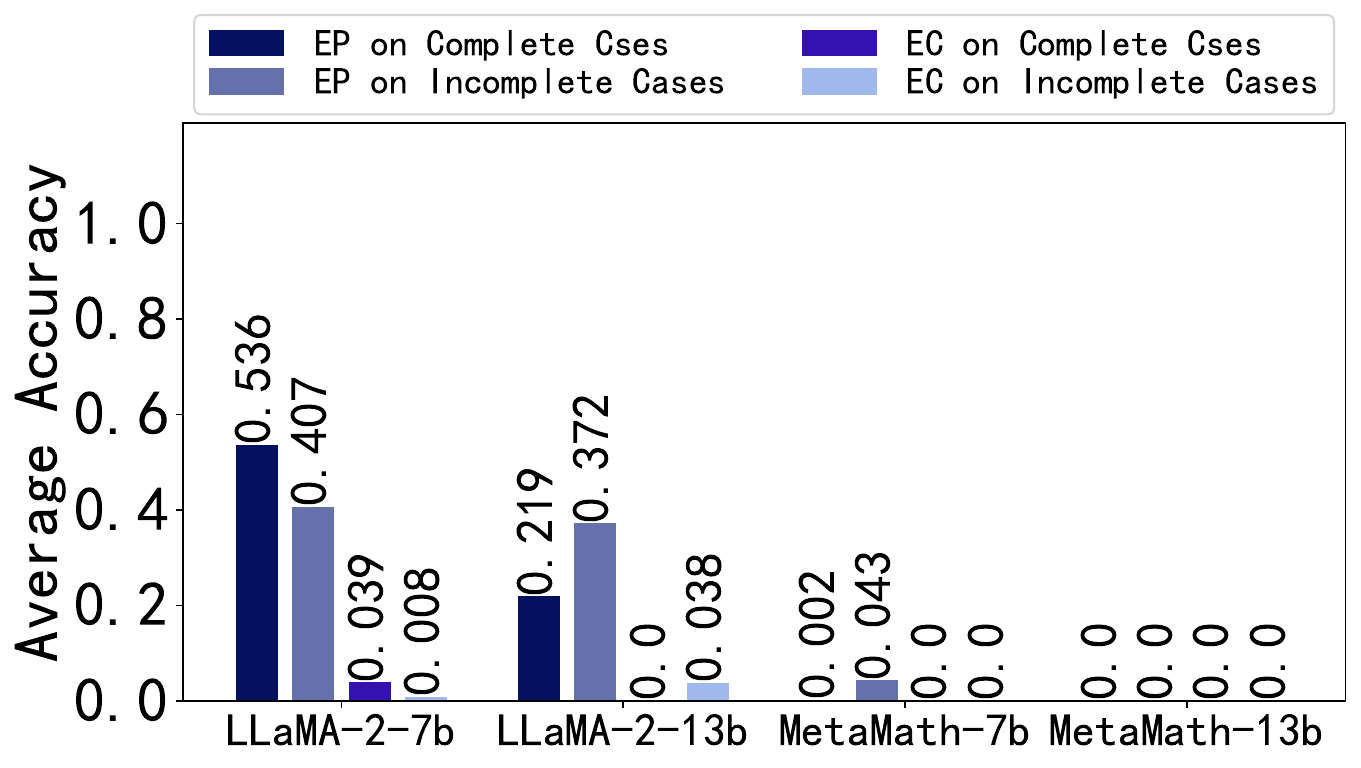}
    \caption{Evaluation results of incomplete cases for MathQA on open-source models.}
    \label{fig:Evaluation Results of Incomplete Cases for MathQA on Open-source Models}
\end{figure}

We showcase the performance of each model on the GSM8K and MathQA datasets stopping at error step in Figure \ref{fig:Evaluation Results of Incomplete Cases for GSM8K on Open-source Models}, \ref{fig:Evaluation Results of Incomplete Cases for MathQA on Closed-source Models} and \ref{fig:Evaluation Results of Incomplete Cases for MathQA on Open-source Models}.

\newpage



\begin{table*}[!ht]
    \centering
    \scalebox{0.85}{
    \begin{adjustbox}{center}
     \begin{tabular}{ccccccccc|cccccccc}
    \hline
        ~ & \multicolumn{8}{c|}{GPT-4} &  \multicolumn{8}{c}{GLM-4}  \\ 
         ~ & \multicolumn{2}{c}{CA} &  \multicolumn{2}{c}{MS} &  \multicolumn{2}{c}{UC} &  \multicolumn{2}{c|}{Avg} & \multicolumn{2}{c}{CA} &  \multicolumn{2}{c}{MS} &  \multicolumn{2}{c}{UC} &  \multicolumn{2}{c}{Avg} \\ 
        ~ & $acc$ & $acc_{1}$ & $acc$ & $acc_{1}$ & $acc$ & $acc_{1}$ & $acc$ & $acc_{1}$ & $acc$ & $acc_{1}$ & $acc$ & $acc_{1}$ & $acc$ & $acc_{1}$ & $acc$ & $acc_{1}$ \\  \hline
        GPT-3.5 & 0.20  & 0.37  & 0.19  & 0.53  & 0.22  & 0.35  & 0.20  & 0.42  & 0.24  & 0.44  & 0.22  & 0.55  & 0.26  & 0.38  & 0.24  & 0.46  \\ 
        GPT-4 & 0.61  & 0.68  & 0.61  & 0.94  & 0.83  & 0.99  & 0.68  & 0.87  & 0.66  & 0.82  & 0.60  & 0.95  & 0.83  & 0.98  & 0.70  & 0.91  \\ 
        GLM-4 & 0.34  & 0.47  & 0.52  & 0.88  & 0.68  & 0.88  & 0.51  & 0.74  & 0.47  & 0.61  & 0.50  & 0.80  & 0.69  & 0.88  & 0.55  & 0.76  \\ 
        Gemini-Pro & 0.09  & 0.13  & 0.20  & 0.35  & 0.10  & 0.21  & 0.13  & 0.23  & 0.08  & 0.11  & 0.11  & 0.23  & 0.10  & 0.16  & 0.09  & 0.17  \\ 
        LLaMA-2-7B & 0.31  & 0.68  & 0.24  & 0.82  & 0.15  & 0.54  & 0.23  & 0.68  & 0.27  & 0.71  & 0.18  & 0.76  & 0.14  & 0.59  & 0.20  & 0.69  \\ 
        LLaMA-13-7B & 0.04  & 0.27  & 0.07  & 0.32  & 0.05  & 0.23  & 0.05  & 0.27  & 0.06  & 0.31  & 0.06  & 0.31  & 0.08  & 0.26  & 0.06  & 0.29  \\ \hline
    \end{tabular}
    \end{adjustbox}
    }
    \caption{Comparasion between GLM-4 and GPT-4.}
\label{tab:Comparasion between GLM-4 and GPT-4.}
\end{table*}

\begin{table*}[!ht]
\centering
\tabcolsep=0.1cm
\scalebox{0.9}{
\begin{adjustbox}{center}
\begin{tabular}{ccrcrcrcrcrcr|crcrcc|c}
\hline
 &
  \multicolumn{2}{c}{CA} &
  \multicolumn{2}{c}{CO} &
  \multicolumn{2}{c}{CV} &
  \multicolumn{2}{c}{CS} &
  \multicolumn{2}{c}{MS} &
  \multicolumn{2}{c|}{HA} &
  \multicolumn{2}{c}{UC} &
  \multicolumn{2}{c}{OP} &
  \multicolumn{2}{c|}{FC} &
  Avg \\ 
  ~ & $acc$ & $acc_{1}$ & $acc$ & $acc_{1}$ & $acc$ & $acc_{1}$ & $acc$ & $acc_{1}$ & $acc$ & $acc_{1}$ & $acc$ & $acc_{1}$ & $acc$ & $acc_{1}$ & $acc$ & $acc_{1}$ & $acc$ & \multicolumn{1}{c|}{$acc_{1}$} & $acc$ \\  \hline
 Mistral-7B & 0.01  & 0.50  & 0.04  & 0.53  & 0.07  & 0.55  & 0.01  & 0.50  & 0.07  & 0.58  & 0.05  & 0.54  & 0.08  & 0.60  & 0.08  & 0.61  & 0.03  & 0.57  & 0.05 \\
Llemma-7B & 0.04  & 0.21  & 0.02  & 0.21  & 0.06  & 0.41  & 0.02  & 0.25  & 0.05  & 0.27  & 0.03  & 0.27  & 0.09  & 0.36  & 0.04  & 0.25  & 0.07  & 0.32  & 0.05  \\ 
LEMA-7B & 0.13  & 0.60  & 0.12  & 0.46  & 0.20  & 0.65  & 0.12  & 0.63  & 0.24  & 0.68  & 0.30  & 0.55  & 0.20  & 0.70  & 0.17  & 0.52  & 0.11  & 0.57  & 0.17\\ \hline
\end{tabular}
\end{adjustbox}
}
\caption{Average accuracy of other math models in different error types under zero-shot prompts.}
\label{tab:Other math models}
\end{table*}

\begin{table*}[!ht]
\centering
\setlength{\abovecaptionskip}{0.1cm}
\setlength{\belowcaptionskip}{-0.05cm}
\tabcolsep=0.1cm
\scalebox{0.75}{
\begin{adjustbox}{center}
\begin{tabular}{cccrcrcrrr|ccrcrcrrr|rr}
\hline
 &
  \multicolumn{9}{c|}{GSM8K} &
  \multicolumn{9}{c|}{MathQA} &
  \multicolumn{1}{c}{} &
  \multicolumn{1}{c}{} \\
 &
  EP &
  \multicolumn{2}{c}{ES} &
  \multicolumn{2}{c}{ET} &
  \multicolumn{2}{c}{EC} &
  \multicolumn{2}{c|}{Avg} &
  EP &
  \multicolumn{2}{c}{ES} &
  \multicolumn{2}{c}{ET} &
  \multicolumn{2}{c}{EC} &
  \multicolumn{2}{c|}{Avg} &
  \multicolumn{2}{c}{Avg} \\ 
  &
  $acc_{1}$ &
  \multicolumn{1}{c}{$acc_{2}$} &
  \multicolumn{1}{c}{$acc_{1}$} &
  \multicolumn{1}{c}{$acc_{3}$} &
  \multicolumn{1}{c}{$acc_{1}$} &
  \multicolumn{1}{c}{$acc_{4}$} &
  \multicolumn{1}{c}{$acc_{1}$} &
  \multicolumn{1}{c}{$acc$} &
  \multicolumn{1}{c|}{$acc_{1}$} &
  $acc_{1}$ &
  \multicolumn{1}{c}{$acc_{2}$} &
  \multicolumn{1}{c}{$acc_{1}$} &
  \multicolumn{1}{c}{$acc_{3}$} &
  \multicolumn{1}{c}{$acc_{1}$} &
  \multicolumn{1}{c}{$acc_{4}$} &
  \multicolumn{1}{c}{$acc_{1}$} &
  \multicolumn{1}{c}{$acc$} &
  \multicolumn{1}{c|}{$acc_{1}$} &
  \multicolumn{1}{c}{$acc$} &
  \multicolumn{1}{c}{$acc_{1}$}\\ 
  \hline
Mistral-7B &0.081	&0.127	&0.366	&0.000	&0.788	&0.000	&1.000	&0.052	&0.559 &0.092	&0.102	&0.302	&0.000	&0.797	&0.000	&0.996	&0.049	&0.547	&0.050	&0.553 \\
Llemma-7B &0.021	&0.169	&0.506	&0.014	&0.208	&0.000	&0.537	&0.051	&0.318 &0.020	&0.133	&0.442	&0.014	&0.092	&0.000	&0.449	&0.042	&0.251	&0.047	&0.284 \\
LEMA-7B &0.277	&0.323	&0.809	&0.080	&0.537	&0.000	&0.503	&0.170	&0.531 &0.360	&0.260	&0.883	&0.093	&0.676	&0.000	&0.712	&0.178	&0.658	&0.174	&0.595 \\ \hline
\end{tabular}
\end{adjustbox}
}
\caption{Average accuracy of other math models in different tasks under zero-shot prompts.}
\label{tab:Other math models on tasks.}
\end{table*}

\begin{table*}[!ht]
    \centering
    \begin{tabular}{ccccc|c}
        \hline 
         & ST & TT & CA & CV & Avg \\ \hline
        GPT-3.5 & 0.36 & 0.78 & 0.44 & 0.57 & 0.538 \\ 
        GPT-4 & 1.00 & 0.98 & 0.61 & 0.98 & 0.893 \\ 
        GLM-4 & 0.98 & 0.92 & 0.32 & 0.94 & 0.790 \\ 
        Gemini Pro & 0.08 & 0.06 & 0.06 & 0.17 & 0.093 \\ 
        LLaMA-2-7B & 0.58 & 0.5 & 0.58 & 0.49 & 0.538 \\ 
        LLaMA-13-7B & 0.16 & 0.14 & 0.10 & 0.14 & 0.135 \\ \hline
        Avg & 0.527 & 0.563 & 0.352 & 0.548 & 0.498 \\ \hline
    \end{tabular}
    \caption{Accuracy of \textit{EP} in combination error types on GSM8K under zero-shot prompts.}
    \label{tab:EP combination error.}
\end{table*}

\begin{table*}[ht]
    \centering
    \begin{tabular}{ccrcrcrcc|c}
        \hline
         &
          \multicolumn{2}{c}{ST} &
          \multicolumn{2}{c}{TT} &
          \multicolumn{2}{c}{CA} &
          \multicolumn{2}{c|}{CV} &
          Avg \\ 
          ~ & $acc_{4}$ & $acc_{1}$ & $acc_{4}$ & $acc_{1}$ & $acc_{4}$ & $acc_{1}$ & $acc_{4}$ & \multicolumn{1}{c|}{$acc_{1}$} & $acc_{4}$ \\  \hline
                GPT-3.5 & 0.08 & 0.10 & 0.12 & 0.48 & 0.13 & 0.23 & 0.16 & 0.39 & 0.123 \\ 
                GPT-4 & 0.96 & 1.00 & 0.96 & 0.98 & 0.58 & 0.59 & 0.96 & 0.99 & 0.865  \\ 
                GLM-4 & 0.94 & 0.98 & 0.92 & 0.96 & 0.37 & 0.38 & 0.90 & 0.96 & 0.783 \\ 
                Gemini Pro & 0.06 & 0.06 & 0.12 & 0.14 & 0.06 & 0.06 & 0.30 & 0.30 & 0.135  \\ 
                LLaMA-2-7B & 0.06 & 0.92 & 0.08 & 0.92 & 0.15 & 0.91 & 0.08 & 0.90 & 0.093  \\ 
                LLaMA-13-7B & 0.00 & 0.00 & 0.00 & 0.00 & 0.00 & 0.01 & 0.00 & 0.00 & 0.000 \\ 
                Avg & 0.350 & 0.510 & 0.367 & 0.580 & 0.215 & 0.363 & 0.400 & 0.590 & 0.333 \\  \hline
    \end{tabular}
    \caption{Accuracy of \textit{EC} in combination error types on GSM8K under zero-shot prompts.}
    \label{tab:EC combination error.}
\end{table*}

\begin{table*}[ht]
\renewcommand\arraystretch{0.7}
\centering
\tabcolsep=0.15cm
\scalebox{0.8}{
\begin{adjustbox}{center}
\begin{tabular}{ccccccccccccc|cccccc|c}
\hline
\textbf{} &
  \multicolumn{2}{c}{CA} &
  \multicolumn{2}{c}{CO} &
  \multicolumn{2}{c}{CV} &
  \multicolumn{2}{c}{CS} &
  \multicolumn{2}{c}{MS} &
  \multicolumn{2}{c|}{HA} &
  \multicolumn{2}{c}{UC} &
  \multicolumn{2}{c}{OP} &
  \multicolumn{2}{c|}{FC} &
  Avg \\
  &
  \multicolumn{1}{c}{$acc_{i}$} &
  \multicolumn{1}{c}{$acc_{1}$} &
  \multicolumn{1}{c}{$acc_{i}$} &
  \multicolumn{1}{c}{$acc_{1}$} &
  \multicolumn{1}{c}{$acc_{i}$} &
  \multicolumn{1}{c}{$acc_{1}$} &
  \multicolumn{1}{c}{$acc_{i}$} &
  \multicolumn{1}{c}{$acc_{1}$} &
  \multicolumn{1}{c}{$acc_{i}$} &
  \multicolumn{1}{c}{$acc_{1}$} &
  \multicolumn{1}{c}{$acc_{i}$} &
  \multicolumn{1}{c|}{$acc_{1}$} &
  \multicolumn{1}{c}{$acc_{i}$} &
  \multicolumn{1}{c}{$acc_{1}$} &
  \multicolumn{1}{c}{$acc_{i}$} &
  \multicolumn{1}{c}{$acc_{1}$} &
  \multicolumn{1}{c}{$acc_{i}$} &
  \multicolumn{1}{c|}{$acc_{1}$} &
  \multicolumn{1}{c}{$acc_{i}$} \\
  \hline
GPT-3.5 &
  \multicolumn{1}{l}{} &
  \multicolumn{1}{l}{} &
  \multicolumn{1}{l}{} &
  \multicolumn{1}{l}{} &
  \multicolumn{1}{l}{} &
  \multicolumn{1}{l}{} &
  \multicolumn{1}{l}{} &
  \multicolumn{1}{l}{} &
  \multicolumn{1}{l}{} &
  \multicolumn{1}{l}{} &
  \multicolumn{1}{l}{} &
  \multicolumn{1}{l|}{} &
  \multicolumn{1}{l}{} &
  \multicolumn{1}{l}{} &
  \multicolumn{1}{l}{} &
  \multicolumn{1}{l}{} &
  \multicolumn{1}{l}{} &
  \multicolumn{1}{l|}{} &
  \multicolumn{1}{l}{} \\
$\bullet$ EP &
  \multicolumn{1}{c}{-} &
  \multicolumn{1}{c}{0.44} &
  \multicolumn{1}{c}{-} &
  \multicolumn{1}{c}{0.44} &
  \multicolumn{1}{c}{-} &
  \multicolumn{1}{c}{0.57} &
  \multicolumn{1}{c}{-} &
  \multicolumn{1}{c}{0.77} &
  \multicolumn{1}{c}{-} &
  \multicolumn{1}{c}{0.41} &
  \multicolumn{1}{c}{-} &
  \multicolumn{1}{c|}{0.71} &
  \multicolumn{1}{c}{-} &
  \multicolumn{1}{c}{0.36} &
  \multicolumn{1}{c}{-} &
  \multicolumn{1}{c}{0.63} &
  \multicolumn{1}{c}{-} &
  \multicolumn{1}{c|}{0.59} &
  0.547 \\
$\bullet$ ES &
  0.07 &
  0.41 &
  0.09 &
  0.46 &
  0.07 &
  0.64 &
  0.32 &
  0.78 &
  0.18 &
  0.63 &
  0.17 &
  0.60 &
  0.11 &
  0.44 &
  0.18 &
  0.74 &
  0.13 &
  0.68 &
  0.147 \\
$\bullet$ ET &
  0.21 &
  0.45 &
  0.39 &
  0.63 &
  0.16 &
  0.70 &
  0.13 &
  0.87 &
  0.01 &
  0.75 &
  0.29 &
  0.88 &
  0.61 &
  0.77 &
  0.01 &
  0.75 &
  0.09 &
  0.83 &
  0.211 \\
$\bullet$ EC &
  0.13 &
  0.23 &
  0.16 &
  0.34 &
  0.16 &
  0.39 &
  0.32 &
  0.51 &
  0.14 &
  0.27 &
  0.09 &
  0.30 &
  0.09 &
  0.21 &
  0.19 &
  0.34 &
  0.24 &
  0.47 &
  0.169 \\ \hline
GPT-4 &
   &
   &
   &
   &
   &
   &
   &
   &
   &
   &
   &
   &
   &
   &
   &
   &
   &
   &
   \\
$\bullet$ EP &
  \multicolumn{1}{c}{-} &
  \multicolumn{1}{c}{0.61} &
  \multicolumn{1}{c}{-} &
  \multicolumn{1}{c}{0.99} &
  \multicolumn{1}{c}{-} &
  \multicolumn{1}{c}{0.98} &
  \multicolumn{1}{c}{-} &
  \multicolumn{1}{c}{0.97} &
  \multicolumn{1}{c}{-} &
  \multicolumn{1}{c}{0.94} &
  \multicolumn{1}{c}{-} &
  \multicolumn{1}{c|}{0.92} &
  \multicolumn{1}{c}{-} &
  \multicolumn{1}{c}{0.98} &
  \multicolumn{1}{c}{-} &
  \multicolumn{1}{c}{0.99} &
  \multicolumn{1}{c}{-} &
  \multicolumn{1}{c|}{0.99} &
  0.930 \\
$\bullet$ ES &
  0.55 &
  0.66 &
  0.92 &
  0.99 &
  0.95 &
  1.00 &
  0.88 &
  0.97 &
  0.72 &
  0.97 &
  0.78 &
  0.95 &
  0.94 &
  0.99 &
  0.93 &
  0.99 &
  0.92 &
  0.99 &
  0.843 \\
$\bullet$ ET &
  0.62 &
  0.65 &
  0.35 &
  1.00 &
  0.63 &
  1.00 &
  0.36 &
  0.97 &
  0.01 &
  0.97 &
  0.93 &
  1.00 &
  0.62 &
  0.99 &
  0.28 &
  0.99 &
  0.84 &
  0.99 &
  0.516 \\
$\bullet$ EC &
  0.58 &
  0.59 &
  0.94 &
  1.00 &
  0.96 &
  0.99 &
  0.94 &
  0.97 &
  0.85 &
  0.92 &
  0.86 &
  0.93 &
  0.93 &
  0.99 &
  0.95 &
  0.98 &
  0.94 &
  0.99 &
  0.883 \\ \hline
GLM-4 &
   &
   &
   &
   &
   &
   &
   &
   &
   &
   &
   &
   &
   &
   &
   &
   &
   &
   &
   \\
$\bullet$ EP &
  \multicolumn{1}{c}{-} &
  \multicolumn{1}{c}{0.32} &
  \multicolumn{1}{c}{-} &
  \multicolumn{1}{c}{0.85} &
  \multicolumn{1}{c}{-} &
  \multicolumn{1}{c}{0.94} &
  \multicolumn{1}{c}{-} &
  \multicolumn{1}{c}{0.95} &
  \multicolumn{1}{c}{-} &
  \multicolumn{1}{c}{0.81} &
  \multicolumn{1}{c}{-} &
  \multicolumn{1}{c|}{0.98} &
  \multicolumn{1}{c}{-} &
  \multicolumn{1}{c}{0.87} &
  \multicolumn{1}{c}{-} &
  \multicolumn{1}{c}{0.93} &
  \multicolumn{1}{c}{-} &
  \multicolumn{1}{c|}{0.99} &
  0.849 \\
$\bullet$ ES &
  0.23 &
  0.47 &
  0.50 &
  0.50 &
  0.69 &
  0.97 &
  0.70 &
  0.99 &
  0.69 &
  0.98 &
  0.68 &
  0.68 &
  0.69 &
  0.97 &
  0.75 &
  0.98 &
  0.83 &
  0.83 &
  0.640 \\
$\bullet$ ET &
  0.45 &
  0.56 &
  0.60 &
  0.97 &
  0.05 &
  0.98 &
  0.05 &
  0.99 &
  0.03 &
  0.99 &
  0.81 &
  1.00 &
  0.68 &
  1.00 &
  0.02 &
  0.98 &
  0.45 &
  1.00 &
  0.349 \\
$\bullet$ EC &
  0.37 &
  0.38 &
  0.89 &
  0.89 &
  0.90 &
  0.96 &
  0.88 &
  0.96 &
  0.81 &
  0.92 &
  0.86 &
  0.96 &
  0.77 &
  0.90 &
  0.87 &
  0.96 &
  0.89 &
  1.00 &
  0.804 \\ \hline
Gemini Pro &
  \multicolumn{2}{c}{} &
  \multicolumn{2}{c}{} &
  \multicolumn{2}{c}{} &
  \multicolumn{2}{c}{} &
  \multicolumn{2}{c}{} &
  \multicolumn{2}{c|}{} &
  \multicolumn{2}{c}{} &
  \multicolumn{2}{c}{} &
  \multicolumn{2}{c|}{} &
   \\
$\bullet$ EP &
  \multicolumn{1}{c}{-} &
  \multicolumn{1}{c}{0.06} &
  \multicolumn{1}{c}{-} &
  \multicolumn{1}{c}{0.12} &
  \multicolumn{1}{c}{-} &
  \multicolumn{1}{c}{0.17} &
  \multicolumn{1}{c}{-} &
  \multicolumn{1}{c}{0.14} &
  \multicolumn{1}{c}{-} &
  \multicolumn{1}{c}{0.18} &
  \multicolumn{1}{c}{-} &
  \multicolumn{1}{c|}{0.42} &
  \multicolumn{1}{c}{-} &
  \multicolumn{1}{c}{0.14} &
  \multicolumn{1}{c}{-} &
  \multicolumn{1}{c}{0.41} &
  \multicolumn{1}{c}{-} &
  \multicolumn{1}{c|}{0.31} &
  0.217 \\
$\bullet$ ES &
  0.10 &
  0.22 &
  0.24 &
  0.37 &
  0.49 &
  0.66 &
  0.32 &
  0.62 &
  0.44 &
  0.59 &
  0.67 &
  0.85 &
  0.15 &
  0.36 &
  0.49 &
  0.69 &
  0.33 &
  0.51 &
  0.359 \\
$\bullet$ ET &
  0.07 &
  0.07 &
  0.01 &
  0.14 &
  0.06 &
  0.31 &
  0.01 &
  0.23 &
  0.02 &
  0.25 &
  0.47&
  0.69 &
  0.06 &
  0.19 &
  0.05 &
  0.50 &
  0.06 &
  0.43 &
  0.090 \\
$\bullet$ EC &
  0.06 &
  0.06 &
  0.11 &
  0.11 &
  0.30 &
  0.30 &
  0.22 &
  0.22 &
  0.20 &
  0.27 &
  0.45 &
  0.60 &
  0.20 &
  0.22 &
  0.36 &
  0.39 &
  0.33 &
  0.34 &
  0.248 \\ \hline \hline
LLaMA-2-7B &
  \multicolumn{1}{l}{} &
  \multicolumn{1}{l}{} &
  \multicolumn{1}{l}{} &
  \multicolumn{1}{l}{} &
  \multicolumn{1}{l}{} &
  \multicolumn{1}{l}{} &
  \multicolumn{1}{l}{} &
  \multicolumn{1}{l}{} &
  \multicolumn{1}{l}{} &
  \multicolumn{1}{l}{} &
  \multicolumn{1}{l}{} &
  \multicolumn{1}{l|}{} &
  \multicolumn{1}{l}{} &
  \multicolumn{1}{l}{} &
  \multicolumn{1}{l}{} &
  \multicolumn{1}{l}{} &
  \multicolumn{1}{l}{} &
  \multicolumn{1}{l|}{} &
  \multicolumn{1}{l}{} \\
$\bullet$ EP &
  \multicolumn{1}{c}{-} &
  \multicolumn{1}{c}{0.58} &
  \multicolumn{1}{c}{-} &
  \multicolumn{1}{c}{0.37} &
  \multicolumn{1}{c}{-} &
  \multicolumn{1}{c}{0.49} &
  \multicolumn{1}{c}{-} &
  \multicolumn{1}{c}{0.61} &
  \multicolumn{1}{c}{-} &
  \multicolumn{1}{c}{0.70} &
  \multicolumn{1}{c}{-} &
  \multicolumn{1}{c|}{0.56} &
  \multicolumn{1}{c}{-} &
  \multicolumn{1}{c}{0.45} &
  \multicolumn{1}{c}{-} &
  \multicolumn{1}{c}{0.68} &
  \multicolumn{1}{c}{-} &
  \multicolumn{1}{c|}{0.40} &
  0.538 \\
$\bullet$ ES &
  0.30 &
  0.94 &
  0.15 &
  0.91 &
  0.09 &
  0.93 &
  0.29 &
  0.92 &
  0.09 &
  0.97 &
  0.38 &
  0.97 &
  0.17 &
  0.90 &
  0.13 &
  0.94 &
  0.06 &
  0.75 &
  0.184 \\
$\bullet$ ET &
  0.38 &
  0.40 &
  0.00 &
  0.19 &
  0.00 &
  0.40 &
  0.00 &
  0.47 &
  0.02 &
  0.60 &
  0.00 &
  0.42 &
  0.03 &
  0.21 &
  0.00 &
  0.53 &
  0.00 &
  0.34 &
  0.048 \\
$\bullet$ EC &
  0.15 &
  0.91 &
  0.02 &
  0.86 &
  0.08 &
  0.90 &
  0.11 &
  0.91 &
  0.08 &
  0.97 &
  0.09 &
  0.94 &
  0.05 &
  0.76 &
  0.01 &
  0.91 &
  0.01 &
  0.68 &
  0.067 \\ \hline
LLaMA-2-13B &
  \multicolumn{1}{l}{} &
  \multicolumn{1}{l}{} &
  \multicolumn{1}{l}{} &
  \multicolumn{1}{l}{} &
  \multicolumn{1}{l}{} &
  \multicolumn{1}{l}{} &
  \multicolumn{1}{l}{} &
  \multicolumn{1}{l}{} &
  \multicolumn{1}{l}{} &
  \multicolumn{1}{l}{} &
  \multicolumn{1}{l}{} &
  \multicolumn{1}{l|}{} &
  \multicolumn{1}{l}{} &
  \multicolumn{1}{l}{} &
  \multicolumn{1}{l}{} &
  \multicolumn{1}{l}{} &
  \multicolumn{1}{l}{} &
  \multicolumn{1}{l|}{} &
  \multicolumn{1}{l}{} \\
$\bullet$ EP &
  \multicolumn{1}{c}{-} &
  \multicolumn{1}{c}{0.10} &
  \multicolumn{1}{c}{-} &
  \multicolumn{1}{c}{0.08} &
  \multicolumn{1}{c}{-} &
  \multicolumn{1}{c}{0.14} &
  \multicolumn{1}{c}{-} &
  \multicolumn{1}{c}{0.17} &
  \multicolumn{1}{c}{-} &
  \multicolumn{1}{c}{0.25} &
  \multicolumn{1}{c}{-} &
  \multicolumn{1}{c|}{0.18} &
  \multicolumn{1}{c}{-} &
  \multicolumn{1}{c}{0.02} &
  \multicolumn{1}{c}{-} &
  \multicolumn{1}{c}{0.32} &
  \multicolumn{1}{c}{-} &
  \multicolumn{1}{c|}{0.23} &
  0.166 \\
$\bullet$ ES &
  0.01 &
  0.02 &
  0.03 &
  0.06 &
  0.00 &
  0.02 &
  0.00 &
  0.02 &
  0.01 &
  0.06 &
  0.00 &
  0.00 &
  0.01 &
  0.02 &
  0.00 &
  0.04 &
  0.00 &
  0.00 &
  0.007 \\
$\bullet$ ET &
  0.01 &
  0.82 &
  0.01 &
  0.86 &
  0.12 &
  0.91 &
  0.49 &
  0.86 &
  0.01 &
  0.85 &
  0.38 &
  0.90 &
  0.12 &
  0.78 &
  0.00 &
  0.88 &
  0.00 &
  0.73 &
  0.127 \\
$\bullet$ EC &
  0.00 &
  0.01 &
  0.00 &
  0.00 &
  0.00 &
  0.00 &
  0.00 &
  0.01 &
  0.00 &
  0.01 &
  0.00 &
  0.02 &
  0.00 &
  0.00 &
  0.00 &
  0.02 &
  0.00 &
  0.00 &
  0.000 \\ \hline
  MetaMath-7B &
  \multicolumn{1}{l}{} &
  \multicolumn{1}{l}{} &
  \multicolumn{1}{l}{} &
  \multicolumn{1}{l}{} &
  \multicolumn{1}{l}{} &
  \multicolumn{1}{l}{} &
  \multicolumn{1}{l}{} &
  \multicolumn{1}{l}{} &
  \multicolumn{1}{l}{} &
  \multicolumn{1}{l}{} &
  \multicolumn{1}{l}{} &
  \multicolumn{1}{l|}{} &
  \multicolumn{1}{l}{} &
  \multicolumn{1}{l}{} &
  \multicolumn{1}{l}{} &
  \multicolumn{1}{l}{} &
  \multicolumn{1}{l}{} &
  \multicolumn{1}{l|}{} &
  \multicolumn{1}{l}{} \\
$\bullet$ EP &
  \multicolumn{1}{c}{-} &
  \multicolumn{1}{c}{0.00} &
  \multicolumn{1}{c}{-} &
  \multicolumn{1}{c}{0.00} &
  \multicolumn{1}{c}{-} &
  \multicolumn{1}{c}{0.00} &
  \multicolumn{1}{c}{-} &
  \multicolumn{1}{c}{0.00} &
  \multicolumn{1}{c}{-} &
  \multicolumn{1}{c}{0.01} &
  \multicolumn{1}{c}{-} &
  \multicolumn{1}{c|}{0.00} &
  \multicolumn{1}{c}{-} &
  \multicolumn{1}{c}{0.00} &
  \multicolumn{1}{c}{-} &
  \multicolumn{1}{c}{0.01} &
  \multicolumn{1}{c}{-} &
  \multicolumn{1}{c|}{0.00} &
  0.002 \\
$\bullet$ ES &
  0.00 &
  0.00 &
  0.00 &
  0.00 &
  0.00 &
  0.00 &
  0.00 &
  0.00 &
  0.00 &
  0.00 &
  0.00 &
  0.00 &
  0.00 &
  0.01 &
  0.00 &
  0.00 &
  0.00 &
  0.00 &
  0.000 \\
$\bullet$ ET &
  0.00 &
  0.00 &
  0.00 &
  0.00 &
  0.00 &
  0.00 &
  0.00 &
  0.00 &
  0.00 &
  0.00 &
  0.00 &
  0.00 &
  0.00 &
  0.00 &
  0.00 &
  0.00 &
  0.00 &
  0.00 &
  0.000 \\
$\bullet$ EC &
  0.00 &
  0.00 &
  0.00 &
  0.00 &
  0.00 &
  0.00 &
  0.00 &
  0.01 &
  0.00 &
  0.00 &
  0.00 &
  0.00 &
  0.00 &
  0.00 &
  0.00 &
  0.00 &
  0.00 &
  0.00 &
  0.000 \\ \hline
MetaMath-13B &
  \multicolumn{2}{l}{} &
  \multicolumn{2}{l}{} &
  \multicolumn{2}{l}{} &
  \multicolumn{2}{l}{} &
  \multicolumn{2}{l}{} &
  \multicolumn{2}{l|}{} &
  \multicolumn{2}{l}{} &
  \multicolumn{2}{l}{} &
  \multicolumn{2}{l|}{} &
   \\
$\bullet$ EP &
\multicolumn{1}{c}{-} &
  \multicolumn{1}{c}{0.00} &
  \multicolumn{1}{c}{-} &
  \multicolumn{1}{c}{0.00} &
  \multicolumn{1}{c}{-} &
  \multicolumn{1}{c}{0.00} &
  \multicolumn{1}{c}{-} &
  \multicolumn{1}{c}{0.00} &
  \multicolumn{1}{c}{-} &
  \multicolumn{1}{c}{0.00} &
  \multicolumn{1}{c}{-} &
  \multicolumn{1}{c|}{0.00} &
  \multicolumn{1}{c}{-} &
  \multicolumn{1}{c}{0.00} &
  \multicolumn{1}{c}{-} &
  \multicolumn{1}{c}{0.00} &
  \multicolumn{1}{c}{-} &
  \multicolumn{1}{c|}{0.00} &
  0.000 \\
$\bullet$ ES &
  0.00 &
  0.01 &
  0.00 &
  0.00 &
  0.00 &
  0.00 &
  0.00 &
  0.00 &
  0.00 &
  0.00 &
  0.00 &
  0.01 &
  0.00 &
  0.00 &
  0.00 &
  0.00 &
  0.00 &
  0.00 &
  0.000 \\
$\bullet$ ET &
  0.00 &
  0.00 &
  0.00 &
  0.00 &
  0.00 &
  0.00 &
  0.00 &
  0.00 &
  0.00 &
  0.00 &
  0.00 &
  0.00 &
  0.00 &
  0.00 &
  0.00 &
  0.00 &
  0.00 &
  0.00 &
  0.000 \\
$\bullet$ EC &
  0.00 &
  0.00 &
  0.00 &
  0.00 &
  0.00 &
  0.00 &
  0.00 &
  0.00 &
  0.00 &
  0.00 &
  0.00 &
  0.00 &
  0.00 &
  0.00 &
  0.00 &
  0.00 &
  0.00 &
  0.01 &
  0.000 \\ \hline
\end{tabular}
\end{adjustbox}
}
\caption{Accuracy of different closed-source and open-source models on different error types of GSM8K. We use zero-shot prompts uniformly here. Different error types are replaced with the first two letters of their names, for example, calculation error is represented by CA. For \textit{EP}, the results of $acc_{1}$ are presented; for \textit{ES}, the results of both $acc_{2}$ and $acc_{1}$ are showcased; for \textit{ET}, the results of both $acc_{3}$ and $acc_{1}$ are displayed; for \textit{EC}, the results of both $acc_{4}$ and $acc_{1}$ are exhibited. And we calculate the average $acc_{i}(i = 1, 2, 3, 4)$ as \textit{Avg} for \textit{EP}, \textit{ES}, \textit{ET} and \textit{EC}.}
\label{tab:GSM8K main experiment}
\end{table*}

\begin{table*}[ht]
\renewcommand\arraystretch{0.7}
\centering
\tabcolsep=0.15cm
\scalebox{0.8}{
\begin{adjustbox}{center}
\begin{tabular}{ccccccccccccc|cccccc|c}
\hline
\textbf{} &
  \multicolumn{2}{c}{CA} &
  \multicolumn{2}{c}{CO} &
  \multicolumn{2}{c}{CV} &
  \multicolumn{2}{c}{CS} &
  \multicolumn{2}{c}{MS} &
  \multicolumn{2}{c|}{HA} &
  \multicolumn{2}{c}{UC} &
  \multicolumn{2}{c}{OP} &
  \multicolumn{2}{c|}{FC} &
  Avg \\ 
  &
  \multicolumn{1}{c}{$acc_{i}$} &
  \multicolumn{1}{c}{$acc_{1}$} &
  \multicolumn{1}{c}{$acc_{i}$} &
  \multicolumn{1}{c}{$acc_{1}$} &
  \multicolumn{1}{c}{$acc_{i}$} &
  \multicolumn{1}{c}{$acc_{1}$} &
  \multicolumn{1}{c}{$acc_{i}$} &
  \multicolumn{1}{c}{$acc_{1}$} &
  \multicolumn{1}{c}{$acc_{i}$} &
  \multicolumn{1}{c}{$acc_{1}$} &
  \multicolumn{1}{c}{$acc_{i}$} &
  \multicolumn{1}{c|}{$acc_{1}$} &
  \multicolumn{1}{c}{$acc_{i}$} &
  \multicolumn{1}{c}{$acc_{1}$} &
  \multicolumn{1}{c}{$acc_{i}$} &
  \multicolumn{1}{c}{$acc_{1}$} &
  \multicolumn{1}{c}{$acc_{i}$} &
  \multicolumn{1}{c|}{$acc_{1}$} &
  \multicolumn{1}{c}{$acc_{i}$} \\
  \hline
GPT-3.5 &
  \multicolumn{1}{l}{} &
  \multicolumn{1}{l}{} &
  \multicolumn{1}{l}{} &
  \multicolumn{1}{l}{} &
  \multicolumn{1}{l}{} &
  \multicolumn{1}{l}{} &
  \multicolumn{1}{l}{} &
  \multicolumn{1}{l}{} &
  \multicolumn{1}{l}{} &
  \multicolumn{1}{l}{} &
  \multicolumn{1}{l}{} &
  \multicolumn{1}{l|}{} &
  \multicolumn{1}{l}{} &
  \multicolumn{1}{l}{} &
  \multicolumn{1}{l}{} &
  \multicolumn{1}{l}{} &
  \multicolumn{1}{l}{} &
  \multicolumn{1}{l|}{} &
  \multicolumn{1}{l}{} \\
$\bullet$ EP &
  \multicolumn{1}{c}{-} &
  \multicolumn{1}{c}{0.33} &
  \multicolumn{1}{c}{-} &
  \multicolumn{1}{c}{0.53} &
  \multicolumn{1}{c}{-} &
  \multicolumn{1}{c}{0.61} &
  \multicolumn{1}{c}{-} &
  \multicolumn{1}{c}{0.56} &
  \multicolumn{1}{c}{-} &
  \multicolumn{1}{c}{0.44} &
  \multicolumn{1}{c}{-} &
  \multicolumn{1}{c|}{0.62} &
  \multicolumn{1}{c}{-} &
  \multicolumn{1}{c}{0.17} &
  \multicolumn{1}{c}{-} &
  \multicolumn{1}{c}{0.59} &
  \multicolumn{1}{c}{-} &
  \multicolumn{1}{c|}{0.59} &
  0.493 \\
$\bullet$ ES &
  0.07 &
  0.34 &
  0.09 &
  0.56 &
  0.26 &
  0.67 &
  0.19 &
  0.56 &
  0.20 &
  0.50 &
  0.20 &
  0.56 &
  0.07 &
  0.31 &
  0.21 &
  0.61 &
  0.12 &
  0.58 &
  0.157 \\
$\bullet$ ET &
  0.20 &
  0.28 &
  0.15 &
  0.46 &
  0.04 &
  0.59 &
  0.05 &
  0.45 &
  0.03 &
  0.50 &
  0.36 &
  0.68 &
  0.08 &
  0.20 &
  0.11 &
  0.55 &
  0.14 &
  0.45 &
  0.129 \\
$\bullet$ EC &
  0.07 &
  0.16 &
  0.19 &
  0.34 &
  0.18 &
  0.32 &
  0.16 &
  0.27 &
  0.12 &
  0.29 &
  0.13 &
  0.37 &
  0.02 &
  0.10 &
  0.18 &
  0.46 &
  0.22 &
  0.41 &
  0.141 \\ \hline
GPT-4 &
   &
   &
   &
   &
   &
   &
   &
   &
   &
   &
   &
   &
   &
   &
   &
   &
   &
   &
   \\
$\bullet$ EP &
  \multicolumn{1}{c}{-} &
  \multicolumn{1}{c}{0.66} &
  \multicolumn{1}{c}{-} &
  \multicolumn{1}{c}{0.87} &
  \multicolumn{1}{c}{-} &
  \multicolumn{1}{c}{0.98} &
  \multicolumn{1}{c}{-} &
  \multicolumn{1}{c}{0.88} &
  \multicolumn{1}{c}{-} &
  \multicolumn{1}{c}{0.91} &
  \multicolumn{1}{c}{-} &
  \multicolumn{1}{c|}{0.98} &
  \multicolumn{1}{c}{-} &
  \multicolumn{1}{c}{0.98} &
  \multicolumn{1}{c}{-} &
  \multicolumn{1}{c}{0.99} &
  \multicolumn{1}{c}{-} &
  \multicolumn{1}{c|}{1.00} &
  0.917 \\
$\bullet$ ES &
  0.52 &
  0.84 &
  0.71 &
  0.92 &
  0.79 &
  0.99 &
  0.68 &
  0.96 &
  0.58 &
  0.96 &
  0.64 &
  0.99 &
  0.82 &
  0.99 &
  0.79 &
  1.00 &
  0.69 &
  1.00 &
  0.691 \\
$\bullet$ ET &
  0.69 &
  0.89 &
  0.31 &
  0.97 &
  0.55 &
  1.00 &
  0.11 &
  0.96 &
  0.12 &
  0.96 &
  0.95 &
  1.00 &
  0.73 &
  1.00 &
  0.50 &
  1.00 &
  0.72 &
  1.00 &
  0.520 \\
$\bullet$ EC &
  0.57 &
  0.65 &
  0.78 &
  0.87 &
  0.85 &
  0.97 &
  0.76 &
  0.85 &
  0.76 &
  0.89 &
  0.89 &
  0.98 &
  0.88 &
  0.98 &
  0.89 &
  0.99 &
  0.91 &
  1.00 &
  0.810 \\ \hline
GLM-4 &
   &
   &
   &
   &
   &
   &
   &
   &
   &
   &
   &
   &
   &
   &
   &
   &
   &
   &
   \\
$\bullet$ EP & 
\multicolumn{1}{c}{-} &
  \multicolumn{1}{c}{0.34} & \multicolumn{1}{c}{-} &
  \multicolumn{1}{c}{0.74} & \multicolumn{1}{c}{-} &
  \multicolumn{1}{c}{0.84} & \multicolumn{1}{c}{-} &
  \multicolumn{1}{c}{0.66} & \multicolumn{1}{c}{-} &
  \multicolumn{1}{c}{0.76} & \multicolumn{1}{c}{-} &
  \multicolumn{1}{c|}{0.99} & \multicolumn{1}{c}{-} &
  \multicolumn{1}{c}{0.74} & \multicolumn{1}{c}{-} &
  \multicolumn{1}{c}{0.89} & \multicolumn{1}{c}{-} &
  \multicolumn{1}{c|}{0.99} & 
  0.772 \\
$\bullet$ ES &
  0.23 &
  0.57 &
  0.50 &
  0.95 &
  0.62 &
  0.96 &
  \multicolumn{1}{r}{0.53} &
  \multicolumn{1}{r}{0.80} &
  0.52 &
  0.92 &
  0.45 &
  1.00 &
  0.56 &
  0.96 &
  0.59 &
  0.97 &
  0.46 &
  1.00 &
  0.496 \\
$\bullet$ ET &
  0.46 &
  0.86 &
  0.50 &
  \multicolumn{1}{r}{0.95} &
  0.09 &
  \multicolumn{1}{r}{0.92} &
  0.24 &
  0.87 &
  0.02 &
  \multicolumn{1}{r}{0.92} &
  0.18 &
  \multicolumn{1}{r|}{1.00} &
  0.28 &
  \multicolumn{1}{r}{0.96} &
  0.01 &
  \multicolumn{1}{r}{0.98} &
  0.18 &
  \multicolumn{1}{r|}{0.99} &
  0.218 \\
$\bullet$ EC &
  0.25 &
  0.41 &
  0.61 &
  0.85 &
  0.70 &
  0.86 &
  0.55 &
  0.73 &
  0.53 &
  0.80 &
  0.69 &
  0.98 &
  0.54 &
  0.73 &
  0.62 &
  0.92 &
  0.68 &
  0.99 &
  0.574 \\ \hline
Gemini Pro &
  \multicolumn{2}{c}{} &
  \multicolumn{2}{c}{} &
  \multicolumn{2}{c}{} &
  \multicolumn{2}{c}{} &
  \multicolumn{2}{c}{} &
  \multicolumn{2}{c|}{} &
  \multicolumn{2}{c}{} &
  \multicolumn{2}{c}{} &
  \multicolumn{2}{c|}{} &
   \\
$\bullet$ EP & \multicolumn{1}{c}{-} &
  \multicolumn{1}{c}{0.06} & \multicolumn{1}{c}{-} &
  \multicolumn{1}{c}{0.27} & \multicolumn{1}{c}{-} &
  \multicolumn{1}{c}{0.25} & \multicolumn{1}{c}{-} &
  \multicolumn{1}{c}{0.06} & \multicolumn{1}{c}{-} &
  \multicolumn{1}{c}{0.24} & \multicolumn{1}{c}{-} &
  \multicolumn{1}{c|}{0.23} & \multicolumn{1}{c}{-} &
  \multicolumn{1}{c}{0.03} & \multicolumn{1}{c}{-} &
  \multicolumn{1}{c}{0.31} & \multicolumn{1}{c}{-} &
  \multicolumn{1}{c|}{0.32} &
  0.197 \\
$\bullet$ ES &
  0.00 &
  0.01 &
  0.05 &
  0.11 &
  0.02 &
  0.07 &
  0.01 &
  0.01 &
  0.03 &
  0.05 &
  0.03 &
  0.05 &
  0.00 &
  0.01 &
  0.08 &
  0.18 &
  0.06 &
  0.09 &
  0.031 \\
$\bullet$ ET &
  0.02 &
  \multicolumn{1}{r}{0.03} &
  0.03 &
  \multicolumn{1}{r}{0.21} &
  0.05 &
  \multicolumn{1}{r}{0.22} &
  0.00 &
  \multicolumn{1}{r}{0.07} &
  0.03 &
  \multicolumn{1}{r}{0.20} &
  0.20 &
  \multicolumn{1}{r|}{0.34} &
  0.01 &
  \multicolumn{1}{r}{0.03} &
  0.02 &
  \multicolumn{1}{r}{0.27} &
  0.11 &
  \multicolumn{1}{r|}{0.25} &
  0.052 \\
$\bullet$ EC &
  \multicolumn{1}{r}{0.04} &
  \multicolumn{1}{r}{0.07} &
  \multicolumn{1}{r}{0.28} &
  \multicolumn{1}{r}{0.38} &
  \multicolumn{1}{r}{0.25} &
  \multicolumn{1}{r}{0.28} &
  \multicolumn{1}{r}{0.13} &
  \multicolumn{1}{r}{0.14} &
  \multicolumn{1}{r}{0.20} &
  \multicolumn{1}{r}{0.30} &
  \multicolumn{1}{r}{0.33} &
  \multicolumn{1}{r|}{0.40} &
  \multicolumn{1}{r}{0.07} &
  \multicolumn{1}{r}{0.07} &
  \multicolumn{1}{r}{0.25} &
  \multicolumn{1}{r}{0.36} &
  \multicolumn{1}{r}{0.25} &
  \multicolumn{1}{r|}{0.34} &
  0.200 \\ \hline \hline
LLaMA-2-7B &
  \multicolumn{1}{l}{} &
  \multicolumn{1}{l}{} &
  \multicolumn{1}{l}{} &
  \multicolumn{1}{l}{} &
  \multicolumn{1}{l}{} &
  \multicolumn{1}{l}{} &
  \multicolumn{1}{l}{} &
  \multicolumn{1}{l}{} &
  \multicolumn{1}{l}{} &
  \multicolumn{1}{l}{} &
  \multicolumn{1}{l}{} &
  \multicolumn{1}{l|}{} &
  \multicolumn{1}{l}{} &
  \multicolumn{1}{l}{} &
  \multicolumn{1}{l}{} &
  \multicolumn{1}{l}{} &
  \multicolumn{1}{l}{} &
  \multicolumn{1}{l|}{} &
   \\
$\bullet$ EP & \multicolumn{1}{c}{-} &
  \multicolumn{1}{c}{0.56} & \multicolumn{1}{c}{-} &
  \multicolumn{1}{c}{0.37} & \multicolumn{1}{c}{-} &
  \multicolumn{1}{c}{0.63} & \multicolumn{1}{c}{-} &
  \multicolumn{1}{c}{0.58} & \multicolumn{1}{c}{-} &
  \multicolumn{1}{c}{0.75} & \multicolumn{1}{c}{-} &
  \multicolumn{1}{c|}{0.49} & \multicolumn{1}{c}{-} &
  \multicolumn{1}{c}{0.31} & \multicolumn{1}{c}{-} &
  \multicolumn{1}{c}{0.66} & \multicolumn{1}{c}{-} &
  \multicolumn{1}{c|}{0.47} &
  0.536 \\
$\bullet$ ES &
  0.15 &
  0.88 &
  0.09 &
  0.70 &
  0.22 &
  0.90 &
  0.20 &
  0.88 &
  0.13 &
  0.99 &
  0.28 &
  0.82 &
  0.13 &
  0.83 &
  0.19 &
  0.88 &
  0.19 &
  0.87 &
  0.176 \\
$\bullet$ ET &
  0.32 &
  0.35 &
  0.01 &
  0.23 &
  0.00 &
  0.46 &
  0.06 &
  0.35 &
  0.06 &
  0.64 &
  0.00 &
  0.20 &
  0.02 &
  0.14 &
  0.00 &
  0.45 &
  0.00 &
  0.40 &
  0.052 \\
$\bullet$ EC &
  0.04 &
  0.78 &
  0.04 &
  0.63 &
  0.05 &
  0.85 &
  0.06 &
  0.86 &
  0.06 &
  0.95 &
  0.07 &
  0.73 &
  0.02 &
  0.72 &
  0.01 &
  0.83 &
  0.00 &
  0.78 &
  0.039 \\ \hline
LLaMA-2-13B &
  \multicolumn{1}{l}{} &
  \multicolumn{1}{l}{} &
  \multicolumn{1}{l}{} &
  \multicolumn{1}{l}{} &
  \multicolumn{1}{l}{} &
  \multicolumn{1}{l}{} &
  \multicolumn{1}{l}{} &
  \multicolumn{1}{l}{} &
  \multicolumn{1}{l}{} &
  \multicolumn{1}{l}{} &
  \multicolumn{1}{l}{} &
  \multicolumn{1}{l|}{} &
  \multicolumn{1}{l}{} &
  \multicolumn{1}{l}{} &
  \multicolumn{1}{l}{} &
  \multicolumn{1}{l}{} &
  \multicolumn{1}{l}{} &
  \multicolumn{1}{l|}{} &
   \\
$\bullet$ EP & \multicolumn{1}{c}{-} &
  \multicolumn{1}{c}{0.14} & \multicolumn{1}{c}{-} &
  \multicolumn{1}{c}{0.15} & \multicolumn{1}{c}{-} &
  \multicolumn{1}{c}{0.30} & \multicolumn{1}{c}{-} &
  \multicolumn{1}{c}{0.27} & \multicolumn{1}{c}{-} &
  \multicolumn{1}{c}{0.29} & \multicolumn{1}{c}{-} &
  \multicolumn{1}{c|}{0.23} & \multicolumn{1}{c}{-} &
  \multicolumn{1}{c}{0.08} & \multicolumn{1}{c}{-} &
  \multicolumn{1}{c}{0.38} & \multicolumn{1}{c}{-} &
  \multicolumn{1}{c|}{0.13} &
  0.219 \\
$\bullet$ ES &
  0.01 &
  0.10 &
  0.02 &
  0.06 &
  0.01 &
  0.12 &
  0.02 &
  0.10 &
  0.01 &
  0.09 &
  0.00 &
  0.06 &
  0.00 &
  0.05 &
  0.00 &
  0.03 &
  0.01 &
  0.03 &
  0.009 \\
$\bullet$ ET &
  0.02 &
  0.92 &
  0.05 &
  0.86 &
  0.13 &
  0.95 &
  0.38 &
  0.95 &
  0.00 &
  0.99 &
  0.26 &
  0.92 &
  0.20 &
  0.92 &
  0.00 &
  0.94 &
  0.00 &
  1.00 &
  0.116 \\
$\bullet$ EC &
  0.00 &
  0.01 &
  0.00 &
  0.01 &
  0.00 &
  0.01 &
  0.00 &
  0.01 &
  0.00 &
  0.00 &
  0.00 &
  0.04 &
  0.00 &
  0.00 &
  0.00 &
  0.01 &
  0.00 &
  0.00 &
  0.000 \\ \hline
MetaMath-7B &
  \multicolumn{1}{l}{} &
  \multicolumn{1}{l}{} &
  \multicolumn{1}{l}{} &
  \multicolumn{1}{l}{} &
  \multicolumn{1}{l}{} &
  \multicolumn{1}{l}{} &
  \multicolumn{1}{l}{} &
  \multicolumn{1}{l}{} &
  \multicolumn{1}{l}{} &
  \multicolumn{1}{l}{} &
  \multicolumn{1}{l}{} &
  \multicolumn{1}{l|}{} &
  \multicolumn{1}{l}{} &
  \multicolumn{1}{l}{} &
  \multicolumn{1}{l}{} &
  \multicolumn{1}{l}{} &
  \multicolumn{1}{l}{} &
  \multicolumn{1}{l|}{} &
  \multicolumn{1}{l}{} \\
$\bullet$ EP & \multicolumn{1}{c}{-} &
  \multicolumn{1}{c}{0.01} & \multicolumn{1}{c}{-} &
  \multicolumn{1}{c}{0.01} & \multicolumn{1}{c}{-} &
  \multicolumn{1}{c}{0.00} & \multicolumn{1}{c}{-} &
  \multicolumn{1}{c}{0.00} & \multicolumn{1}{c}{-} &
  \multicolumn{1}{c}{0.00} & \multicolumn{1}{c}{-} &
  \multicolumn{1}{c|}{0.00} & \multicolumn{1}{c}{-} &
  \multicolumn{1}{c}{0.00} & \multicolumn{1}{c}{-} &
  \multicolumn{1}{c}{0.00} & \multicolumn{1}{c}{-} &
  \multicolumn{1}{c|}{0.00} &
  0.002 \\
$\bullet$ ES &
  0.00 &
  0.00 &
  0.00 &
  0.00 &
  0.00 &
  0.00 &
  0.00 &
  0.01 &
  0.00 &
  0.01 &
  0.00 &
  0.00 &
  0.00 &
  0.00 &
  0.00 &
  0.00 &
  0.00 &
  0.00 &
  0.000 \\
$\bullet$ ET &
  0.00 &
  0.00 &
  0.00 &
  0.00 &
  0.00 &
  0.00 &
  0.00 &
  0.00 &
  0.00 &
  0.00 &
  0.00 &
  0.00 &
  0.00 &
  0.00 &
  0.00 &
  0.00 &
  0.00 &
  0.00 &
  0.000 \\
$\bullet$ EC &
  0.00 &
  0.00 &
  0.00 &
  0.00 &
  0.00 &
  0.00 &
  0.00 &
  0.00 &
  0.00 &
  0.00 &
  0.00 &
  0.00 &
  0.00 &
  0.00 &
  0.00 &
  0.00 &
  0.00 &
  0.00 &
  0.000 \\ \hline
MetaMath-13B &
  \multicolumn{2}{l}{} &
  \multicolumn{2}{l}{} &
  \multicolumn{2}{l}{} &
  \multicolumn{2}{l}{} &
  \multicolumn{2}{l}{} &
  \multicolumn{2}{l|}{} &
  \multicolumn{2}{l}{} &
  \multicolumn{2}{l}{} &
  \multicolumn{2}{l|}{} &
   \\
$\bullet$ EP & \multicolumn{1}{c}{-} &
  \multicolumn{1}{c}{0.00} & \multicolumn{1}{c}{-} &
  \multicolumn{1}{c}{0.00} & \multicolumn{1}{c}{-} &
  \multicolumn{1}{c}{0.00} & \multicolumn{1}{c}{-} &
  \multicolumn{1}{c}{0.00} & \multicolumn{1}{c}{-} &
  \multicolumn{1}{c}{0.00} & \multicolumn{1}{c}{-} &
  \multicolumn{1}{c|}{0.00} & \multicolumn{1}{c}{-} &
  \multicolumn{1}{c}{0.00} & \multicolumn{1}{c}{-} &
  \multicolumn{1}{c}{0.00} & \multicolumn{1}{c}{-} &
  \multicolumn{1}{c|}{0.00} &
  0.000 \\
$\bullet$ ES &
  0.00 &
  0.01 &
  0.00 &
  0.00 &
  0.00 &
  0.00 &
  0.00 &
  0.00 &
  0.00 &
  0.00 &
  0.00 &
  0.00 &
  0.00 &
  0.00 &
  0.00 &
  0.00 &
  0.00 &
  0.00 &
  0.000 \\
$\bullet$ ET &
  0.00 &
  0.00 &
  0.00 &
  0.00 &
  0.00 &
  0.00 &
  0.00 &
  0.00 &
  0.00 &
  0.00 &
  0.00 &
  0.00 &
  0.00 &
  0.00 &
  0.00 &
  0.00 &
  0.00 &
  0.00 &
  0.000 \\
$\bullet$ EC &
  0.00 &
  0.00 &
  0.00 &
  0.00 &
  0.00 &
  0.00 &
  0.00 &
  0.00 &
  0.00 &
  0.00 &
  0.00 &
  0.00 &
  0.00 &
  0.00 &
  0.00 &
  0.00 &
  0.00 &
  0.00 &
  0.000 \\ \hline
\end{tabular}
\end{adjustbox}
}
\caption{Accuracy of different closed-source and open-source models on different error types of MathQA. We use zero-shot prompts uniformly here. Different error types are replaced with the first two letters of their names, for example, calculation error is represented by CA. For \textit{EP}, the results of $acc_{1}$ are presented; for \textit{ES}, the results of both $acc_{2}$ and $acc_{1}$ are showcased; for \textit{ET}, the results of both $acc_{3}$ and $acc_{1}$ are displayed; for \textit{EC}, the results of both $acc_{4}$ and $acc_{1}$ are exhibited. And we calculate the average $acc_{i}(i = 1, 2, 3, 4)$ as \textit{Avg} for \textit{EP}, \textit{ES}, \textit{ET} and \textit{EC}.}
\label{tab:MathQA main experiment}
\end{table*}

\begin{table*}[ht]
\centering
\tabcolsep=0.1cm
\scalebox{0.7}{
\begin{adjustbox}{center}
\begin{tabular}{ccccccccrr|ccrclclrr|rr}
\hline
 &
  \multicolumn{9}{c|}{GSM8K} &
  \multicolumn{9}{c|}{MathQA} &
  \multicolumn{1}{c}{} &
  \multicolumn{1}{c}{} \\
 &
  EP &
  \multicolumn{2}{c}{ES} &
  \multicolumn{2}{c}{ET} &
  \multicolumn{2}{c}{EC} &
  \multicolumn{2}{c|}{Avg} &
  EP &
  \multicolumn{2}{c}{ES} &
  \multicolumn{2}{c}{ET} &
  \multicolumn{2}{c}{EC} &
  \multicolumn{2}{c|}{Avg} &
  \multicolumn{2}{c}{Avg} \\ 
  &
  \multicolumn{1}{c}{$acc_{1}$} &
  \multicolumn{1}{c}{$acc_{2}$} &
  \multicolumn{1}{c}{$acc_{1}$} &
  \multicolumn{1}{c}{$acc_{3}$} &
  \multicolumn{1}{c}{$acc_{1}$} &
  \multicolumn{1}{c}{$acc_{4}$} &
  \multicolumn{1}{c}{$acc_{1}$} &
  \multicolumn{1}{c}{$acc$} &
  \multicolumn{1}{c|}{$acc_{1}$} &
   \multicolumn{1}{c}{$acc_{1}$} &
  \multicolumn{1}{c}{$acc_{2}$} &
  \multicolumn{1}{c}{$acc_{1}$} &
  \multicolumn{1}{c}{$acc_{3}$} &
  \multicolumn{1}{c}{$acc_{1}$} &
  \multicolumn{1}{c}{$acc_{4}$} &
  \multicolumn{1}{c}{$acc_{1}$} &
  \multicolumn{1}{c}{$acc$} &
  \multicolumn{1}{c|}{$acc_{1}$} &
  \multicolumn{1}{c}{$acc$} &
  \multicolumn{1}{c}{$acc_{1}$} \\
  \hline
MetaMath-7B &
  0.002 &
  0.000 &
  0.001 &
  0.000 &
  0.000 &
  0.000 &
  0.001 &
  0.001 &
  0.001 &
  0.002 &
  0.000 &
  0.002 &
  0.000 &
  0.000 &
  0.000 &
  0.000 &
  0.001 &
  0.001 &
  0.001 &
  0.001 \\
MetaMath-13B &
  0.000 &
  0.000 &
  0.002 &
  0.000 &
  0.000 &
  0.000 &
  0.001 &
  0.000 &
  0.001 &
  0.000 &
  0.000 &
  0.001 &
  0.000 &
  0.000 &
  0.000 &
  0.000 &
  0.000 &
  0.000 &
  0.000 &
  0.001 \\
\hline
\end{tabular}
\end{adjustbox}
}
\caption{Average accuracy of MetaMath models in different tasks on GSM8K and MathQA under zero-shot prompts.}
\label{tab:Based on Model-MetaMath, main table}
\end{table*}

\begin{table*}[htbp]
\centering
\tabcolsep=0.1cm
\scalebox{0.7}{
\begin{adjustbox}{center}
\centering
\begin{tabular}{lcccccccccccc|cccccc|cc}
\hline
\multicolumn{1}{c}{} &
  \multicolumn{2}{c}{CA} &
  \multicolumn{2}{c}{CO} &
  \multicolumn{2}{c}{CV} &
  \multicolumn{2}{c}{CS} &
  \multicolumn{2}{c}{MS} &
  \multicolumn{2}{c|}{HA} &
  \multicolumn{2}{c}{UC} &
  \multicolumn{2}{c}{OP} &
  \multicolumn{2}{c|}{FC} &
  \multicolumn{2}{c}{Avg} \\ 
  &
  \multicolumn{1}{c}{$acc$} &
  \multicolumn{1}{c}{$acc_{1}$} &
  \multicolumn{1}{c}{$acc$} &
  \multicolumn{1}{c}{$acc_{1}$} &
  \multicolumn{1}{c}{$acc$} &
  \multicolumn{1}{c}{$acc_{1}$} &
  \multicolumn{1}{c}{$acc$} &
  \multicolumn{1}{c}{$acc_{1}$} &
  \multicolumn{1}{c}{$acc$} &
  \multicolumn{1}{c}{$acc_{1}$} &
  \multicolumn{1}{c}{$acc$} &
  \multicolumn{1}{c|}{$acc_{1}$} &
  \multicolumn{1}{c}{$acc$} &
  \multicolumn{1}{c}{$acc_{1}$} &
  \multicolumn{1}{c}{$acc$} &
  \multicolumn{1}{c}{$acc_{1}$} &
  \multicolumn{1}{c}{$acc$} &
  \multicolumn{1}{c|}{$acc_{1}$} &
  \multicolumn{1}{c}{$acc$} &
  \multicolumn{1}{c}{$acc_{1}$} \\
  \hline
MetaMath-7B &
  0.001 &
  0.001 &
  0.001 &
  0.001 &
  0.000 &
  0.000 &
  0.000 &
  0.003 &
  0.001 &
  0.003 &
  0.000 &
  0.001 &
  0.001 &
  0.001 &
  0.000 &
  0.000 &
  0.000 &
  0.000 &
  0.001 &
  0.001 \\
MetaMath-13B &
  0.000 &
  0.003 &
  0.000 &
  0.000 &
  0.000 &
  0.000 &
  0.000 &
  0.000 &
  0.000 &
  0.000 &
  0.000 &
  0.000 &
  0.000 &
  0.000 &
  0.000 &
  0.001 &
  0.000 &
  0.001 &
  0.000 &
  0.001 \\ \hline
\multicolumn{1}{c}{Avg} &
  0.001 &
  0.002 &
  0.001 &
  0.001 &
  0.000 &
  0.000 &
  0.000 &
  0.001 &
  0.001 &
  0.001 &
  0.000 &
  0.001 &
  0.001 &
  0.001 &
  0.000 &
  0.001 &
  0.000 &
  0.001 &
  0.000 &
  0.001 \\ \hline
\end{tabular}
\end{adjustbox}
}
\caption{Average accuracy of MetaMath models in different error types on GSM8K and MathQA under zero-shot prompts.}
\label{tab:Based on Type, MetaMath main table}
\end{table*}

\begin{table*}[ht]
\centering
\tabcolsep=0.15cm
\scalebox{0.7}{
\begin{adjustbox}{center}
\begin{tabular}{ccccccccccccc|cccccc|cc}
\hline
 &
  \multicolumn{2}{c}{CA} &
  \multicolumn{2}{c}{CO} &
  \multicolumn{2}{c}{CV} &
  \multicolumn{2}{c}{CS} &
  \multicolumn{2}{c}{MS} &
  \multicolumn{2}{c|}{HA} &
  \multicolumn{2}{c}{UC} &
  \multicolumn{2}{c}{OP} &
  \multicolumn{2}{c|}{FC} &
  \multicolumn{2}{c}{Avg} \\ 
  &
  \multicolumn{1}{c}{$acc_{i}$} &
  \multicolumn{1}{c}{$acc_{1}$} &
  \multicolumn{1}{c}{$acc_{i}$} &
  \multicolumn{1}{c}{$acc_{1}$} &
  \multicolumn{1}{c}{$acc_{i}$} &
  \multicolumn{1}{c}{$acc_{1}$} &
  \multicolumn{1}{c}{$acc_{i}$} &
  \multicolumn{1}{c}{$acc_{1}$} &
  \multicolumn{1}{c}{$acc_{i}$} &
  \multicolumn{1}{c}{$acc_{1}$} &
  \multicolumn{1}{c}{$acc_{i}$} &
  \multicolumn{1}{c|}{$acc_{1}$} &
  \multicolumn{1}{c}{$acc_{i}$} &
  \multicolumn{1}{c}{$acc_{1}$} &
  \multicolumn{1}{c}{$acc_{i}$} &
  \multicolumn{1}{c}{$acc_{1}$} &
  \multicolumn{1}{c}{$acc_{i}$} &
  \multicolumn{1}{c|}{$acc_{1}$} &
  \multicolumn{1}{c}{$acc_{i}$} &
  \multicolumn{1}{c}{$acc_{1}$} \\
  \hline
EP &
  \multicolumn{1}{c}{-} &
  \multicolumn{1}{c}{0.003} &
  \multicolumn{1}{c}{-} &
  \multicolumn{1}{c}{0.003} &
  \multicolumn{1}{c}{-} &
  \multicolumn{1}{c}{0.000} &
  \multicolumn{1}{c}{-} &
  \multicolumn{1}{c}{0.000} &
  \multicolumn{1}{c}{-} &
  \multicolumn{1}{c}{0.003} &
  \multicolumn{1}{c}{-} &
  \multicolumn{1}{c|}{0.000} &
  \multicolumn{1}{c}{-} &
  \multicolumn{1}{c}{0.003} &
  \multicolumn{1}{c}{-} &
  \multicolumn{1}{c}{0.000} &
  \multicolumn{1}{c}{-} &
  \multicolumn{1}{c|}{0.000} &
  \multicolumn{1}{c}{-} &
  \multicolumn{1}{c}{0.001} \\
ES & 0.000 & 0.005 & 0.000 & 0.000 & 0.000 & 0.000 & 0.000 & 0.003 & 0.000 & 0.003 & 0.000 & 0.003 & 0.000 & 0.000 & 0.000 & 0.000 & 0.000 & 0.003 & 0.000 & 0.002 \\
ET & 0.000 & 0.000 & 0.000 & 0.000 & 0.000 & 0.000 & 0.000 & 0.000 & 0.000 & 0.000 & 0.000 & 0.000 & 0.000 & 0.000 & 0.000 & 0.000 & 0.000 & 0.000 & 0.000 & 0.000 \\
\hline
\end{tabular}
\end{adjustbox}
}
\caption{Average accuracy of MetaMath models on four tasks in different error types on GSM8K and MathQA under zero-shot prompts.}
\label{tab:Based on Task, MetaMath main table}
\end{table*}

\begin{table*}[ht]
\centering
\adjustbox{center}{
  \begin{minipage}[t]{0.48\linewidth}
    \centering
    \scalebox{0.8}{
    \begin{tabular}{l|clclclclclclclclcl}
\hline
 &
  \multicolumn{2}{c}{CA} &
  \multicolumn{2}{c}{CO} &
  \multicolumn{2}{c}{CV} &
  \multicolumn{2}{c}{CS} &
  \multicolumn{2}{c}{MS} &
  \multicolumn{2}{c}{HA} &
  \multicolumn{2}{c}{UC} &
  \multicolumn{2}{c}{OP} &
  \multicolumn{2}{c}{FC} \\ \hline
CA &
  \multicolumn{2}{c}{\cellcolor[HTML]{340096}{\color[HTML]{FFFFFF} 147}} &
  \multicolumn{2}{c}{\cellcolor[HTML]{3303CF}{\color[HTML]{FFFFFF} 19}} &
  \multicolumn{2}{c}{\cellcolor[HTML]{6665CD}{\color[HTML]{000000} 12}} &
  \multicolumn{2}{c}{\cellcolor[HTML]{6665CD}{\color[HTML]{000000} 14}} &
  \multicolumn{2}{c}{\cellcolor[HTML]{CBCEFB}{\color[HTML]{000000} 1}} &
  \multicolumn{2}{c}{\cellcolor[HTML]{9698ED}{\color[HTML]{000000} 5}} &
  \multicolumn{2}{c}{\cellcolor[HTML]{6665CD}{\color[HTML]{000000} 13}} &
  \multicolumn{2}{c}{\cellcolor[HTML]{6200C9}{\color[HTML]{FFFFFF} 35}} &
  \multicolumn{2}{c}{\cellcolor[HTML]{CBCEFB}{\color[HTML]{000000} 2}} \\
CO &
  \multicolumn{2}{c}{\cellcolor[HTML]{340096}{\color[HTML]{FFFFFF} 127}} &
  \multicolumn{2}{c}{\cellcolor[HTML]{303498}{\color[HTML]{FFFFFF} 83}} &
  \multicolumn{2}{c}{\cellcolor[HTML]{303498}{\color[HTML]{FFFFFF} 75}} &
  \multicolumn{2}{c}{\cellcolor[HTML]{6200C9}{\color[HTML]{FFFFFF} 64}} &
  \multicolumn{2}{c}{\cellcolor[HTML]{9698ED}{\color[HTML]{000000} 5}} &
  \multicolumn{2}{c}{\cellcolor[HTML]{CBCEFB}{\color[HTML]{000000} 4}} &
  \multicolumn{2}{c}{\cellcolor[HTML]{9698ED}{\color[HTML]{000000} 5}} &
  \multicolumn{2}{c}{\cellcolor[HTML]{6665CD}{\color[HTML]{000000} 12}} &
  \multicolumn{2}{c}{\cellcolor[HTML]{FFFFFF}{\color[HTML]{000000} 0}} \\
\hline
\end{tabular}
    }
    \caption{GPT-3.5 error type analysis on MathQA.}
    \label{tab:MathQA GPT-3.5 error type analysis}
  \end{minipage}
}
\end{table*}

\begin{table*}[ht]
\centering
\adjustbox{center}{
  \begin{minipage}[t]{0.48\linewidth}
    \centering
    \scalebox{0.8}{
    \begin{tabular}{l|clclclclclclclclcl}
\hline
 &
  \multicolumn{2}{c}{CA} &
  \multicolumn{2}{c}{CO} &
  \multicolumn{2}{c}{CV} &
  \multicolumn{2}{c}{CS} &
  \multicolumn{2}{c}{MS} &
  \multicolumn{2}{c}{HA} &
  \multicolumn{2}{c}{UC} &
  \multicolumn{2}{c}{OP} &
  \multicolumn{2}{c}{FC} \\ \hline
CA &
  \multicolumn{2}{c}{\cellcolor[HTML]{340096}{\color[HTML]{FFFFFF} 401}} &
  \multicolumn{2}{c}{\cellcolor[HTML]{CBCEFB}{\color[HTML]{000000} 1}} &
  \multicolumn{2}{c}{\cellcolor[HTML]{9698ED}{\color[HTML]{000000} 4}} &
  \multicolumn{2}{c}{\cellcolor[HTML]{9698ED}{\color[HTML]{000000} 3}} &
  \multicolumn{2}{c}{\cellcolor[HTML]{6665CD}{\color[HTML]{000000} 6}} &
  \multicolumn{2}{c}{\cellcolor[HTML]{FFFFFF}{\color[HTML]{000000} 0}} &
  \multicolumn{2}{c}{\cellcolor[HTML]{FFFFFF}{\color[HTML]{000000} 0}} &
  \multicolumn{2}{c}{\cellcolor[HTML]{CBCEFB}{\color[HTML]{000000} 1}} &
  \multicolumn{2}{c}{\cellcolor[HTML]{FFFFFF}{\color[HTML]{000000} 0}} \\
CO &
  \multicolumn{2}{c}{\cellcolor[HTML]{340096}{\color[HTML]{FFFFFF} 256}} &
  \multicolumn{2}{c}{\cellcolor[HTML]{303498}{\color[HTML]{FFFFFF} 234}} &
  \multicolumn{2}{c}{\cellcolor[HTML]{6665CD}{\color[HTML]{000000} 22}} &
  \multicolumn{2}{c}{\cellcolor[HTML]{3303CF}{\color[HTML]{FFFFFF} 62}} &
  \multicolumn{2}{c}{\cellcolor[HTML]{CBCEFB}{\color[HTML]{000000} 8}} &
  \multicolumn{2}{c}{\cellcolor[HTML]{FFFFFF}{\color[HTML]{000000} 0}} &
  \multicolumn{2}{c}{\cellcolor[HTML]{FFFFFF}{\color[HTML]{000000} 0}} &
  \multicolumn{2}{c}{\cellcolor[HTML]{9698ED}{\color[HTML]{000000} 18}} &
  \multicolumn{2}{c}{\cellcolor[HTML]{FFFFFF}{\color[HTML]{000000} 0}} \\
\hline
\end{tabular}
}\caption{GPT-4 error type analysis on GSM8K.}
\label{tab:GSM8K GPT-4 error type analysis}
  \end{minipage}%
  \hspace{1.5cm}
  \begin{minipage}[t]{0.48\linewidth}
    \centering
    \scalebox{0.8}{
    \begin{tabular}{l|rlrlrlrlrlclllrlrl}
\hline
 &
  \multicolumn{2}{c}{CA} &
  \multicolumn{2}{c}{CO} &
  \multicolumn{2}{c}{CV} &
  \multicolumn{2}{c}{CS} &
  \multicolumn{2}{c}{MS} &
  \multicolumn{2}{c}{HA} &
  \multicolumn{2}{c}{UC} &
  \multicolumn{2}{c}{OP} &
  \multicolumn{2}{c}{FC} \\ \hline
CA &
  \multicolumn{2}{c}{\cellcolor[HTML]{340096}{\color[HTML]{FFFFFF} 428}} &
  \multicolumn{2}{c}{\cellcolor[HTML]{6665CD}{\color[HTML]{000000} 15}} &
  \multicolumn{2}{c}{\cellcolor[HTML]{3303CF}{\color[HTML]{FFFFFF} 19}} &
  \multicolumn{2}{c}{\cellcolor[HTML]{6200C9}{\color[HTML]{FFFFFF} 23}} &
  \multicolumn{2}{c}{\cellcolor[HTML]{9698ED}{\color[HTML]{000000} 8}} &
  \multicolumn{2}{c}{\cellcolor[HTML]{FFFFFF}{\color[HTML]{000000} 0}} &
  \multicolumn{2}{c}{\cellcolor[HTML]{FFFFFF}{\color[HTML]{000000} 0}} &
  \multicolumn{2}{c}{\cellcolor[HTML]{CBCEFB}{\color[HTML]{000000} 7}} &
  \multicolumn{2}{c}{\cellcolor[HTML]{9698ED}{\color[HTML]{000000} 8}} \\
CO &
  \multicolumn{2}{c}{\cellcolor[HTML]{340096}{\color[HTML]{FFFFFF} 248}} &
  \multicolumn{2}{c}{\cellcolor[HTML]{303498}{\color[HTML]{FFFFFF} 167}} &
  \multicolumn{2}{c}{\cellcolor[HTML]{6200C9}{\color[HTML]{FFFFFF} 83}} &
  \multicolumn{2}{c}{\cellcolor[HTML]{3303CF}{\color[HTML]{FFFFFF} 46}} &
  \multicolumn{2}{c}{\cellcolor[HTML]{CBCEFB}{\color[HTML]{000000} 4}} &
  \multicolumn{2}{c}{\cellcolor[HTML]{CBCEFB}{\color[HTML]{000000} 2}} &
  \multicolumn{2}{c}{\cellcolor[HTML]{FFFFFF}{\color[HTML]{000000} 0}} &
  \multicolumn{2}{c}{\cellcolor[HTML]{6665CD}{\color[HTML]{000000} 21}} &
  \multicolumn{2}{c}{\cellcolor[HTML]{9698ED}{\color[HTML]{000000} 10}} \\
\hline
\end{tabular}
}\caption{GPT-4 error type analysis on MathQA.}
\label{tab:MathQA GPT-4 error type analysis}
  \end{minipage}
}
\end{table*}

\begin{table*}[ht]
\centering
\adjustbox{center}{
  \begin{minipage}[t]{0.48\linewidth}
    \centering
    \scalebox{0.8}{
\begin{tabular}{l|clclclclclclclclcl}
\hline
 &
  \multicolumn{2}{c}{CA} &
  \multicolumn{2}{c}{CO} &
  \multicolumn{2}{c}{CV} &
  \multicolumn{2}{c}{CS} &
  \multicolumn{2}{c}{MS} &
  \multicolumn{2}{c}{HA} &
  \multicolumn{2}{c}{UC} &
  \multicolumn{2}{c}{OP} &
  \multicolumn{2}{c}{FC} \\ \hline
CA &
  \multicolumn{2}{c}{\cellcolor[HTML]{340096}{\color[HTML]{FFFFFF} 242}} &
  \multicolumn{2}{c}{\cellcolor[HTML]{3303CF}{\color[HTML]{FFFFFF} 25}} &
  \multicolumn{2}{c}{\cellcolor[HTML]{9698ED}{\color[HTML]{000000} 7}} &
  \multicolumn{2}{c}{\cellcolor[HTML]{6665CD}{\color[HTML]{000000} 16}} &
  \multicolumn{2}{c}{\cellcolor[HTML]{9698ED}{\color[HTML]{000000} 8}} &
  \multicolumn{2}{c}{\cellcolor[HTML]{CBCEFB}{\color[HTML]{000000} 1}} &
  \multicolumn{2}{c}{\cellcolor[HTML]{FFFFFF}{\color[HTML]{000000} 0}} &
  \multicolumn{2}{c}{\cellcolor[HTML]{CBCEFB}{\color[HTML]{000000} 2}} &
  \multicolumn{2}{c}{\cellcolor[HTML]{FFFFFF}{\color[HTML]{000000} 0}} \\
CO &
  \multicolumn{2}{c}{\cellcolor[HTML]{303498}{\color[HTML]{FFFFFF} 180}} &
  \multicolumn{2}{c}{\cellcolor[HTML]{340096}{\color[HTML]{FFFFFF} 351}} &
  \multicolumn{2}{c}{\cellcolor[HTML]{CBCEFB}{\color[HTML]{000000} 1}} &
  \multicolumn{2}{c}{\cellcolor[HTML]{6665CD}{\color[HTML]{000000} 12}} &
  \multicolumn{2}{c}{\cellcolor[HTML]{3303CF}{\color[HTML]{FFFFFF} 31}} &
  \multicolumn{2}{c}{\cellcolor[HTML]{FFFFFF}{\color[HTML]{000000} 0}} &
  \multicolumn{2}{c}{\cellcolor[HTML]{FFFFFF}{\color[HTML]{000000} 0}} &
  \multicolumn{2}{c}{\cellcolor[HTML]{CBCEFB}{\color[HTML]{000000} 2}} &
  \multicolumn{2}{c}{\cellcolor[HTML]{FFFFFF}{\color[HTML]{000000} 0}} \\
\hline
\end{tabular}
}\caption{GLM-4 error type analysis on GSM8K.}
\label{tab:GSM8K GLM-4 error type analysis}
  \end{minipage}%
  \hspace{1.5cm}
  \begin{minipage}[t]{0.48\linewidth}
    \centering
    \scalebox{0.8}{
\begin{tabular}{l|clclclclclclclclcl}
\hline
 &
  \multicolumn{2}{c}{CA} &
  \multicolumn{2}{c}{CO} &
  \multicolumn{2}{c}{CV} &
  \multicolumn{2}{c}{CS} &
  \multicolumn{2}{c}{MS} &
  \multicolumn{2}{c}{HA} &
  \multicolumn{2}{c}{UC} &
  \multicolumn{2}{c}{OP} &
  \multicolumn{2}{c}{FC} \\ \hline
CA &
  \multicolumn{2}{c}{\cellcolor[HTML]{340096}{\color[HTML]{FFFFFF} 241}} &
  \multicolumn{2}{c}{\cellcolor[HTML]{6665CD}12} &
  \multicolumn{2}{c}{\cellcolor[HTML]{6665CD}11} &
  \multicolumn{2}{c}{\cellcolor[HTML]{3303CF}{\color[HTML]{FFFFFF} 31}} &
  \multicolumn{2}{c}{\cellcolor[HTML]{9698ED}9} &
  \multicolumn{2}{c}{0} &
  \multicolumn{2}{c}{\cellcolor[HTML]{3303CF}{\color[HTML]{FFFFFF} 32}} &
  \multicolumn{2}{c}{\cellcolor[HTML]{CBCEFB}3} &
  \multicolumn{2}{c}{\cellcolor[HTML]{6665CD}12} \\
CO &
  \multicolumn{2}{c}{\cellcolor[HTML]{303498}{\color[HTML]{FFFFFF} 206}} &
  \multicolumn{2}{c}{\cellcolor[HTML]{340096}{\color[HTML]{FFFFFF} 238}} &
  \multicolumn{2}{c}{\cellcolor[HTML]{9698ED}14} &
  \multicolumn{2}{c}{\cellcolor[HTML]{3303CF}{\color[HTML]{FFFFFF} 44}} &
  \multicolumn{2}{c}{\cellcolor[HTML]{6665CD}32} &
  \multicolumn{2}{c}{0} &
  \multicolumn{2}{c}{0} &
  \multicolumn{2}{c}{\cellcolor[HTML]{CBCEFB}6} &
  \multicolumn{2}{c}{\cellcolor[HTML]{CBCEFB}5} \\
CV &
  \multicolumn{2}{c}{\cellcolor[HTML]{340096}{\color[HTML]{FFFFFF} 356}} &
  \multicolumn{2}{c}{\cellcolor[HTML]{6665CD}10} &
  \multicolumn{2}{c}{\cellcolor[HTML]{303498}{\color[HTML]{FFFFFF} 98}} &
  \multicolumn{2}{c}{\cellcolor[HTML]{6200C9}{\color[HTML]{FFFFFF} 46}} &
  \multicolumn{2}{c}{\cellcolor[HTML]{9698ED}5} &
  \multicolumn{2}{c}{\cellcolor[HTML]{CBCEFB}1} &
  \multicolumn{2}{c}{\cellcolor[HTML]{9698ED}6} &
  \multicolumn{2}{c}{\cellcolor[HTML]{6665CD}8} &
  \multicolumn{2}{c}{\cellcolor[HTML]{3303CF}{\color[HTML]{FFFFFF} 24}} \\
\hline
\end{tabular}
}\caption{GLM-4 error type analysis on MathQA. }
\label{tab:MathQA GLM-4 error type analysis}
  \end{minipage}
}
\end{table*}

\begin{table*}[ht]
\centering
\adjustbox{center}{
  \begin{minipage}[t]{0.48\linewidth}
    \centering
    \scalebox{0.8}{
\begin{tabular}{l|clclclclclclclclcl}
\hline
 &
  \multicolumn{2}{c}{CA} &
  \multicolumn{2}{c}{CO} &
  \multicolumn{2}{c}{CV} &
  \multicolumn{2}{c}{CS} &
  \multicolumn{2}{c}{MS} &
  \multicolumn{2}{c}{HA} &
  \multicolumn{2}{c}{UC} &
  \multicolumn{2}{c}{OP} &
  \multicolumn{2}{c}{FC} \\ \hline
CA &
  \multicolumn{2}{c}{\cellcolor[HTML]{340096}{\color[HTML]{FFFFFF} 115}} &
  \multicolumn{2}{c}{0} &
  \multicolumn{2}{c}{\cellcolor[HTML]{6665CD}14} &
  \multicolumn{2}{c}{\cellcolor[HTML]{3303CF}{\color[HTML]{FFFFFF} 23}} &
  \multicolumn{2}{c}{\cellcolor[HTML]{CBCEFB}1} &
  \multicolumn{2}{c}{0} &
  \multicolumn{2}{c}{0} &
  \multicolumn{2}{c}{\cellcolor[HTML]{CBCEFB}1} &
  \multicolumn{2}{c}{0} \\
CO &
  \multicolumn{2}{c}{\cellcolor[HTML]{340096}{\color[HTML]{FFFFFF} 134}} &
  \multicolumn{2}{c}{\cellcolor[HTML]{9698ED}6} &
  \multicolumn{2}{c}{\cellcolor[HTML]{3303CF}{\color[HTML]{FFFFFF} 33}} &
  \multicolumn{2}{c}{\cellcolor[HTML]{6200C9}{\color[HTML]{FFFFFF} 45}} &
  \multicolumn{2}{c}{\cellcolor[HTML]{CBCEFB}3} &
  \multicolumn{2}{c}{0} &
  \multicolumn{2}{c}{0} &
  \multicolumn{2}{c}{0} &
  \multicolumn{2}{c}{0} \\
CV &
  \multicolumn{2}{c}{\cellcolor[HTML]{340096}{\color[HTML]{FFFFFF} 167}} &
  \multicolumn{2}{c}{0} &
  \multicolumn{2}{c}{\cellcolor[HTML]{6200C9}{\color[HTML]{FFFFFF} 74}} &
  \multicolumn{2}{c}{\cellcolor[HTML]{6200C9}{\color[HTML]{FFFFFF} 77}} &
  \multicolumn{2}{c}{0} &
  \multicolumn{2}{c}{0} &
  \multicolumn{2}{c}{0} &
  \multicolumn{2}{c}{\cellcolor[HTML]{CBCEFB}2} &
  \multicolumn{2}{c}{0} \\
CS &
  \multicolumn{2}{c}{\cellcolor[HTML]{340096}{\color[HTML]{FFFFFF} 205}} &
  \multicolumn{2}{c}{0} &
  \multicolumn{2}{c}{\cellcolor[HTML]{3303CF}{\color[HTML]{FFFFFF} 22}} &
  \multicolumn{2}{c}{\cellcolor[HTML]{6200C9}{\color[HTML]{FFFFFF} 75}} &
  \multicolumn{2}{c}{0} &
  \multicolumn{2}{c}{0} &
  \multicolumn{2}{c}{\cellcolor[HTML]{9698ED}2} &
  \multicolumn{2}{c}{\cellcolor[HTML]{CBCEFB}1} &
  \multicolumn{2}{c}{\cellcolor[HTML]{CBCEFB}1} \\
\hline
\end{tabular}
}\caption{Gemini Pro error type analysis on GSM8K.}
\label{tab:GSM8K Gemini Pro error type analysis}
  \end{minipage}%
  \hspace{1.2cm}
  \begin{minipage}[t]{0.48\linewidth}
    \centering
    \scalebox{0.8}{
\begin{tabular}{l|clclclclclclclclcl}
\hline
 &
  \multicolumn{2}{c}{CA} &
  \multicolumn{2}{c}{CO} &
  \multicolumn{2}{c}{CV} &
  \multicolumn{2}{c}{CS} &
  \multicolumn{2}{c}{MS} &
  \multicolumn{2}{c}{HA} &
  \multicolumn{2}{c}{UC} &
  \multicolumn{2}{c}{OP} &
  \multicolumn{2}{c}{FC} \\ \hline
CA &
  \multicolumn{2}{c}{\cellcolor[HTML]{340096}{\color[HTML]{FFFFFF} 80}} &
  \multicolumn{2}{c}{0} &
  \multicolumn{2}{c}{\cellcolor[HTML]{3303CF}{\color[HTML]{FFFFFF} 21}} &
  \multicolumn{2}{c}{\cellcolor[HTML]{6665CD}17} &
  \multicolumn{2}{c}{0} &
  \multicolumn{2}{c}{0} &
  \multicolumn{2}{c}{0} &
  \multicolumn{2}{c}{0} &
  \multicolumn{2}{c}{\cellcolor[HTML]{CBCEFB}1} \\
CO &
  \multicolumn{2}{c}{\cellcolor[HTML]{340096}{\color[HTML]{FFFFFF} 87}} &
  \multicolumn{2}{c}{\cellcolor[HTML]{6665CD}18} &
  \multicolumn{2}{c}{\cellcolor[HTML]{340096}{\color[HTML]{FFFFFF} 88}} &
  \multicolumn{2}{c}{\cellcolor[HTML]{303498}{\color[HTML]{FFFFFF} 74}} &
  \multicolumn{2}{c}{\cellcolor[HTML]{9698ED}3} &
  \multicolumn{2}{c}{0} &
  \multicolumn{2}{c}{\cellcolor[HTML]{CBCEFB}2} &
  \multicolumn{2}{c}{\cellcolor[HTML]{9698ED}3} &
  \multicolumn{2}{c}{\cellcolor[HTML]{CBCEFB}1} \\
CV &
  \multicolumn{2}{c}{\cellcolor[HTML]{340096}{\color[HTML]{FFFFFF} 140}} &
  \multicolumn{2}{c}{0} &
  \multicolumn{2}{c}{\cellcolor[HTML]{6200C9}{\color[HTML]{FFFFFF} 87}} &
  \multicolumn{2}{c}{\cellcolor[HTML]{3303CF}{\color[HTML]{FFFFFF} 43}} &
  \multicolumn{2}{c}{\cellcolor[HTML]{6665CD}5} &
  \multicolumn{2}{c}{0} &
  \multicolumn{2}{c}{\cellcolor[HTML]{CBCEFB}1} &
  \multicolumn{2}{c}{\cellcolor[HTML]{CBCEFB}1} &
  \multicolumn{2}{c}{\cellcolor[HTML]{9698ED}3} \\
\hline
\end{tabular}
}
\caption{Gemini Pro error type analysis on MathQA.}
\label{tab:MathQA Gemini Pro error type analysis}
  \end{minipage}
}
\end{table*}

\begin{table*}[ht]
\centering
\adjustbox{center}{
  \begin{minipage}[t]{0.48\linewidth}
    \centering
    \scalebox{0.8}{
    \begin{tabular}{l|clclclclclclclclcl}
\hline
 &
  \multicolumn{2}{c}{CA} &
  \multicolumn{2}{c}{CO} &
  \multicolumn{2}{c}{CV} &
  \multicolumn{2}{c}{CS} &
  \multicolumn{2}{c}{MS} &
  \multicolumn{2}{c}{HA} &
  \multicolumn{2}{c}{UC} &
  \multicolumn{2}{c}{OP} &
  \multicolumn{2}{c}{FC} \\ \hline
CA &
  \multicolumn{2}{c}{\cellcolor[HTML]{340096}{\color[HTML]{FFFFFF} 191}} &
  \multicolumn{2}{c}{0} &
  \multicolumn{2}{c}{0} &
  \multicolumn{2}{c}{0} &
  \multicolumn{2}{c}{\cellcolor[HTML]{6200C9}{\color[HTML]{FFFFFF} 145}} &
  \multicolumn{2}{c}{\cellcolor[HTML]{3303CF}{\color[HTML]{FFFFFF} 76}} &
  \multicolumn{2}{c}{\cellcolor[HTML]{CBCEFB}6} &
  \multicolumn{2}{c}{0} &
  \multicolumn{2}{c}{0} \\
CO &
  \multicolumn{2}{c}{\cellcolor[HTML]{340096}{\color[HTML]{FFFFFF} 166}} &
  \multicolumn{2}{c}{0} &
  \multicolumn{2}{c}{0} &
  \multicolumn{2}{c}{0} &
  \multicolumn{2}{c}{\cellcolor[HTML]{6200C9}{\color[HTML]{FFFFFF} 119}} &
  \multicolumn{2}{c}{\cellcolor[HTML]{3303CF}{\color[HTML]{FFFFFF} 66}} &
  \multicolumn{2}{c}{0} &
  \multicolumn{2}{c}{0} &
  \multicolumn{2}{c}{0} \\
CV &
  \multicolumn{2}{c}{\cellcolor[HTML]{340096}{\color[HTML]{FFFFFF} 204}} &
  \multicolumn{2}{c}{\cellcolor[HTML]{CBCEFB}1} &
  \multicolumn{2}{c}{0} &
  \multicolumn{2}{c}{0} &
  \multicolumn{2}{c}{\cellcolor[HTML]{6200C9}{\color[HTML]{FFFFFF} 120}} &
  \multicolumn{2}{c}{\cellcolor[HTML]{3303CF}{\color[HTML]{FFFFFF} 89}} &
  \multicolumn{2}{c}{\cellcolor[HTML]{CBCEFB}2} &
  \multicolumn{2}{c}{0} &
  \multicolumn{2}{c}{0} \\
CS &
  \multicolumn{2}{c}{\cellcolor[HTML]{340096}{\color[HTML]{FFFFFF} 198}} &
  \multicolumn{2}{c}{0} &
  \multicolumn{2}{c}{\cellcolor[HTML]{CBCEFB}1} &
  \multicolumn{2}{c}{0} &
  \multicolumn{2}{c}{\cellcolor[HTML]{6200C9}{\color[HTML]{FFFFFF} 140}} &
  \multicolumn{2}{c}{\cellcolor[HTML]{3303CF}{\color[HTML]{FFFFFF} 82}} &
  \multicolumn{2}{c}{\cellcolor[HTML]{CBCEFB}2} &
  \multicolumn{2}{c}{0} &
  \multicolumn{2}{c}{0} \\
\hline
\end{tabular}
}\caption{LLaMA-2-7B error type analysis on GSM8K.}
\label{tab:GSM8K  LLaMA-2-7B error type analysis}
  \end{minipage}%
  \hspace{1.2cm}
  \begin{minipage}[t]{0.48\linewidth}
    \centering
    \scalebox{0.8}{
\begin{tabular}{l|clclclclclclclclcl}
\hline
 &
  \multicolumn{2}{c}{CA} &
  \multicolumn{2}{c}{CO} &
  \multicolumn{2}{c}{CV} &
  \multicolumn{2}{c}{CS} &
  \multicolumn{2}{c}{MS} &
  \multicolumn{2}{c}{HA} &
  \multicolumn{2}{c}{UC} &
  \multicolumn{2}{c}{OP} &
  \multicolumn{2}{c}{FC} \\ \hline
CA &
  \multicolumn{2}{c}{\cellcolor[HTML]{340096}{\color[HTML]{FFFFFF}191}} &
  \multicolumn{2}{c}{0} &
  \multicolumn{2}{c}{0} &
  \multicolumn{2}{c}{0} &
  \multicolumn{2}{c}{\cellcolor[HTML]{303498}{\color[HTML]{FFFFFF}112}} &
  \multicolumn{2}{c}{\cellcolor[HTML]{3303CF}{\color[HTML]{FFFFFF}43}} &
  \multicolumn{2}{c}{\cellcolor[HTML]{6665CD}17} &
  \multicolumn{2}{c}{0} &
  \multicolumn{2}{c}{0} \\
CO &
  \multicolumn{2}{c}{\cellcolor[HTML]{340096}{\color[HTML]{FFFFFF}163}} &
  \multicolumn{2}{c}{\cellcolor[HTML]{9698ED}4} &
  \multicolumn{2}{c}{0} &
  \multicolumn{2}{c}{0} &
  \multicolumn{2}{c}{\cellcolor[HTML]{6200C9}{\color[HTML]{FFFFFF}82}} &
  \multicolumn{2}{c}{\cellcolor[HTML]{3303CF}{\color[HTML]{FFFFFF}41}} &
  \multicolumn{2}{c}{\cellcolor[HTML]{CBCEFB}1} &
  \multicolumn{2}{c}{0} &
  \multicolumn{2}{c}{0} \\
CV &
  \multicolumn{2}{c}{\cellcolor[HTML]{340096}{\color[HTML]{FFFFFF}210}} &
  \multicolumn{2}{c}{0} &
  \multicolumn{2}{c}{0} &
  \multicolumn{2}{c}{0} &
  \multicolumn{2}{c}{\cellcolor[HTML]{303498}{\color[HTML]{FFFFFF}129}} &
  \multicolumn{2}{c}{\cellcolor[HTML]{3303CF}{\color[HTML]{FFFFFF}53}} &
  \multicolumn{2}{c}{\cellcolor[HTML]{9698ED}10} &
  \multicolumn{2}{c}{0} &
  \multicolumn{2}{c}{0} \\
CS &
  \multicolumn{2}{c}{\cellcolor[HTML]{340096}{\color[HTML]{FFFFFF}181}} &
  \multicolumn{2}{c}{0} &
  \multicolumn{2}{c}{\cellcolor[HTML]{CBCEFB}1} &
  \multicolumn{2}{c}{\cellcolor[HTML]{CBCEFB}1} &
  \multicolumn{2}{c}{\cellcolor[HTML]{303498}{\color[HTML]{FFFFFF}122}} &
  \multicolumn{2}{c}{\cellcolor[HTML]{3303CF}{\color[HTML]{FFFFFF}57}} &
  \multicolumn{2}{c}{\cellcolor[HTML]{9698ED}11} &
  \multicolumn{2}{c}{0} &
  \multicolumn{2}{c}{0} \\
\hline
\end{tabular}
}
\caption{LLaMA-2-7B error type analysis on MathQA.}
\label{tab:MathQA  LLaMA-2-7B error type analysis}
  \end{minipage}
}
\end{table*}

\begin{table*}[ht]
\centering
\adjustbox{center}{
  \begin{minipage}[t]{0.48\linewidth}
    \centering
    \scalebox{0.8}{
    \begin{tabular}{l|clclclclclclclclcl}
\hline
 &
  \multicolumn{2}{c}{CA} &
  \multicolumn{2}{c}{CO} &
  \multicolumn{2}{c}{CV} &
  \multicolumn{2}{c}{CS} &
  \multicolumn{2}{c}{MS} &
  \multicolumn{2}{c}{HA} &
  \multicolumn{2}{c}{UC} &
  \multicolumn{2}{c}{OP} &
  \multicolumn{2}{c}{FC} \\ \hline
CA &
  \multicolumn{2}{c}{\cellcolor[HTML]{9698ED}14} &
  \multicolumn{2}{c}{\cellcolor[HTML]{CBCEFB}2} &
  \multicolumn{2}{c}{\cellcolor[HTML]{9698ED}16} &
  \multicolumn{2}{c}{\cellcolor[HTML]{340096}{\color[HTML]{FFFFFF} 194}} &
  \multicolumn{2}{c}{\cellcolor[HTML]{3303CF}{\color[HTML]{FFFFFF} 32}} &
  \multicolumn{2}{c}{\cellcolor[HTML]{6200C9}{\color[HTML]{FFFFFF} 100}} &
  \multicolumn{2}{c}{\cellcolor[HTML]{CBCEFB}1} &
  \multicolumn{2}{c}{0} &
  \multicolumn{2}{c}{\cellcolor[HTML]{6665CD}19} \\
CO &
  \multicolumn{2}{c}{\cellcolor[HTML]{9698ED}8} &
  \multicolumn{2}{c}{\cellcolor[HTML]{9698ED}8} &
  \multicolumn{2}{c}{\cellcolor[HTML]{6665CD}26} &
  \multicolumn{2}{c}{\cellcolor[HTML]{340096}{\color[HTML]{FFFFFF} 174}} &
  \multicolumn{2}{c}{\cellcolor[HTML]{3303CF}{\color[HTML]{FFFFFF} 40}} &
  \multicolumn{2}{c}{\cellcolor[HTML]{6200C9}{\color[HTML]{FFFFFF} 86}} &
  \multicolumn{2}{c}{0} &
  \multicolumn{2}{c}{0} &
  \multicolumn{2}{c}{\cellcolor[HTML]{6665CD}27} \\
CV &
  \multicolumn{2}{c}{\cellcolor[HTML]{6665CD}22} &
  \multicolumn{2}{c}{\cellcolor[HTML]{CBCEFB}2} &
  \multicolumn{2}{c}{\cellcolor[HTML]{9698ED}13} &
  \multicolumn{2}{c}{\cellcolor[HTML]{340096}{\color[HTML]{FFFFFF} 185}} &
  \multicolumn{2}{c}{\cellcolor[HTML]{3303CF}{\color[HTML]{FFFFFF} 35}} &
  \multicolumn{2}{c}{\cellcolor[HTML]{303498}{\color[HTML]{FFFFFF} 113}} &
  \multicolumn{2}{c}{0} &
  \multicolumn{2}{c}{\cellcolor[HTML]{CBCEFB}1} &
  \multicolumn{2}{c}{\cellcolor[HTML]{6665CD}18} \\
\hline
\end{tabular}
}\caption{LLaMA-2-13B error type analysis on GSM8K.}
\label{tab:GSM8K  LLaMA-2-13B error type analysis}
  \end{minipage}%
  \hspace{1.2cm}
  \begin{minipage}[t]{0.48\linewidth}
    \centering
    \scalebox{0.8}{
\begin{tabular}{l|clclclclclclclclcl}
\hline
 &
  \multicolumn{2}{c}{CA} &
  \multicolumn{2}{c}{CO} &
  \multicolumn{2}{c}{CV} &
  \multicolumn{2}{c}{CS} &
  \multicolumn{2}{c}{MS} &
  \multicolumn{2}{c}{HA} &
  \multicolumn{2}{c}{UC} &
  \multicolumn{2}{c}{OP} &
  \multicolumn{2}{c}{FC} \\ \hline
CA &
  \multicolumn{2}{c}{\cellcolor[HTML]{3303CF}{\color[HTML]{FFFFFF} 28}} &
  \multicolumn{2}{c}{0} &
  \multicolumn{2}{c}{\cellcolor[HTML]{6665CD}14} &
  \multicolumn{2}{c}{\cellcolor[HTML]{340096}{\color[HTML]{FFFFFF} 167}} &
  \multicolumn{2}{c}{\cellcolor[HTML]{6200C9}{\color[HTML]{FFFFFF} 33}} &
  \multicolumn{2}{c}{\cellcolor[HTML]{303498}{\color[HTML]{FFFFFF} 118}} &
  \multicolumn{2}{c}{\cellcolor[HTML]{9698ED}8} &
  \multicolumn{2}{c}{\cellcolor[HTML]{CBCEFB}2} &
  \multicolumn{2}{c}{\cellcolor[HTML]{6665CD}19} \\
CO &
  \multicolumn{2}{c}{\cellcolor[HTML]{6665CD}23} &
  \multicolumn{2}{c}{\cellcolor[HTML]{6665CD}25} &
  \multicolumn{2}{c}{\cellcolor[HTML]{3303CF}{\color[HTML]{FFFFFF} 38}} &
  \multicolumn{2}{c}{\cellcolor[HTML]{340096}{\color[HTML]{FFFFFF} 149}} &
  \multicolumn{2}{c}{\cellcolor[HTML]{3303CF}{\color[HTML]{FFFFFF} 43}} &
  \multicolumn{2}{c}{\cellcolor[HTML]{6200C9}{\color[HTML]{FFFFFF} 80}} &
  \multicolumn{2}{c}{0} &
  \multicolumn{2}{c}{0} &
  \multicolumn{2}{c}{\cellcolor[HTML]{9698ED}10} \\
CV &
  \multicolumn{2}{c}{\cellcolor[HTML]{3303CF}{\color[HTML]{FFFFFF} 34}} &
  \multicolumn{2}{c}{0} &
  \multicolumn{2}{c}{\cellcolor[HTML]{9698ED}13} &
  \multicolumn{2}{c}{\cellcolor[HTML]{340096}{\color[HTML]{FFFFFF} 159}} &
  \multicolumn{2}{c}{\cellcolor[HTML]{6665CD}30} &
  \multicolumn{2}{c}{\cellcolor[HTML]{303498}{\color[HTML]{FFFFFF} 136}} &
  \multicolumn{2}{c}{\cellcolor[HTML]{9698ED}10} &
  \multicolumn{2}{c}{\cellcolor[HTML]{CBCEFB}2} &
  \multicolumn{2}{c}{\cellcolor[HTML]{6665CD}29} \\
\hline
\end{tabular}
}
\caption{LLaMA-2-13B error type analysis on MathQA.}
\label{tab:MathQA  LLaMA-2-13B error type analysis}
  \end{minipage}
}
\end{table*}


\begin{table*}[ht]

\centering
\scalebox{1}{
\begin{tabular}{ccccccccccccc|cccccc|c}
\hline
 &
  \multicolumn{2}{c}{CA} &
  \multicolumn{2}{c}{CO} &
  \multicolumn{2}{c}{CV} &
  \multicolumn{2}{c}{CS} &
  \multicolumn{2}{c}{MS} &
  \multicolumn{2}{c|}{HA} &
  \multicolumn{2}{c}{UC} &
  \multicolumn{2}{c}{OP} &
  \multicolumn{2}{c|}{FC} &
  Avg \\ \hline
Normal &
   &
   &
   &
   &
   &
   &
   &
   &
   &
   &
   &
   &
   &
   &
   &
   &
   &
   &
   \\
$\bullet$ GPT-3.5 &
  \multicolumn{2}{c}{0.72} &
  \multicolumn{2}{c}{0.71} &
  \multicolumn{2}{c}{0.77} &
  \multicolumn{2}{c}{0.84} &
  \multicolumn{2}{c}{0.71} &
  \multicolumn{2}{c|}{0.82} &
  \multicolumn{2}{c}{0.71} &
  \multicolumn{2}{c}{0.79} &
  \multicolumn{2}{c|}{0.77} &
  \multicolumn{1}{r}{0.759} \\
$\bullet$ GPT-4 &
  \multicolumn{2}{c}{0.76} &
  \multicolumn{2}{c}{0.89} &
  \multicolumn{2}{c}{0.89} &
  \multicolumn{2}{c}{0.89} &
  \multicolumn{2}{c}{0.87} &
  \multicolumn{2}{c|}{0.88} &
  \multicolumn{2}{c}{0.90} &
  \multicolumn{2}{c}{0.89} &
  \multicolumn{2}{c|}{0.89} &
  \multicolumn{1}{r}{0.875} \\
$\bullet$ GLM-4 &
  \multicolumn{2}{c}{0.63} &
  \multicolumn{2}{c}{0.80} &
  \multicolumn{2}{c}{0.83} &
  \multicolumn{2}{c}{0.83} &
  \multicolumn{2}{c}{0.76} &
  \multicolumn{2}{c|}{0.84} &
  \multicolumn{2}{c}{0.80} &
  \multicolumn{2}{c}{0.81} &
  \multicolumn{2}{c|}{0.85} &
  \multicolumn{1}{r}{0.795} \\
$\bullet$ Gemini Pro &
  \multicolumn{2}{c}{0.67} &
  \multicolumn{2}{c}{0.67} &
  \multicolumn{2}{c}{0.68} &
  \multicolumn{2}{c}{0.68} &
  \multicolumn{2}{c}{0.67} &
  \multicolumn{2}{c|}{0.78} &
  \multicolumn{2}{c}{0.69} &
  \multicolumn{2}{c}{0.75} &
  \multicolumn{2}{c|}{0.75} &
  \multicolumn{1}{r}{0.705} \\
$\bullet$ LLaMA-2-7B &
  \multicolumn{2}{c}{0.44} &
  \multicolumn{2}{c}{0.40} &
  \multicolumn{2}{c}{0.42} &
  \multicolumn{2}{c}{0.46} &
  \multicolumn{2}{c}{0.51} &
  \multicolumn{2}{c|}{0.45} &
  \multicolumn{2}{c}{0.41} &
  \multicolumn{2}{c}{0.48} &
  \multicolumn{2}{c|}{0.42} &
  0.445 \\
\hline
\end{tabular}
}
\caption{\textit{EP} prompt robustness testing on GSM8K.}
\label{tab:GSM8K EP prompt robustness testing，F1-score}
\end{table*}

\begin{table*}[ht]
\centering
\scalebox{1}{
\begin{tabular}{ccccccccccccc|cccccc|c}
\hline
 &
  \multicolumn{2}{c}{CA} &
  \multicolumn{2}{c}{CO} &
  \multicolumn{2}{c}{CV} &
  \multicolumn{2}{c}{CS} &
  \multicolumn{2}{c}{MS} &
  \multicolumn{2}{c|}{HA} &
  \multicolumn{2}{c}{UC} &
  \multicolumn{2}{c}{OP} &
  \multicolumn{2}{c|}{FC} &
  Avg \\ \hline
Normal &
   &
   &
   &
   &
   &
   &
   &
   &
   &
   &
   &
   &
   &
   &
   &
   &
   &
   &
   \\
$\bullet$ GPT-3.5 &
  \multicolumn{2}{c}{0.63} &
  \multicolumn{2}{c}{0.69} &
  \multicolumn{2}{c}{0.71} &
  \multicolumn{2}{c}{0.72} &
  \multicolumn{2}{c}{0.64} &
  \multicolumn{2}{c|}{0.72} &
  \multicolumn{2}{c}{0.61} &
  \multicolumn{2}{c}{0.72} &
  \multicolumn{2}{c|}{0.68} &
  0.680 \\
$\bullet$ GPT-4 &
  \multicolumn{2}{c}{0.64} &
  \multicolumn{2}{c}{0.73} &
  \multicolumn{2}{c}{0.78} &
  \multicolumn{2}{c}{0.73} &
  \multicolumn{2}{c}{0.73} &
  \multicolumn{2}{c|}{0.77} &
  \multicolumn{2}{c}{0.79} &
  \multicolumn{2}{c}{0.73} &
  \multicolumn{2}{c|}{0.77} &
  0.741 \\
$\bullet$ GLM-4 &
  \multicolumn{2}{c}{0.58} &
  \multicolumn{2}{c}{0.70} &
  \multicolumn{2}{c}{0.74} &
  \multicolumn{2}{c}{0.68} &
  \multicolumn{2}{c}{0.72} &
  \multicolumn{2}{c|}{0.81} &
  \multicolumn{2}{c}{0.71} &
  \multicolumn{2}{c}{0.76} &
  \multicolumn{2}{c|}{0.80} &
  0.722 \\
$\bullet$ Gemini Pro &
  \multicolumn{2}{c}{0.67} &
  \multicolumn{2}{c}{0.75} &
  \multicolumn{2}{c}{0.74} &
  \multicolumn{2}{c}{0.68} &
  \multicolumn{2}{c}{0.73} &
  \multicolumn{2}{c|}{0.72} &
  \multicolumn{2}{c}{0.68} &
  \multicolumn{2}{c}{0.75} &
  \multicolumn{2}{c|}{0.75} &
  0.718 \\
$\bullet$ LLaMA-2-7B &
  \multicolumn{2}{c}{0.69} &
  \multicolumn{2}{c}{0.65} &
  \multicolumn{2}{c}{0.76} &
  \multicolumn{2}{c}{0.73} &
  \multicolumn{2}{c}{0.77} &
  \multicolumn{2}{c|}{0.68} &
  \multicolumn{2}{c}{0.63} &
  \multicolumn{2}{c}{0.74} &
  \multicolumn{2}{c|}{0.70} &
  0.705 \\
\hline
\end{tabular}
}
\caption{\textit{EP} prompt robustness testing on MathQA.}
\label{tab:MathQA EP prompt robustness testing，F1-score}
\end{table*}

\begin{table*}[ht]
\centering
\tabcolsep=0.1cm
\scalebox{0.9}{
\begin{adjustbox}{center}
\begin{tabular}{ccccccccccccc|cccccc|c}
\hline
 &
  \multicolumn{2}{c}{CA} &
  \multicolumn{2}{c}{CO} &
  \multicolumn{2}{c}{CV} &
  \multicolumn{2}{c}{CS} &
  \multicolumn{2}{c}{MS} &
  \multicolumn{2}{c|}{HA} &
  \multicolumn{2}{c}{UC} &
  \multicolumn{2}{c}{OP} &
  \multicolumn{2}{c|}{FC} &
  Avg \\ 
   &
  \multicolumn{1}{c}{$acc_{2}$} &
  \multicolumn{1}{c}{$acc_{1}$} &
  \multicolumn{1}{c}{$acc_{2}$} &
  \multicolumn{1}{c}{$acc_{1}$} &
  \multicolumn{1}{c}{$acc_{2}$} &
  \multicolumn{1}{c}{$acc_{1}$} &
  \multicolumn{1}{c}{$acc_{2}$} &
  \multicolumn{1}{c}{$acc_{1}$} &
  \multicolumn{1}{c}{$acc_{2}$} &
  \multicolumn{1}{c}{$acc_{1}$} &
  \multicolumn{1}{c}{$acc_{2}$} &
  \multicolumn{1}{c|}{$acc_{1}$} &
  \multicolumn{1}{c}{$acc_{2}$} &
  \multicolumn{1}{c}{$acc_{1}$} &
  \multicolumn{1}{c}{$acc_{2}$} &
  \multicolumn{1}{c}{$acc_{1}$} &
  \multicolumn{1}{c}{$acc_{2}$} &
  \multicolumn{1}{c|}{$acc_{1}$} &
  \multicolumn{1}{c}{$acc_{2}$} \\
  \hline
Few-shot &
   &
   &
   &
   &
   &
   &
   &
   &
   &
   &
   &
   &
   &
   &
   &
   &
   &
   &
   \\
$\bullet$ GPT-3.5 &
  0.06 &
  0.28 &
  0.10 &
  0.21 &
  0.22 &
  0.51 &
  0.45 &
  0.78 &
  0.12 &
  0.30 &
  0.13 &
  0.37 &
  0.23 &
  0.57 &
  0.22 &
  0.58 &
  0.25 &
  0.53 &
  0.198 \\
$\bullet$ GPT-4 &
  0.59 &
  0.75 &
  0.89 &
  1.00 &
  0.95 &
  1.00 &
  0.89 &
  0.97 &
  0.76 &
  0.96 &
  0.73 &
  0.96 &
  0.96 &
  0.99 &
  0.93 &
  0.99 &
  0.87 &
  0.99 &
  0.841 \\
$\bullet$ GLM-4 &
  0.24 &
  0.68 &
  0.47 &
  0.97 &
  0.80 &
  0.99 &
  0.73 &
  0.99 &
  0.72 &
  0.97 &
  0.60 &
  0.99 &
  0.71 &
  0.99 &
  0.71 &
  0.99 &
  0.71 &
  1.00 &
  0.632 \\
$\bullet$ Gemini Pro &
  0.00 &
  0.01 &
  0.05 &
  0.09 &
  0.04 &
  0.06 &
  0.01 &
  0.04 &
  0.01 &
  0.06 &
  0.04 &
  0.09 &
  0.08 &
  0.12 &
  0.13 &
  0.25 &
  0.11 &
  0.17 &
  0.052 \\
\hline
\end{tabular}
\end{adjustbox}
}
\caption{\textit{ES} prompt robustness testing on GSM8K.}
\label{tab:GSM8K ES prompt robustness testing，Accuracy}
\end{table*}

\begin{table*}[ht]
\centering
\tabcolsep=0.1cm
\scalebox{0.9}{
\begin{adjustbox}{center}
\begin{tabular}{ccccccccccccc|cccccc|c}
\hline
 &
  \multicolumn{2}{c}{CA} &
  \multicolumn{2}{c}{CO} &
  \multicolumn{2}{c}{CV} &
  \multicolumn{2}{c}{CS} &
  \multicolumn{2}{c}{MS} &
  \multicolumn{2}{c|}{HA} &
  \multicolumn{2}{c}{UC} &
  \multicolumn{2}{c}{OP} &
  \multicolumn{2}{c|}{FC} &
  Avg \\ 
     &
  \multicolumn{1}{c}{$acc_{2}$} &
  \multicolumn{1}{c}{$acc_{1}$} &
  \multicolumn{1}{c}{$acc_{2}$} &
  \multicolumn{1}{c}{$acc_{1}$} &
  \multicolumn{1}{c}{$acc_{2}$} &
  \multicolumn{1}{c}{$acc_{1}$} &
  \multicolumn{1}{c}{$acc_{2}$} &
  \multicolumn{1}{c}{$acc_{1}$} &
  \multicolumn{1}{c}{$acc_{2}$} &
  \multicolumn{1}{c}{$acc_{1}$} &
  \multicolumn{1}{c}{$acc_{2}$} &
  \multicolumn{1}{c|}{$acc_{1}$} &
  \multicolumn{1}{c}{$acc_{2}$} &
  \multicolumn{1}{c}{$acc_{1}$} &
  \multicolumn{1}{c}{$acc_{2}$} &
  \multicolumn{1}{c}{$acc_{1}$} &
  \multicolumn{1}{c}{$acc_{2}$} &
  \multicolumn{1}{c|}{$acc_{1}$} &
  \multicolumn{1}{c}{$acc_{2}$} \\
  \hline
Few-shot &
   &
   &
   &
   &
   &
   &
   &
   &
   &
   &
   &
   &
   &
   &
   &
   &
   &
   &
   \\
$\bullet$ GPT-3.5 &
  0.07 &
  0.34 &
  0.09 &
  0.56 &
  0.26 &
  0.67 &
  0.19 &
  0.56 &
  0.20 &
  0.50 &
  0.20 &
  0.56 &
  0.07 &
  0.31 &
  0.21 &
  0.61 &
  0.12 &
  0.58 &
  0.157 \\
$\bullet$ GPT-4 &
  0.52 &
  0.84 &
  0.71 &
  0.92 &
  0.79 &
  0.99 &
  0.68 &
  0.96 &
  0.58 &
  0.96 &
  0.64 &
  0.99 &
  0.82 &
  0.99 &
  0.79 &
  1.00 &
  0.69 &
  1.00 &
  0.691 \\
$\bullet$ GLM-4 &
  0.23 &
  0.57 &
  0.50 &
  0.95 &
  0.62 &
  0.96 &
  \multicolumn{1}{r}{0.53} &
  \multicolumn{1}{r}{0.80} &
  0.52 &
  0.92 &
  0.45 &
  1.00 &
  0.56 &
  0.96 &
  0.59 &
  0.97 &
  0.46 &
  1.00 &
  0.496 \\
$\bullet$ Gemini Pro &
  0.00 &
  0.01 &
  0.05 &
  0.11 &
  0.02 &
  0.07 &
  0.01 &
  0.01 &
  0.03 &
  0.05 &
  0.03 &
  0.05 &
  0.00 &
  0.01 &
  0.08 &
  0.18 &
  0.06 &
  0.09 &
  0.031 \\
\hline
\end{tabular}
\end{adjustbox}
}
\caption{\textit{ES} prompt robustness testing on MathQA.}
\label{tab:MathQA ES prompt robustness testing，Accuracy}
\end{table*}

\begin{table*}[ht]
\centering
\tabcolsep=0.1cm
\scalebox{0.9}{
\begin{adjustbox}{center}
\begin{tabular}{ccccccccccccc|cccccc|c}
\hline
 &
  \multicolumn{2}{c}{CA} &
  \multicolumn{2}{c}{CO} &
  \multicolumn{2}{c}{CV} &
  \multicolumn{2}{c}{CS} &
  \multicolumn{2}{c}{MS} &
  \multicolumn{2}{c|}{HA} &
  \multicolumn{2}{c}{UC} &
  \multicolumn{2}{c}{OP} &
  \multicolumn{2}{c|}{FC} &
  Avg \\ 
     &
  \multicolumn{1}{c}{$acc_{3}$} &
  \multicolumn{1}{c}{$acc_{1}$} &
  \multicolumn{1}{c}{$acc_{3}$} &
  \multicolumn{1}{c}{$acc_{1}$} &
  \multicolumn{1}{c}{$acc_{3}$} &
  \multicolumn{1}{c}{$acc_{1}$} &
  \multicolumn{1}{c}{$acc_{3}$} &
  \multicolumn{1}{c}{$acc_{1}$} &
  \multicolumn{1}{c}{$acc_{3}$} &
  \multicolumn{1}{c}{$acc_{1}$} &
  \multicolumn{1}{c}{$acc_{3}$} &
  \multicolumn{1}{c|}{$acc_{1}$} &
  \multicolumn{1}{c}{$acc_{3}$} &
  \multicolumn{1}{c}{$acc_{1}$} &
  \multicolumn{1}{c}{$acc_{3}$} &
  \multicolumn{1}{c}{$acc_{1}$} &
  \multicolumn{1}{c}{$acc_{3}$} &
  \multicolumn{1}{c|}{$acc_{1}$} &
  \multicolumn{1}{c}{$acc_{3}$} \\
  \hline
Zero-shot &
   &
  \multicolumn{1}{r}{} &
   &
  \multicolumn{1}{r}{} &
   &
  \multicolumn{1}{r}{} &
   &
  \multicolumn{1}{r}{} &
   &
  \multicolumn{1}{r}{} &
   &
  \multicolumn{1}{r|}{} &
   &
  \multicolumn{1}{r}{} &
   &
  \multicolumn{1}{r}{} &
   &
  \multicolumn{1}{r|}{} &
   \\
$\bullet$ GPT-3.5 &
  0.21 &
  \multicolumn{1}{r}{0.45} &
  0.39 &
  \multicolumn{1}{r}{0.63} &
  0.16 &
  \multicolumn{1}{r}{0.70} &
  0.13 &
  \multicolumn{1}{r}{0.87} &
  0.01 &
  \multicolumn{1}{r}{0.75} &
  0.29 &
  \multicolumn{1}{r|}{0.88} &
  0.61 &
  \multicolumn{1}{r}{0.77} &
  0.01 &
  \multicolumn{1}{r}{0.75} &
  0.09 &
  \multicolumn{1}{r|}{0.83} &
  0.211 \\
$\bullet$ GPT-4 &
  0.62 &
  \multicolumn{1}{r}{0.65} &
  0.35 &
  \multicolumn{1}{r}{1.00} &
  0.63 &
  \multicolumn{1}{r}{1.00} &
  0.36 &
  \multicolumn{1}{r}{0.97} &
  0.01 &
  \multicolumn{1}{r}{0.97} &
  0.93 &
  \multicolumn{1}{r|}{1.00} &
  0.62 &
  \multicolumn{1}{r}{0.99} &
  0.28 &
  \multicolumn{1}{r}{0.99} &
  0.84 &
  \multicolumn{1}{r|}{0.99} &
  0.516 \\
\hline
\end{tabular}
\end{adjustbox}
}
\caption{\textit{ET} prompt robustness testing on GSM8K.}
\label{tab:GSM8K ET prompt robustness testing，Accuracy}
\end{table*}

\begin{table*}[ht]
\centering
\tabcolsep=0.1cm
\scalebox{0.9}{
\begin{adjustbox}{center}
\begin{tabular}{ccrcrcrcrcrcr|crcrcr|c}
\hline
 &
  \multicolumn{2}{c}{CA} &
  \multicolumn{2}{c}{CO} &
  \multicolumn{2}{c}{CV} &
  \multicolumn{2}{c}{CS} &
  \multicolumn{2}{c}{MS} &
  \multicolumn{2}{c|}{HA} &
  \multicolumn{2}{c}{UC} &
  \multicolumn{2}{c}{OP} &
  \multicolumn{2}{c|}{FC} &
  Avg \\ 
     &
  \multicolumn{1}{c}{$acc_{3}$} &
  \multicolumn{1}{c}{$acc_{1}$} &
  \multicolumn{1}{c}{$acc_{3}$} &
  \multicolumn{1}{c}{$acc_{1}$} &
  \multicolumn{1}{c}{$acc_{3}$} &
  \multicolumn{1}{c}{$acc_{1}$} &
  \multicolumn{1}{c}{$acc_{3}$} &
  \multicolumn{1}{c}{$acc_{1}$} &
  \multicolumn{1}{c}{$acc_{3}$} &
  \multicolumn{1}{c}{$acc_{1}$} &
  \multicolumn{1}{c}{$acc_{3}$} &
  \multicolumn{1}{c|}{$acc_{1}$} &
  \multicolumn{1}{c}{$acc_{3}$} &
  \multicolumn{1}{c}{$acc_{1}$} &
  \multicolumn{1}{c}{$acc_{3}$} &
  \multicolumn{1}{c}{$acc_{1}$} &
  \multicolumn{1}{c}{$acc_{3}$} &
  \multicolumn{1}{c|}{$acc_{1}$} &
  \multicolumn{1}{c}{$acc_{3}$} \\
  \hline
Zero-shot &
   &
   &
   &
   &
   &
   &
   &
   &
   &
   &
   &
   &
   &
   &
   &
   &
   &
   &
   \\
$\bullet$ GPT-3.5 &
  0.27 &
  0.47 &
  0.30 &
  0.72 &
  0.12 &
  0.70 &
  0.10 &
  0.65 &
  0.03 &
  0.73 &
  0.30 &
  0.90 &
  0.25 &
  0.43 &
  0.12 &
  0.74 &
  0.07 &
  0.74 &
  0.173 \\
$\bullet$ GPT-4 &
  0.70 &
  0.82 &
  0.30 &
  0.97 &
  0.56 &
  0.98 &
  0.27 &
  0.92 &
  0.08 &
  0.94 &
  0.85 &
  1.00 &
  0.46 &
  0.99 &
  0.43 &
  0.99 &
  0.68 &
  1.00 &
  0.481 \\
$\bullet$ GLM-4 &
  0.45 &
  0.67 &
  0.40 &
  0.97 &
  0.12 &
  0.96 &
  0.07 &
  0.79 &
  0.04 &
  0.88 &
  0.73 &
  1.00 &
  0.52 &
  0.95 &
  0.06 &
  0.97 &
  0.55 &
  1.00 &
  0.327 \\
$\bullet$ Gemini Pro &
  0.24 &
  0.31 &
  0.03 &
  0.64 &
  0.17 &
  0.66 &
  0.03 &
  0.51 &
  0.04 &
  0.56 &
  0.18 &
  0.89 &
  0.01 &
  0.49 &
  0.01 &
  0.70 &
  0.15 &
  0.67 &
  0.096 \\
\hline
\end{tabular}
\end{adjustbox}
}
\caption{\textit{ET} prompt robustness testing on MathQA.}
\label{tab:MathQA ET prompt robustness testing，Accuracy}
\end{table*}

\begin{table*}[ht]
\centering
\tabcolsep=0.1cm
\scalebox{0.9}{
\begin{adjustbox}{center}
\begin{tabular}{ccrcrcrcrcrcr|crcrcr|c}
\hline
 &
  \multicolumn{2}{c}{CA} &
  \multicolumn{2}{c}{CO} &
  \multicolumn{2}{c}{CV} &
  \multicolumn{2}{c}{CS} &
  \multicolumn{2}{c}{MS} &
  \multicolumn{2}{c|}{HA} &
  \multicolumn{2}{c}{UC} &
  \multicolumn{2}{c}{OP} &
  \multicolumn{2}{c|}{FC} &
  Avg \\ 
     &
  \multicolumn{1}{c}{$acc_{4}$} &
  \multicolumn{1}{c}{$acc_{1}$} &
  \multicolumn{1}{c}{$acc_{4}$} &
  \multicolumn{1}{c}{$acc_{1}$} &
  \multicolumn{1}{c}{$acc_{4}$} &
  \multicolumn{1}{c}{$acc_{1}$} &
  \multicolumn{1}{c}{$acc_{4}$} &
  \multicolumn{1}{c}{$acc_{1}$} &
  \multicolumn{1}{c}{$acc_{4}$} &
  \multicolumn{1}{c}{$acc_{1}$} &
  \multicolumn{1}{c}{$acc_{4}$} &
  \multicolumn{1}{c|}{$acc_{1}$} &
  \multicolumn{1}{c}{$acc_{4}$} &
  \multicolumn{1}{c}{$acc_{1}$} &
  \multicolumn{1}{c}{$acc_{4}$} &
  \multicolumn{1}{c}{$acc_{1}$} &
  \multicolumn{1}{c}{$acc_{4}$} &
  \multicolumn{1}{c|}{$acc_{1}$} &
  \multicolumn{1}{c}{$acc_{4}$} \\
  \hline
Zero-shot &
   &
   &
   &
   &
   &
   &
   &
   &
   &
   &
   &
   &
   &
   &
   &
   &
   &
   &
   \\
$\bullet$ GPT-3.5 &
  0.17 &
  0.20 &
  0.14 &
  0.16 &
  0.39 &
  0.43 &
  0.59 &
  0.65 &
  0.17 &
  0.27 &
  0.16 &
  0.33 &
  0.30 &
  0.36 &
  0.36 &
  0.48 &
  0.38 &
  0.40 &
  0.296 \\
$\bullet$ GPT-4 &
  0.66 &
  0.68 &
  0.97 &
  1.00 &
  0.94 &
  0.99 &
  0.95 &
  0.97 &
  0.87 &
  0.95 &
  0.89 &
  0.97 &
  0.93 &
  0.99 &
  0.95 &
  0.98 &
  0.95 &
  0.99 &
  0.901 \\
$\bullet$ GLM-4 &
  0.53 &
  0.55 &
  0.89 &
  0.97 &
  0.93 &
  0.97 &
  0.93 &
  0.98 &
  0.86 &
  0.98 &
  0.78 &
  0.99 &
  0.94 &
  0.99 &
  0.90 &
  0.98 &
  0.92 &
  1.00 &
  0.853 \\
$\bullet$ Gemini Pro &
  0.01 &
  0.01 &
  0.08 &
  0.09 &
  0.08 &
  0.08 &
  0.06 &
  0.07 &
  0.07 &
  0.09 &
  0.12 &
  0.14 &
  0.17 &
  0.17 &
  0.23 &
  0.26 &
  0.23 &
  0.23 &
  0.117 \\
\hline
\end{tabular}
\end{adjustbox}
}
\caption{\textit{EC} prompt robustness testing on GSM8K.}
\label{tab:GSM8K EC prompt robustness testing，Accuracy}
\end{table*}

\begin{table*}[!ht]
\centering
\tabcolsep=0.1cm
\scalebox{0.9}{
\begin{adjustbox}{center}
\begin{tabular}{ccrcrcrcrcrcr|crcrcc|c}
\hline
 &
  \multicolumn{2}{c}{CA} &
  \multicolumn{2}{c}{CO} &
  \multicolumn{2}{c}{CV} &
  \multicolumn{2}{c}{CS} &
  \multicolumn{2}{c}{MS} &
  \multicolumn{2}{c|}{HA} &
  \multicolumn{2}{c}{UC} &
  \multicolumn{2}{c}{OP} &
  \multicolumn{2}{c|}{FC} &
  Avg \\ 
&
  \multicolumn{1}{c}{$acc_{4}$} &
  \multicolumn{1}{c}{$acc_{1}$} &
  \multicolumn{1}{c}{$acc_{4}$} &
  \multicolumn{1}{c}{$acc_{1}$} &
  \multicolumn{1}{c}{$acc_{4}$} &
  \multicolumn{1}{c}{$acc_{1}$} &
  \multicolumn{1}{c}{$acc_{4}$} &
  \multicolumn{1}{c}{$acc_{1}$} &
  \multicolumn{1}{c}{$acc_{4}$} &
  \multicolumn{1}{c}{$acc_{1}$} &
  \multicolumn{1}{c}{$acc_{4}$} &
  \multicolumn{1}{c|}{$acc_{1}$} &
  \multicolumn{1}{c}{$acc_{4}$} &
  \multicolumn{1}{c}{$acc_{1}$} &
  \multicolumn{1}{c}{$acc_{4}$} &
  \multicolumn{1}{c}{$acc_{1}$} &
  \multicolumn{1}{c}{$acc_{4}$} &
  \multicolumn{1}{c|}{$acc_{1}$} &
  \multicolumn{1}{c}{$acc_{4}$} \\
  \hline
Zero-shot &
   &
   &
   &
   &
   &
   &
   &
   &
   &
   &
   &
   &
   &
   &
   &
   &
   &
  \multicolumn{1}{r|}{} &
   \\
$\bullet$ GPT-3.5 &
  0.15 &
  0.20 &
  0.33 &
  0.45 &
  0.40 &
  0.50 &
  0.34 &
  0.41 &
  0.25 &
  0.39 &
  0.20 &
  0.44 &
  0.14 &
  0.17 &
  0.34 &
  0.47 &
  0.32 &
  \multicolumn{1}{r|}{0.42} &
  0.274 \\
$\bullet$ GPT-4 &
  0.71 &
  0.80 &
  0.80 &
  0.89 &
  0.83 &
  0.98 &
  0.81 &
  0.92 &
  0.81 &
  0.95 &
  0.92 &
  0.99 &
  0.89 &
  0.99 &
  0.84 &
  0.99 &
  0.90 &
  \multicolumn{1}{r|}{1.00} &
  0.834 \\
$\bullet$ GLM-4 &
  0.48 &
  0.71 &
  0.69 &
  0.95 &
  0.79 &
  0.94 &
  0.71 &
  0.83 &
  0.58 &
  0.90 &
  0.79 &
  1.00 &
  0.74 &
  0.95 &
  0.67 &
  0.96 &
  0.78 &
  \multicolumn{1}{r|}{1.00} &
  0.692 \\
$\bullet$ Gemini Pro &
  0.00 &
  0.01 &
  0.13 &
  0.18 &
  0.10 &
  0.10 &
  0.01 &
  0.02 &
  0.06 &
  0.09 &
  0.09 &
  0.10 &
  0.05 &
  0.07 &
  0.13 &
  0.18 &
  0.17 &
  \multicolumn{1}{r|}{0.22} &
  0.082 \\
\hline
\end{tabular}
\end{adjustbox}
}
\caption{\textit{EC} prompt robustness testing on MathQA.}
\label{tab:MathQA EC prompt robustness testing，Accuracy}
\end{table*}

\begin{table*}[!ht]
\centering
\scalebox{1}{
\begin{adjustbox}{center}
\begin{tabular}{cclclclclclcl|clclcl|c}
\hline
 &
  \multicolumn{2}{c}{CA} &
  \multicolumn{2}{c}{CO} &
  \multicolumn{2}{c}{CV} &
  \multicolumn{2}{c}{CS} &
  \multicolumn{2}{c}{MS} &
  \multicolumn{2}{c|}{HA} &
  \multicolumn{2}{c}{UC} &
  \multicolumn{2}{c}{OP} &
  \multicolumn{2}{c|}{FC} &
  Avg \\ \hline
$\bullet$ GPT-3.5 &
  \multicolumn{2}{c}{0.83} &
  \multicolumn{2}{c}{0.72} &
  \multicolumn{2}{c}{0.89} &
  \multicolumn{2}{c}{0.81} &
  \multicolumn{2}{c}{0.81} &
  \multicolumn{2}{c|}{0.85} &
  \multicolumn{2}{c}{0.60} &
  \multicolumn{2}{c}{0.77} &
  \multicolumn{2}{c|}{0.76} &
  0.782 \\
$\bullet$ GPT-4 &
  \multicolumn{2}{c}{0.98} &
  \multicolumn{2}{c}{0.90} &
  \multicolumn{2}{c}{0.93} &
  \multicolumn{2}{c}{0.97} &
  \multicolumn{2}{c}{0.91} &
  \multicolumn{2}{c|}{0.98} &
  \multicolumn{2}{c}{0.93} &
  \multicolumn{2}{c}{0.97} &
  \multicolumn{2}{c|}{0.95} &
  0.947 \\
$\bullet$ GLM-4 &
  \multicolumn{2}{c}{0.97} &
  \multicolumn{2}{c}{0.96} &
  \multicolumn{2}{c}{0.93} &
  \multicolumn{2}{c}{0.94} &
  \multicolumn{2}{c}{0.92} &
  \multicolumn{2}{c|}{0.98} &
  \multicolumn{2}{c}{0.86} &
  \multicolumn{2}{c}{0.90} &
  \multicolumn{2}{c|}{0.89} &
  0.928 \\
$\bullet$ Gemini Pro &
  \multicolumn{2}{c}{0.81} &
  \multicolumn{2}{c}{0.70} &
  \multicolumn{2}{c}{0.76} &
  \multicolumn{2}{c}{0.81} &
  \multicolumn{2}{c}{0.81} &
  \multicolumn{2}{c|}{0.85} &
  \multicolumn{2}{c}{0.73} &
  \multicolumn{2}{c}{0.69} &
  \multicolumn{2}{c|}{0.81} &
  0.774 \\ \hline \hline
$\bullet$ LLaMA-2-7B &
  \multicolumn{2}{c}{0.13} &
  \multicolumn{2}{c}{0.05} &
  \multicolumn{2}{c}{0.12} &
  \multicolumn{2}{c}{0.13} &
  \multicolumn{2}{c}{0.19} &
  \multicolumn{2}{c|}{0.23} &
  \multicolumn{2}{c}{0.08} &
  \multicolumn{2}{c}{0.15} &
  \multicolumn{2}{c|}{0.07} &
  0.128 \\
\hline
\end{tabular}
\end{adjustbox}
}
\caption{Accuracy of tranditional task on GSM8K.}
\label{tab:GSM8K Original Cases Accuracy}
\end{table*}

\begin{table*}[!ht]
\centering
\scalebox{1}{
\begin{adjustbox}{center}
\begin{tabular}{cclclclclclcl|clclcl|c}
\hline
 &
  \multicolumn{2}{c}{CA} &
  \multicolumn{2}{c}{CO} &
  \multicolumn{2}{c}{CV} &
  \multicolumn{2}{c}{CS} &
  \multicolumn{2}{c}{MS} &
  \multicolumn{2}{c|}{HA} &
  \multicolumn{2}{c}{UC} &
  \multicolumn{2}{c}{OP} &
  \multicolumn{2}{c|}{FC} &
  Avg \\ \hline
$\bullet$ GPT-3.5 &
  \multicolumn{2}{c}{0.80} &
  \multicolumn{2}{c}{0.59} &
  \multicolumn{2}{c}{0.83} &
  \multicolumn{2}{c}{0.82} &
  \multicolumn{2}{c}{0.73} &
  \multicolumn{2}{c|}{0.75} &
  \multicolumn{2}{c}{0.68} &
  \multicolumn{2}{c}{0.70} &
  \multicolumn{2}{c|}{0.72} &
  0.736 \\
$\bullet$ GPT-4 &
  \multicolumn{2}{c}{0.92} &
  \multicolumn{2}{c}{0.74} &
  \multicolumn{2}{c}{0.89} &
  \multicolumn{2}{c}{0.90} &
  \multicolumn{2}{c}{0.87} &
  \multicolumn{2}{c|}{0.89} &
  \multicolumn{2}{c}{0.86} &
  \multicolumn{2}{c}{0.87} &
  \multicolumn{2}{c|}{0.87} &
  0.868 \\
$\bullet$ GLM-4 &
  \multicolumn{2}{c}{0.70} &
  \multicolumn{2}{c}{0.67} &
  \multicolumn{2}{c}{0.86} &
  \multicolumn{2}{c}{0.78} &
  \multicolumn{2}{c}{0.81} &
  \multicolumn{2}{c|}{0.84} &
  \multicolumn{2}{c}{0.73} &
  \multicolumn{2}{c}{0.80} &
  \multicolumn{2}{c|}{0.81} &
  0.778 \\
$\bullet$ Gemini Pro &
  \multicolumn{2}{c}{0.63} &
  \multicolumn{2}{c}{0.35} &
  \multicolumn{2}{c}{0.68} &
  \multicolumn{2}{c}{0.61} &
  \multicolumn{2}{c}{0.53} &
  \multicolumn{2}{c|}{0.68} &
  \multicolumn{2}{c}{0.52} &
  \multicolumn{2}{c}{0.66} &
  \multicolumn{2}{c|}{0.72} &
  0.598 \\ \hline \hline
$\bullet$ LLaMA-2-7B &
  \multicolumn{2}{c}{0.10} &
  \multicolumn{2}{c}{0.12} &
  \multicolumn{2}{c}{0.12} &
  \multicolumn{2}{c}{0.06} &
  \multicolumn{2}{c}{0.14} &
  \multicolumn{2}{c|}{0.09} &
  \multicolumn{2}{c}{0.04} &
  \multicolumn{2}{c}{0.07} &
  \multicolumn{2}{c|}{0.06} &
  0.089 \\
\hline
\end{tabular}
\end{adjustbox}
}
\caption{Accuracy of tranditional task on MathQA.}
\label{tab:MathQA Original Cases Accuracy}
\end{table*}

\begin{table*}[!ht]
\centering
\tabcolsep=0.15cm
\scalebox{0.8}{
\begin{adjustbox}{center}
\begin{tabular}{ccccccccccccc|cccccc|c}
\hline
 &
  \multicolumn{2}{c}{CA} &
  \multicolumn{2}{c}{CO} &
  \multicolumn{2}{c}{CV} &
  \multicolumn{2}{c}{CS} &
  \multicolumn{2}{c}{MS} &
  \multicolumn{2}{c|}{HA} &
  \multicolumn{2}{c}{UC} &
  \multicolumn{2}{c}{OP} &
  \multicolumn{2}{c|}{FC} &
  Avg \\ 
     &
  \multicolumn{1}{c}{$acc_{4}$} &
  \multicolumn{1}{c}{$acc_{1}$} &
  \multicolumn{1}{c}{$acc_{4}$} &
  \multicolumn{1}{c}{$acc_{1}$} &
  \multicolumn{1}{c}{$acc_{4}$} &
  \multicolumn{1}{c}{$acc_{1}$} &
  \multicolumn{1}{c}{$acc_{4}$} &
  \multicolumn{1}{c}{$acc_{1}$} &
  \multicolumn{1}{c}{$acc_{4}$} &
  \multicolumn{1}{c}{$acc_{1}$} &
  \multicolumn{1}{c}{$acc_{4}$} &
  \multicolumn{1}{c|}{$acc_{1}$} &
  \multicolumn{1}{c}{$acc_{4}$} &
  \multicolumn{1}{c}{$acc_{1}$} &
  \multicolumn{1}{c}{$acc_{4}$} &
  \multicolumn{1}{c}{$acc_{1}$} &
  \multicolumn{1}{c}{$acc_{4}$} &
  \multicolumn{1}{c|}{$acc_{1}$} &
  \multicolumn{1}{c}{$acc_{4}$} \\
  \hline
GPT-3.5 &
  \multicolumn{1}{l}{} &
  \multicolumn{1}{l}{} &
  \multicolumn{1}{l}{} &
  \multicolumn{1}{l}{} &
  \multicolumn{1}{l}{} &
  \multicolumn{1}{l}{} &
  \multicolumn{1}{l}{} &
  \multicolumn{1}{l}{} &
  \multicolumn{1}{l}{} &
  \multicolumn{1}{l}{} &
  \multicolumn{1}{l}{} &
  \multicolumn{1}{l|}{} &
  \multicolumn{1}{l}{} &
  \multicolumn{1}{l}{} &
  \multicolumn{1}{l}{} &
  \multicolumn{1}{l}{} &
  \multicolumn{1}{l}{} &
  \multicolumn{1}{l|}{} &
  \multicolumn{1}{l}{} \\
$\bullet$ EP &  \multicolumn{1}{c}{-} &
  \multicolumn{1}{c}{0.68} & \multicolumn{1}{c}{-} &
  \multicolumn{1}{c}{0.84} & \multicolumn{1}{c}{-} &
  \multicolumn{1}{c}{0.94} & \multicolumn{1}{c}{-} & 
  \multicolumn{1}{c}{0.92} & \multicolumn{1}{c}{-} &
  \multicolumn{1}{c}{0.68} & \multicolumn{1}{c}{-} &
  \multicolumn{1}{c|}{0.83} & \multicolumn{1}{c}{-} &
  \multicolumn{1}{c}{0.75} & \multicolumn{1}{c}{-} &
  \multicolumn{1}{c}{0.89} & \multicolumn{1}{c}{-} &
  \multicolumn{1}{c|}{0.91} & 
  0.827 \\
\hline
\end{tabular}
\end{adjustbox}
}
\caption{Accuracy of incomplete cases on GSM8K.}
\label{tab:GSM8K Incomplete Cases Accuracy}
\end{table*}

\clearpage
\begin{table*}[!ht]
\centering
\tabcolsep=0.15cm
\scalebox{0.8}{
\begin{adjustbox}{center}
\begin{tabular}{ccccccccccccc|cccccc|c}
\hline
 &
  \multicolumn{2}{c}{CA} &
  \multicolumn{2}{c}{CO} &
  \multicolumn{2}{c}{CV} &
  \multicolumn{2}{c}{CS} &
  \multicolumn{2}{c}{MS} &
  \multicolumn{2}{c|}{HA} &
  \multicolumn{2}{c}{UC} &
  \multicolumn{2}{c}{OP} &
  \multicolumn{2}{c|}{FC} &
  Avg \\ 
  &
  \multicolumn{1}{c}{$acc_{4}$} &
  \multicolumn{1}{c}{$acc_{1}$} &
  \multicolumn{1}{c}{$acc_{4}$} &
  \multicolumn{1}{c}{$acc_{1}$} &
  \multicolumn{1}{c}{$acc_{4}$} &
  \multicolumn{1}{c}{$acc_{1}$} &
  \multicolumn{1}{c}{$acc_{4}$} &
  \multicolumn{1}{c}{$acc_{1}$} &
  \multicolumn{1}{c}{$acc_{4}$} &
  \multicolumn{1}{c}{$acc_{1}$} &
  \multicolumn{1}{c}{$acc_{4}$} &
  \multicolumn{1}{c|}{$acc_{1}$} &
  \multicolumn{1}{c}{$acc_{4}$} &
  \multicolumn{1}{c}{$acc_{1}$} &
  \multicolumn{1}{c}{$acc_{4}$} &
  \multicolumn{1}{c}{$acc_{1}$} &
  \multicolumn{1}{c}{$acc_{4}$} &
  \multicolumn{1}{c|}{$acc_{1}$} &
  \multicolumn{1}{c}{$acc_{4}$} \\
  \hline
GPT-3.5 &
  \multicolumn{1}{l}{} &
  \multicolumn{1}{l}{} &
  \multicolumn{1}{l}{} &
  \multicolumn{1}{l}{} &
  \multicolumn{1}{l}{} &
  \multicolumn{1}{l}{} &
  \multicolumn{1}{l}{} &
  \multicolumn{1}{l}{} &
  \multicolumn{1}{l}{} &
  \multicolumn{1}{l}{} &
  \multicolumn{1}{l}{} &
  \multicolumn{1}{l|}{} &
  \multicolumn{1}{l}{} &
  \multicolumn{1}{l}{} &
  \multicolumn{1}{l}{} &
  \multicolumn{1}{l}{} &
  \multicolumn{1}{l}{} &
  \multicolumn{1}{l|}{} &
  \multicolumn{1}{l}{} \\
$\bullet$ EP &  \multicolumn{1}{c}{-} &
  \multicolumn{1}{c}{0.45} &\multicolumn{1}{c}{-} &
  \multicolumn{1}{c}{0.75} &\multicolumn{1}{c}{-} &
  \multicolumn{1}{c}{0.86} &\multicolumn{1}{c}{-} &
  \multicolumn{1}{c}{0.70} &\multicolumn{1}{c}{-} &
  \multicolumn{1}{c}{0.65} &\multicolumn{1}{c}{-} &
  \multicolumn{1}{c|}{0.97} &\multicolumn{1}{c}{-} &
  \multicolumn{1}{c}{0.51} &\multicolumn{1}{c}{-} &
  \multicolumn{1}{c}{0.65} &\multicolumn{1}{c}{-} &
  \multicolumn{1}{c|}{0.86} &
  0.711 \\
\hline
\end{tabular}
\end{adjustbox}
}
\caption{Accuracy of incomplete cases on MathQA.}
\label{tab:MathQA Incomplete Cases Accuracy}
\end{table*}

\end{document}